\documentclass[10pt]{article}
\usepackage{a4wide}
\usepackage[utf8]{inputenc}
\usepackage[T1]{fontenc}
\usepackage{graphicx}
\usepackage{grffile}
\usepackage{longtable}
\usepackage{wrapfig}
\usepackage{rotating}
\usepackage[normalem]{ulem}
\usepackage{amsmath}
\usepackage{textcomp}
\usepackage{amssymb}
\usepackage{capt-of}
\usepackage{hyperref}
\usepackage{gensymb}
\usepackage{multicol}
\usepackage{enumitem}
\usepackage{cancel}
\usepackage[dvipsnames]{xcolor}
\usepackage[linesnumbered,ruled,vlined]{algorithm2e}

\usepackage[normalem]{ulem}
\newcommand{\stkout}[1]{\ifmmode\text{\sout{\ensuremath{#1}}}\else\sout{#1}\fi}
\usepackage{tikz} 
\usepackage{lipsum}
\newcommand{\mybox}[4]{
	\begin{figure}[h!]
		\centering
		\begin{tikzpicture}
			\node[anchor=text,text width=\columnwidth-1.2cm, draw, rounded corners, line width=1pt, fill=#3, inner sep=5mm] (big) {\\#4};
			\node[draw, rounded corners, line width=.5pt, fill=#2, anchor=west, xshift=5mm] (small) at (big.north west) {#1};
		\end{tikzpicture}
	\end{figure}
}

\usepackage{xcolor}
\definecolor{azure}{rgb}{0.1, 0.5, 0.7}
\newcommand{\textbfazure}[1]{\textcolor{azure}{\textbf{#1}}}

\author{Gyubeom Edward Im\thanks{blog: \href{https://alida.tistory.com}{alida.tistory.com}, email: \href{mailto:criterion.im@gmail.com}{criterion.im@gmail.com}}}

\date{June 28, 2024}

\title{Notes on Kalman Filter\\(KF, EKF, ESKF, IEKF, IESKF)}

\usepackage{fancyhdr}
\begin{document}
	
	\maketitle
	
	\tableofcontents

\section{Preliminaries}
\textbfazure{The Kalman filter} is an algorithm that recursively \textbfazure{predicts and updates} the state of a \textbfazure{time-varying system} based on the \textbfazure{predictions} of the system model and \textbfazure{observations} containing noise. 

This section provides a brief overview of the foundational knowledge before delving into the Kalman filter itself.

\subsection{Estimation theory}
\textbfazure{Estimation theory} encompasses various methods for predicting the parameters or states of a model based on observed data. It is widely used in fields such as data analysis, signal processing, machine learning, finance, and robotics, serving as an essential tool for making accurate decisions in the presence of uncertainty. For more detailed information, refer to \href{https://alida.tistory.com/92}{this post}.

\begin{figure}[h!]
	\centering
	\includegraphics[width=10cm]{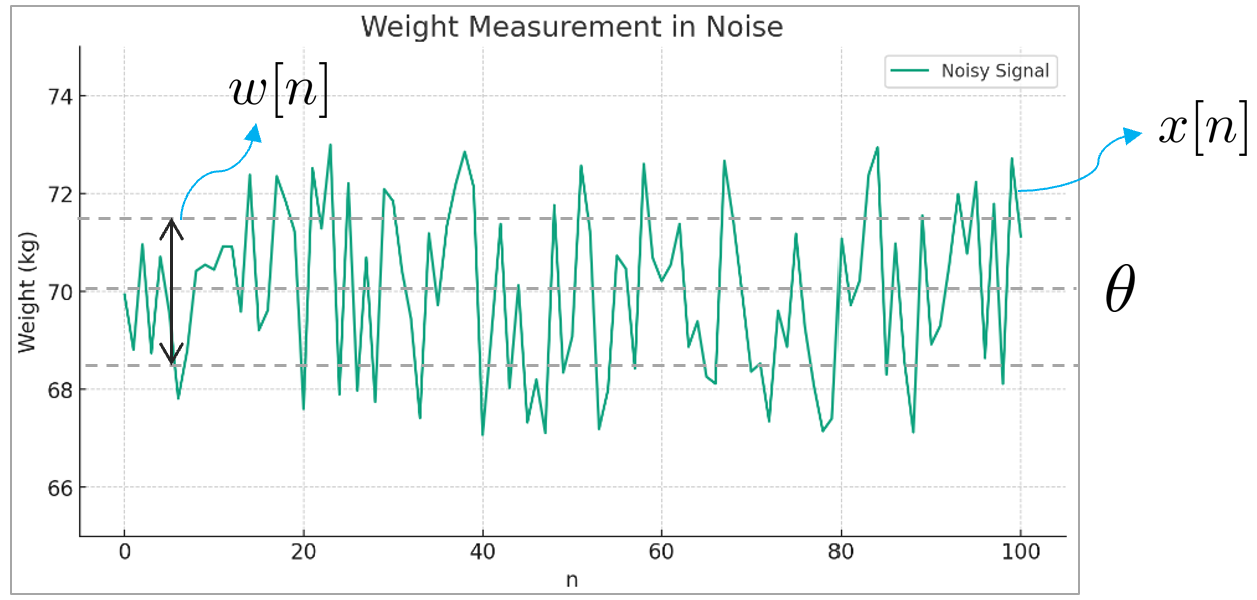}
\end{figure}

For example, suppose we are given data measured over a year for a person whose average weight is 70 kg, as shown in the figure above. The green data points forming the graph are $x[n]$, and the central value of these data points is the parameter $\theta$ that we aim to estimate. The degree of spread (i.e., variance) around $\theta$ can be mathematically represented by $w[n]$.

The problem of estimating the parameter $\theta$ based on the observed data $x[n]$ can be expressed by the following simple mathematical model:
\begin{equation} 
	\begin{aligned} 
		x[n] = \theta + w[n]  \quad n=0,1,\cdots,N-1 
	\end{aligned}  
\end{equation}
- $x[n]$: observed data\\
- $\theta$: parameter to be estimated\\
- $w[n]$: random noise\\

\subsection{Bayesian philosophy}
Estimation theory can be broadly divided into the frequentist and Bayesian perspectives, depending on how the parameter $\theta$ to be estimated is viewed.
\begin{itemize}
	\item \textbfazure{Frequentist}: A frequentist perspective views the parameter $\theta$ as an unknown deterministic parameter.
	\item \textbfazure{Bayesian}: A Bayesian perspective treats the parameter $\theta$ as a random variable with a prior probability distribution.
\end{itemize}

\begin{equation}   
	\begin{aligned}  
		& \text{Frequentist: } \quad \underbrace{x[n]}_{\text{r.v.}} = \underbrace{\theta}_{\text{deterministic}}+ w[n] \\  
		& \text{Bayesian: } \quad \underbrace{x[n]}_{\text{r.v.}} = \underbrace{\theta}_{\text{r.v.}} + w[n] \\  
	\end{aligned}   
\end{equation}

The Bayesian philosophy starts from the premise that if we have prior information about the parameter $\theta$, it can be used to make a better estimate. For this, a prior pdf of $\theta$ must be provided or computable.

In the Bayesian framework, since the parameter $\theta$ can also be modeled as a random variable, the following Bayesian rule holds:
\begin{equation}
	\begin{aligned}
		p(\theta|x) &  = \frac{p(\theta) p(x|\theta)}{p(x)}
	\end{aligned}
\end{equation}
- $p(\theta|x)$: posterior conditional probability distribution of the parameter $\theta$ given the observed data $x$\\
- $p(x|\theta)$: conditional probability distribution of $x$ given $\theta$, or likelihood\\
- $p(x)$: probability distribution of $x$\\
- $p(\theta)$: prior probability distribution of $\theta$\\

\subsection{Estimation problem}
The estimation problem can generally be depicted as follows:
\begin{figure}[h!]
	\centering
	\includegraphics[width=10cm]{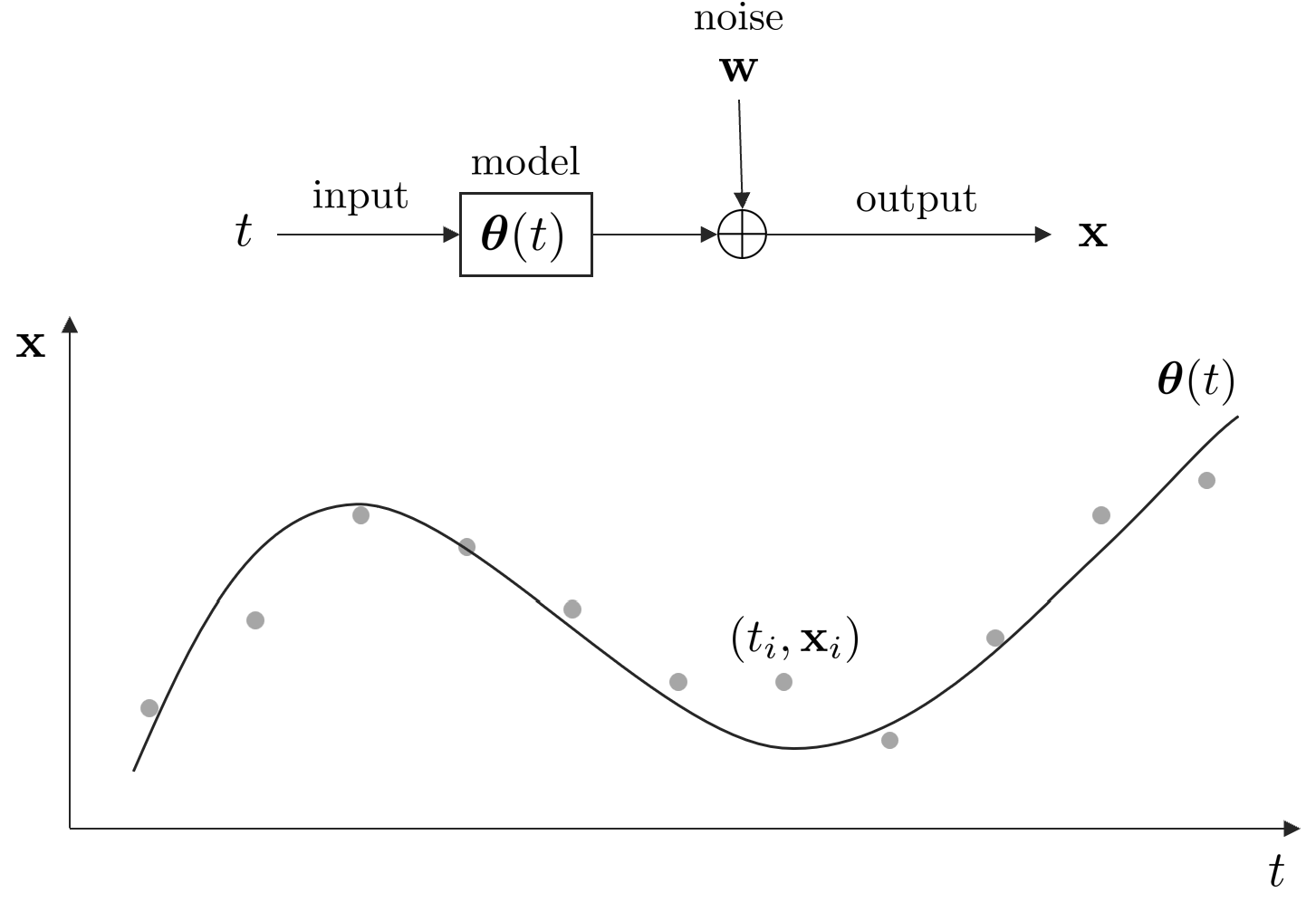}
\end{figure}

The goal of the estimation problem is to find the optimal model $\boldsymbol{\theta}(t)$ using the given data $(t_i, \mathbf{x}_i)$. The type of estimation problem varies depending on which point in time is being estimated.
\begin{figure}[h!]
	\centering
	\includegraphics[width=8cm]{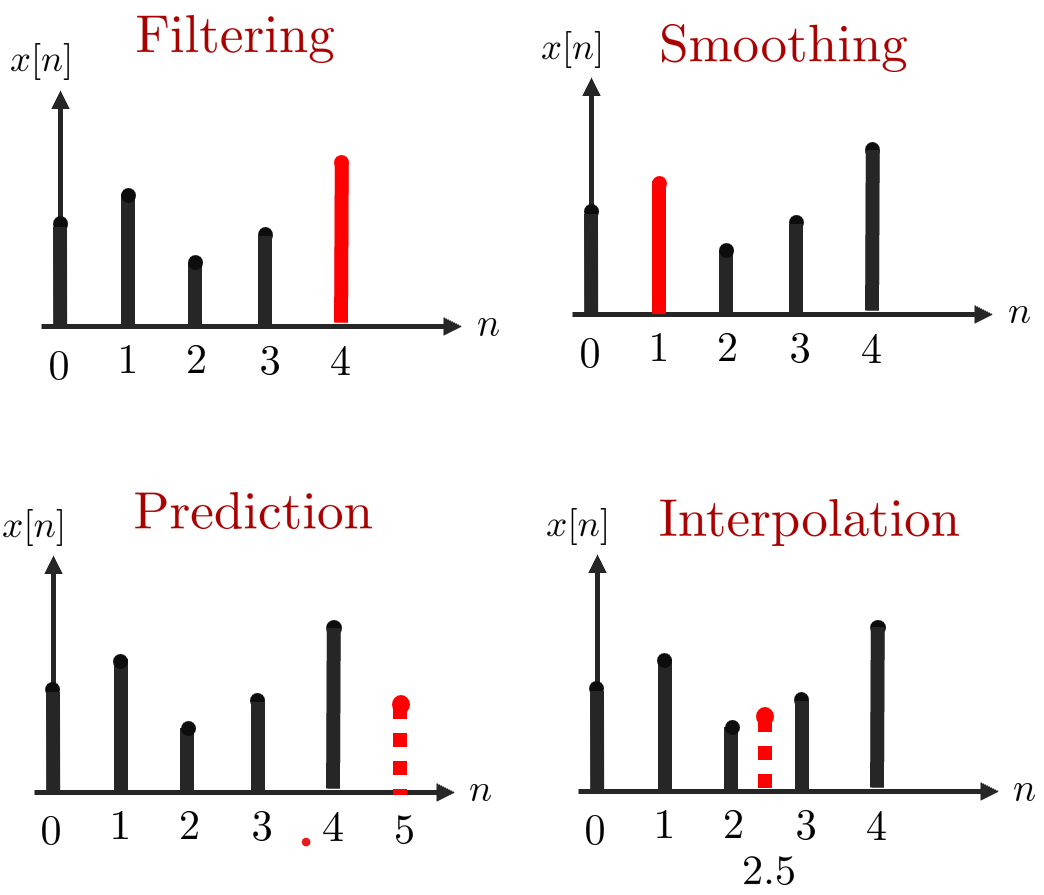}
\end{figure}

\begin{enumerate}
	\item \textbfazure{Filtering}: Filtering refers to the problem of estimating $\theta = x[N-1]$ given the observed data $\{x[0],x[1],\cdots,x[N-1]\}$. By estimating the optimal parameter, we aim to filter out the noise from the signal. Note that in filtering, the parameter depends only on current and past data.
	\item \textbfazure{Smoothing}: Smoothing refers to the problem of estimating an intermediate $\theta = s[n]$ given the observed data $\{x[0],x[1],\cdots,x[N-1]\}$. For instance, to estimate $s[1]$, all observed data are used. Naturally, smoothing cannot be performed until all data are observed.
	\item \textbfazure{Prediction}: Prediction refers to the problem of estimating $\theta = x[N-1+l]$ given the observed data $\{x[0],x[1],\cdots,x[N-1]\}$. Here, $l$ is an arbitrary positive number.
	\item \textbfazure{Interpolation}: Interpolation refers to the problem of estimating $\theta = x[n]$ given the observed data $\{x[0], \cdots ,x[n-1], x[n+1,\cdots,x[N-1]\}$.
\end{enumerate}

\subsection{Dynamic system}
A system with state variables that change over time can be modeled as follows:
\begin{equation}
	\begin{aligned}
		& \text{Motion Model: } & \mathbf{x}_{t} = \mathbf{f}(\mathbf{x}_{t-1}, \mathbf{u}_{t} ) + \mathbf{w}_{t} \\
		& \text{Observation Model: } &  \mathbf{z}_{t} = \mathbf{h}(\mathbf{x}_{t}) + \mathbf{v}_{t}
	\end{aligned}
\end{equation}
- $\mathbf{x}_{t}$: state variables at time $t$\\
- $\mathbf{u}_{t}$: control inputs at time $t$\\
- $\mathbf{z}_{t}$: observations at time $t$\\
- $\mathbf{f}(\mathbf{x}_{t-1}, \mathbf{u}_{t} )$: motion model function predicting the current state $\mathbf{x}_t$ from the previous state $\mathbf{x}_{t-1}$ and current control inputs $\mathbf{u}_{t}$\\
- $\mathbf{h}(\mathbf{x}_{t})$: observation model function converting current state variables $\mathbf{x}_{t}$ to observations $\mathbf{z}_{t}$\\
- $\mathbf{w}_{t}$: noise in the motion model at time $t$\\
- $\mathbf{v}_{t}$: noise in the observation model at time $t$\\

Revisiting the estimation problem using the above notation, we get:
\begin{figure}[h!]
	\centering
	\includegraphics[width=7cm]{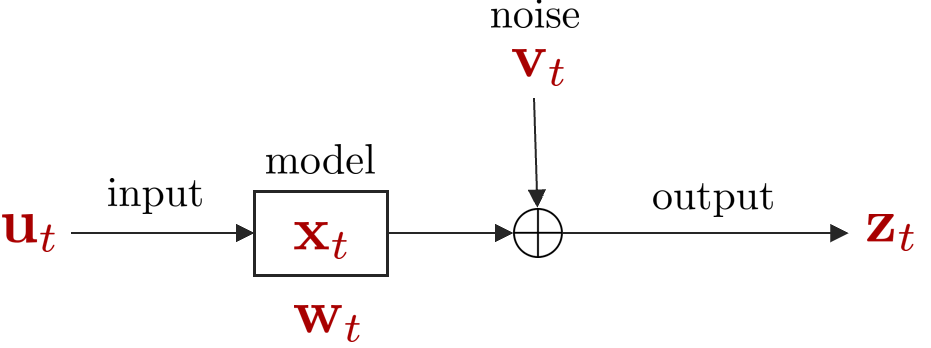}
\end{figure}

The dynamic system can be depicted as a graph as follows:
\begin{figure}[h!]
	\centering
	\includegraphics[width=12cm]{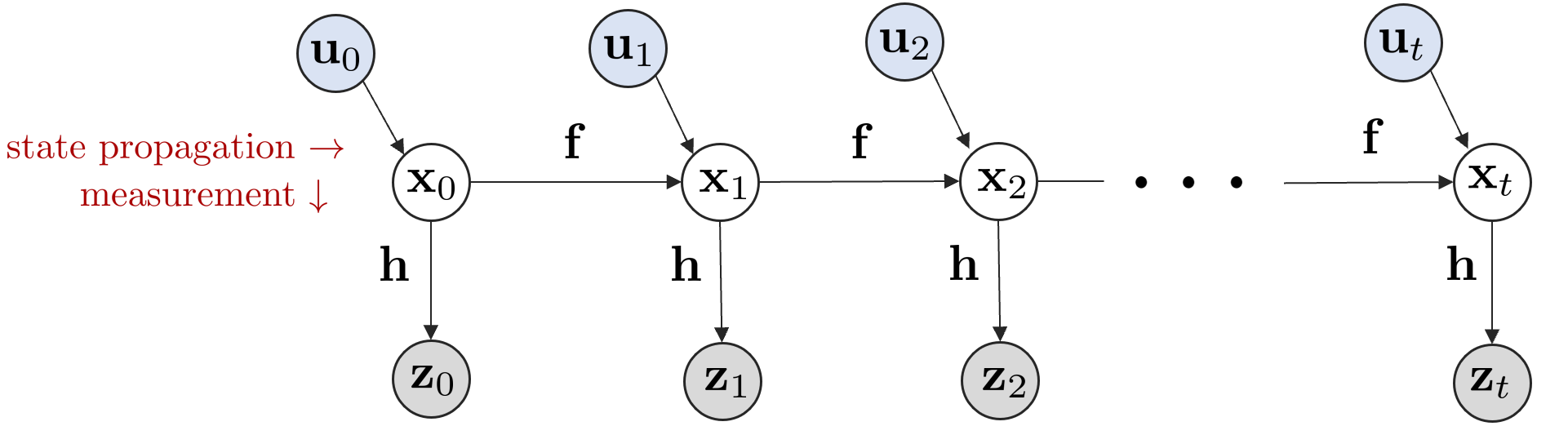}
\end{figure}

A dynamic system typically involves a prediction phase and an update (or correction) phase. The prediction phase refers to predicting the next state $\mathbf{x}_{t+1}$ from all observed values $\mathbf{z}_{0:t}$ and the current state $\mathbf{x}_{t}$.
\begin{figure}[h!]
	\centering
	\includegraphics[width=12cm]{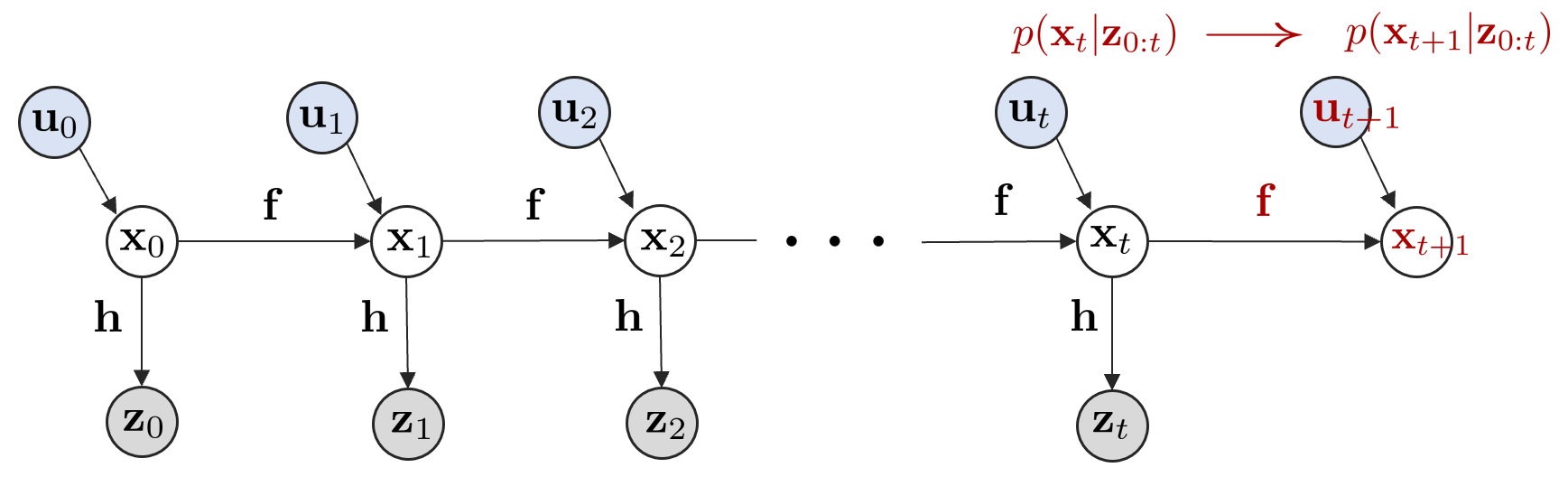}
\end{figure}

The update phase refers to obtaining new observations $\mathbf{z}_{t+1}$ from the next state $\mathbf{x}_{t+1}$.
\begin{figure}[h!]
	\centering
	\includegraphics[width=12cm]{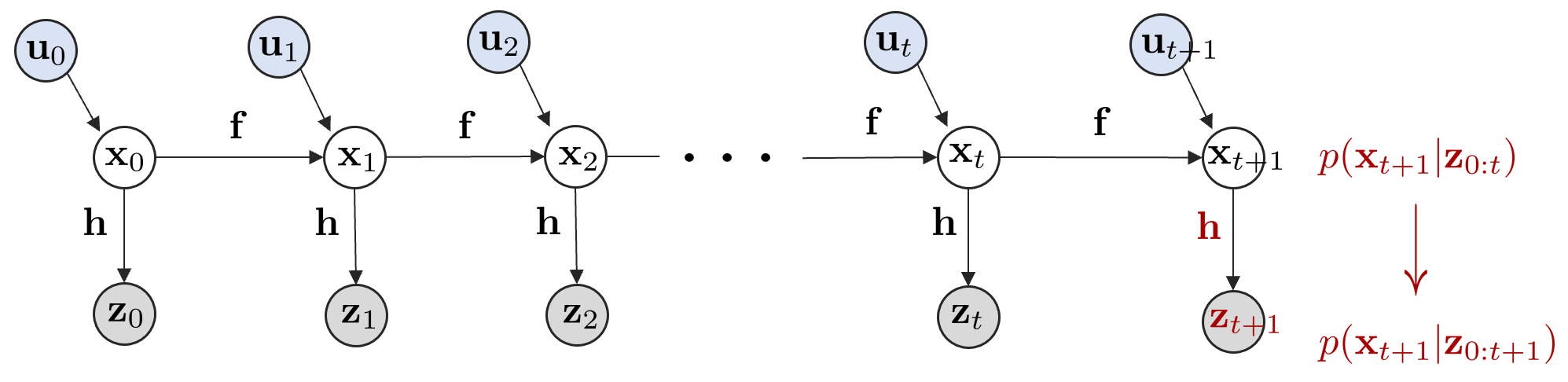}
\end{figure}

\section{Recursive Bayes Filter}
Given control inputs and observations, the degree of reliability of the current state $\mathbf{x}_{t}$, or the Belief $\text{bel}(\mathbf{x}_{t})$ about $\mathbf{x}_{t}$, is defined as follows:
\begin{equation}
	\begin{aligned}
		\text{bel}(\mathbf{x}_{t}) = p(\mathbf{x}_{t} \ | \ \mathbf{z}_{1:t}, \mathbf{u}_{1:t})
	\end{aligned}
\end{equation}
- $\mathbf{x}_{t}$ : State variable at time $t$\\
- $\mathbf{z}_{1:t} = \{\mathbf{z}_{1}, \cdots, \mathbf{z}_{t} \}$ : Observations from time 1 to $t$\\
- $\mathbf{u}_{1:t} = \{\mathbf{u}_{1}, \cdots, \mathbf{u}_{t} \}$ : Control inputs from time 1 to $t$\\
- $\text{bel}(\mathbf{x}_{t})$ : Belief of $\mathbf{x}_{t}$, which represents the probability (degree of belief) that the robot is at $\mathbf{x}_{t}$ based on sensor observations $\mathbf{z}_{1:t}$ and control inputs $\mathbf{u}_{1:t}$ from the start time to $t$ seconds.\\

$\text{bel}(\cdot)$ is expressed and developed according to the Bayesian rule, so it is also called the Bayes filter. Using the Markov assumption and the Bayesian rule, a recursive filter can be induced, and this is called the recursive Bayes filter.
\begin{equation}
	\begin{aligned}
		& \text{bel}(\mathbf{x}_{t}) = \eta \cdot p( \mathbf{z}_{t} \ | \  \mathbf{x}_{t})\overline{\text{bel}}(\mathbf{x}_{t}) \\
		& \overline{\text{bel}}(\mathbf{x}_{t}) = \int p(\mathbf{x}_{t} \ | \ \mathbf{x}_{t-1}, \mathbf{u}_{t})\text{bel}(\mathbf{x}_{t-1}) d \mathbf{x}_{t-1}\\
	\end{aligned}
\end{equation}
- $\eta = 1/p(\mathbf{z}_{t}|\mathbf{z}_{1:t-1}, \mathbf{u}_{1:t})$ : A normalization factor to maintain the definition of the probability distribution by normalizing its width to 1\\
- $p( \mathbf{z}_{t} \ | \ \mathbf{x}_{t})$ : Observation model\\
- $\int p(\mathbf{x}_{t} \ | \ \mathbf{x}_{t-1},\mathbf{u}_{t}) d \mathbf{x}_{t-1}$ : Motion model\\

The Recursive Bayes filter is thus called a recursive filter because it calculates the current step's $\text{bel}(\mathbf{x}_{t})$ from the previous step's $\text{bel}(\mathbf{x}_{t-1})$.

\subsection{Derivation of Recursive Bayes Filter}
The formulation of the Recursive Bayes Filter is derived as follows:
\begin{equation}
	\begin{aligned}
		\text{bel}(\mathbf{x}_{t}) & = p( \mathbf{x}_{t}  | \mathbf{z}_{1:t}, \mathbf{u}_{1:t}) \\
		& = \eta \cdot p(\mathbf{z}_{t} |  \mathbf{x}_{t}, \mathbf{z}_{1:t-1}, \mathbf{u}_{1:t}) \cdot p( \mathbf{x}_{t} | \mathbf{z}_{1:t-1}, \mathbf{u}_{1:t}) \\
		& = \eta \cdot p(\mathbf{z}_{t} |  \mathbf{x}_{t}) \cdot \int_{\mathbf{x}_{t-1}} p( \mathbf{x}_{t} | \mathbf{x}_{t-1},\mathbf{z}_{1:t-1}, \mathbf{u}_{1:t}) \cdot p(\mathbf{x}_{t-1} | \mathbf{z}_{1:t-1}, \mathbf{u}_{1:t}) d\mathbf{x}_{t-1}\\
		& = \eta \cdot p(\mathbf{z}_{t} |  \mathbf{x}_{t}) \cdot \int_{\mathbf{x}_{t-1}} p( \mathbf{x}_{t} | \mathbf{x}_{t-1},\mathbf{u}_{t}) \cdot p(\mathbf{x}_{t-1} | \mathbf{z}_{1:t-1}, \mathbf{u}_{1:t-1}) d\mathbf{x}_{t-1} \\
		& = \eta \cdot p(\mathbf{z}_{t} |  \mathbf{x}_{t}) \cdot \int_{\mathbf{x}_{t-1}} p( \mathbf{x}_{t} | \mathbf{x}_{t-1},\mathbf{u}_{t}) \text{bel}(\mathbf{x}_{t-1}) d \mathbf{x}_{t-1} \\
		& = \eta \cdot p(\mathbf{z}_{t} |  \mathbf{x}_{t}) \cdot \overline{\text{bel}}(\mathbf{x}_{t-1}) \\
	\end{aligned} 
\end{equation}
Steps applied include Bayesian rule, Markov Assumption, and Marginalization to derive the recursive filter process.
	
\textbfazure{Step 1:}
\begin{equation}
	\begin{aligned}
		\text{bel}(\mathbf{x}_{t}) & = p({\color{Cyan} \mathbf{x}_{t} } | {\color{Cyan} \mathbf{z}_{1:t}}, \mathbf{u}_{1:t}) \\
		& = \eta \cdot p({\color{Cyan} \mathbf{z}_{t}} | {\color{Cyan} \mathbf{x}_{t}}, \mathbf{z}_{1:t-1}, \mathbf{u}_{1:t}) \cdot p( {\color{Cyan} \mathbf{x}_{t}} | \mathbf{z}_{1:t-1}, \mathbf{u}_{1:t})
	\end{aligned}
\end{equation}
Apply the Bayesian rule: $p(x|y) = \frac{p(y|x)p(x)}{p(y)}$

\textbfazure{Step 2:}
\begin{equation}
	\begin{aligned}
		\text{bel}(\mathbf{x}_{t}) & = p( \mathbf{x}_{t}  | \mathbf{z}_{1:t}, \mathbf{u}_{1:t}) \\
		& = \eta \cdot p(\mathbf{z}_{t} |  {\color{Cyan} \mathbf{x}_{t}, \mathbf{z}_{1:t-1}, \mathbf{u}_{1:t} } ) \cdot p( \mathbf{x}_{t} | \mathbf{z}_{1:t-1}, \mathbf{u}_{1:t}) \\
		& = \eta \cdot p(\mathbf{z}_{t} |  {\color{Cyan} \mathbf{x}_{t} } ) \cdot p( \mathbf{x}_{t} | \mathbf{z}_{1:t-1}, \mathbf{u}_{1:t}) \\
	\end{aligned} 
\end{equation} 

Apply the Markov Assumption that the current state depends only on the immediately previous state.

\textbfazure{Step 3:}
\begin{equation}
	\begin{aligned}
		\text{bel}(\mathbf{x}_{t}) & = p( \mathbf{x}_{t}  | \mathbf{z}_{1:t}, \mathbf{u}_{1:t}) \\
		& = \eta \cdot p(\mathbf{z}_{t} |  \mathbf{x}_{t}, \mathbf{z}_{1:t-1}, \mathbf{u}_{1:t}) \cdot p( \mathbf{x}_{t} | \mathbf{z}_{1:t-1}, \mathbf{u}_{1:t}) \\
		& = \eta \cdot p(\mathbf{z}_{t} |  \mathbf{x}_{t}) \cdot {\color{Cyan} p( \mathbf{x}_{t} | \mathbf{z}_{1:t-1}, \mathbf{u}_{1:t}) } \\
		& = \eta \cdot p(\mathbf{z}_{t} |  \mathbf{x}_{t}) \cdot {\color{Cyan}  \int_{\mathbf{x}_{t-1}} p( \mathbf{x}_{t} | \mathbf{x}_{t-1},\mathbf{z}_{1:t-1}, \mathbf{u}_{1:t}) \cdot p(\mathbf{x}_{t-1} | \mathbf{z}_{1:t-1}, \mathbf{u}_{1:t}) d\mathbf{x}_{t-1} } \\
	\end{aligned} 
\end{equation}

Apply the Law of Total Probability or Marginalization: $p(x) = \int_{y} p(x|y) \cdot p(y) dy$

\textbfazure{Step 4:}
\begin{equation}
	\begin{aligned}
		\text{bel}(\mathbf{x}_{t}) & = p( \mathbf{x}_{t}  | \mathbf{z}_{1:t}, \mathbf{u}_{1:t}) \\
		& = \eta \cdot p(\mathbf{z}_{t} |  \mathbf{x}_{t}, \mathbf{z}_{1:t-1}, \mathbf{u}_{1:t}) \cdot p( \mathbf{x}_{t} | \mathbf{z}_{1:t-1}, \mathbf{u}_{1:t}) \\
		& = \eta \cdot p(\mathbf{z}_{t} |  \mathbf{x}_{t}) \cdot p( \mathbf{x}_{t} | \mathbf{z}_{1:t-1}, \mathbf{u}_{1:t}) \\
		& = \eta \cdot p(\mathbf{z}_{t} |  \mathbf{x}_{t}) \cdot \int_{\mathbf{x}_{t-1}} p( \mathbf{x}_{t} | \mathbf{x}_{t-1},{\color{Cyan} \mathbf{z}_{1:t-1}, \mathbf{u}_{1:t} } ) \cdot p(\mathbf{x}_{t-1} | \mathbf{z}_{1:t-1}, \mathbf{u}_{1:t}) d\mathbf{x}_{t-1}\\
		& = \eta \cdot p(\mathbf{z}_{t} |  \mathbf{x}_{t}) \cdot \int_{\mathbf{x}_{t-1}} p( \mathbf{x}_{t} | \mathbf{x}_{t-1}, {\color{Cyan} \mathbf{u}_{t} } ) \cdot p(\mathbf{x}_{t-1} | \mathbf{z}_{1:t-1}, \mathbf{u}_{1:t}) d\mathbf{x}_{t-1} \\
	\end{aligned} 
\end{equation}
Apply the Markov Assumption that the current state depends only on the immediately previous state.

\textbfazure{Step 5:}
\begin{equation}
	\begin{aligned}
		\text{bel}(\mathbf{x}_{t}) & = p( \mathbf{x}_{t}  | \mathbf{z}_{1:t}, \mathbf{u}_{1:t}) \\
		& = \eta \cdot p(\mathbf{z}_{t} |  \mathbf{x}_{t}, \mathbf{z}_{1:t-1}, \mathbf{u}_{1:t}) \cdot p( \mathbf{x}_{t} | \mathbf{z}_{1:t-1}, \mathbf{u}_{1:t}) \\
		& = \eta \cdot p(\mathbf{z}_{t} |  \mathbf{x}_{t}) \cdot p( \mathbf{x}_{t} | \mathbf{z}_{1:t-1}, \mathbf{u}_{1:t}) \\
		& = \eta \cdot p(\mathbf{z}_{t} |  \mathbf{x}_{t}) \cdot \int_{\mathbf{x}_{t-1}} p( \mathbf{x}_{t} | \mathbf{x}_{t-1},\mathbf{z}_{1:t-1}, \mathbf{u}_{1:t}) \cdot p(\mathbf{x}_{t-1} | \mathbf{z}_{1:t-1}, \mathbf{u}_{1:t}) d\mathbf{x}_{t-1}\\
		& = \eta \cdot p(\mathbf{z}_{t} |  \mathbf{x}_{t}) \cdot \int_{\mathbf{x}_{t-1}} p( \mathbf{x}_{t} | \mathbf{x}_{t-1},\mathbf{u}_{t}) \cdot p(\mathbf{x}_{t-1} | \mathbf{z}_{1:t-1}, {\color{Cyan}  \mathbf{u}_{1:t} } ) d\mathbf{x}_{t-1} \\
		& = \eta \cdot p(\mathbf{z}_{t} |  \mathbf{x}_{t}) \cdot \int_{\mathbf{x}_{t-1}} p( \mathbf{x}_{t} | \mathbf{x}_{t-1},\mathbf{u}_{t}) \cdot p(\mathbf{x}_{t-1} | \mathbf{z}_{1:t-1}, {\color{Cyan} \mathbf{u}_{1:t-1} } ) d\mathbf{x}_{t-1} \\
	\end{aligned} 
\end{equation}
Apply the Markov Assumption that the current state depends only on the immediately previous state.

\textbfazure{Step 6:}
\begin{equation}
	\begin{aligned}
		\text{bel}(\mathbf{x}_{t}) & = p( \mathbf{x}_{t}  | \mathbf{z}_{1:t}, \mathbf{u}_{1:t}) \\
		& = \eta \cdot p(\mathbf{z}_{t} |  \mathbf{x}_{t}, \mathbf{z}_{1:t-1}, \mathbf{u}_{1:t}) \cdot p( \mathbf{x}_{t} | \mathbf{z}_{1:t-1}, \mathbf{u}_{1:t}) \\
		& = \eta \cdot p(\mathbf{z}_{t} |  \mathbf{x}_{t}) \cdot p( \mathbf{x}_{t} | \mathbf{z}_{1:t-1}, \mathbf{u}_{1:t}) \\
		& = \eta \cdot p(\mathbf{z}_{t} |  \mathbf{x}_{t}) \cdot \int_{\mathbf{x}_{t-1}} p( \mathbf{x}_{t} | \mathbf{x}_{t-1},\mathbf{z}_{1:t-1}, \mathbf{u}_{1:t}) \cdot p(\mathbf{x}_{t-1} | \mathbf{z}_{1:t-1}, \mathbf{u}_{1:t}) d\mathbf{x}_{t-1}\\
		& = \eta \cdot p(\mathbf{z}_{t} |  \mathbf{x}_{t}) \cdot \int_{\mathbf{x}_{t-1}} p( \mathbf{x}_{t} | \mathbf{x}_{t-1},\mathbf{u}_{t}) \cdot p(\mathbf{x}_{t-1} | \mathbf{z}_{1:t-1}, \mathbf{u}_{1:t}) d\mathbf{x}_{t-1} \\
		& = \eta \cdot p(\mathbf{z}_{t} |  \mathbf{x}_{t}) \cdot \int_{\mathbf{x}_{t-1}} p( \mathbf{x}_{t} | \mathbf{x}_{t-1},\mathbf{u}_{t}) \cdot {\color{Cyan} p(\mathbf{x}_{t-1} | \mathbf{z}_{1:t-1}, \mathbf{u}_{1:t-1})} d\mathbf{x}_{t-1} \\
		& = \eta \cdot p(\mathbf{z}_{t} |  \mathbf{x}_{t}) \cdot \int_{\mathbf{x}_{t-1}} p( \mathbf{x}_{t} | \mathbf{x}_{t-1},\mathbf{u}_{t}) \cdot {\color{Cyan} \text{bel}(\mathbf{x}_{t-1}) } d\mathbf{x}_{t-1} \\
	\end{aligned} 
\end{equation}

$p(\mathbf{x}_{t-1} | \mathbf{z}_{1:t-1}, \mathbf{u}_{1:t-1})$ is substituted by $\text{bel}(\mathbf{x}_{t-1})$ as they are equivalent.

\textbfazure{Step 7:}
\begin{equation}
	\begin{aligned}
		\text{bel}(\mathbf{x}_{t}) & = p( \mathbf{x}_{t}  | \mathbf{z}_{1:t}, \mathbf{u}_{1:t}) \\
		& = \eta \cdot p(\mathbf{z}_{t} |  \mathbf{x}_{t}, \mathbf{z}_{1:t-1}, \mathbf{u}_{1:t}) \cdot p( \mathbf{x}_{t} | \mathbf{z}_{1:t-1}, \mathbf{u}_{1:t}) \\
		& = \eta \cdot p(\mathbf{z}_{t} |  \mathbf{x}_{t}) \cdot p( \mathbf{x}_{t} | \mathbf{z}_{1:t-1}, \mathbf{u}_{1:t}) \\
		& = \eta \cdot p(\mathbf{z}_{t} |  \mathbf{x}_{t}) \cdot \int_{\mathbf{x}_{t-1}} p( \mathbf{x}_{t} | \mathbf{x}_{t-1},\mathbf{z}_{1:t-1}, \mathbf{u}_{1:t}) \cdot p(\mathbf{x}_{t-1} | \mathbf{z}_{1:t-1}, \mathbf{u}_{1:t}) d\mathbf{x}_{t-1}\\
		& = \eta \cdot p(\mathbf{z}_{t} |  \mathbf{x}_{t}) \cdot \int_{\mathbf{x}_{t-1}} p( \mathbf{x}_{t} | \mathbf{x}_{t-1},\mathbf{u}_{t}) \cdot p(\mathbf{x}_{t-1} | \mathbf{z}_{1:t-1}, \mathbf{u}_{1:t}) d\mathbf{x}_{t-1} \\
		& = \eta \cdot p(\mathbf{z}_{t} |  \mathbf{x}_{t}) \cdot \int_{\mathbf{x}_{t-1}} p( \mathbf{x}_{t} | \mathbf{x}_{t-1},\mathbf{u}_{t}) \cdot p(\mathbf{x}_{t-1} | \mathbf{z}_{1:t-1}, \mathbf{u}_{1:t-1}) d\mathbf{x}_{t-1} \\
		& = \eta \cdot p(\mathbf{z}_{t} |  \mathbf{x}_{t}) \cdot {\color{Cyan} \int_{\mathbf{x}_{t-1}} p( \mathbf{x}_{t} | \mathbf{x}_{t-1},\mathbf{u}_{t}) \cdot \text{bel}(\mathbf{x}_{t-1}) d \mathbf{x}_{t-1} } \\
		& = \eta \cdot p(\mathbf{z}_{t} |  \mathbf{x}_{t}) \cdot  {\color{Cyan}  \overline{\text{bel}}(\mathbf{x}_{t}) }\\
	\end{aligned} 
\end{equation}

The integral $\int_{\mathbf{x}_{t-1}} p( \mathbf{x}_{t} | \mathbf{x}_{t-1},\mathbf{u}_{t}) \cdot \text{bel}(\mathbf{x}_{t-1}) d \mathbf{x}_{t-1}$ is replaced by $\overline{\text{bel}}(\mathbf{x}_{t})$.

\subsection{Gaussian Belief Case}
If $\text{bel}(\mathbf{x}_{t})$ follows a Gaussian distribution, this is specifically referred to as the Kalman filter.
\begin{equation}
	\begin{aligned}
		& \overline{\text{bel}}(\mathbf{x}_{t}) \sim \mathcal{N}(\hat{\mathbf{x}}_{t|t-1}, \mathbf{P}_{t|t-1}) \quad \text{(Kalman Filter Prediction)} \\ 
		& \text{bel}(\mathbf{x}_{t}) \sim \mathcal{N}(\hat{\mathbf{x}}_{t|t}, \mathbf{P}_{t|t}) \quad \text{(Kalman Filter Correction)}
	\end{aligned}
\end{equation}

Mean and covariance are also denoted as $(\hat{\mathbf{x}}, \mathbf{P})$ or $(\hat{\boldsymbol{\mu}}, \boldsymbol{\Sigma})$, representing the same concept with different notations.

	\section{Kalman Filter (KF)}
	\textbf{NOMENCLATURE of Kalman Filter}
	\begin{itemize}
		\item Scalars are denoted by lowercase letters, e.g., a
		\item Vectors are denoted by bold lowercase letters, e.g., $\mathbf{a}$
		\item Matrices are denoted by bold uppercase letters, e.g., $\mathbf{R}$
		\item prediction: $\overline{\text{bel}}(\mathbf{x}_{t}) \sim \mathcal{N}(\hat{\mathbf{x}}_{t|t-1}, {\mathbf{P}}_{t|t-1})$
		\begin{itemize}
			\item $\hat{\mathbf{x}}_{t|t-1}$ : Mean at step $t$ given the correction value at step $t-1$. Some literature also denotes this as $\mathbf{x}^{-}_{t}$.
			\item $\hat{\mathbf{P}}_{t|t-1}$ : Covariance at step $t$ given the correction value at step $t-1$. Some literature also denotes this as $\mathbf{P}^{-}_{t}$.
		\end{itemize}
		\item correction: $\text{bel}(\mathbf{x}_{t}) \sim \mathcal{N}(\hat{\mathbf{x}}_{t|t}, \mathbf{P}_{t|t})$
		\begin{itemize}
			\item $\hat{\mathbf{x}}_{t|t}$ : Mean at step $t$ given the prediction value at step $t$. Some literature also denotes this as $\mathbf{x}^{+}_{t}$.
			\item $\hat{\mathbf{P}}_{t|t}$ : Covariance at step $t$ given the prediction value at step $t$. Some literature also denotes this as $\mathbf{P}^{+}_{t}$.
		\end{itemize}
	\end{itemize}
	
	~\\
	\begin{figure}[h!]
		\centering
		\includegraphics[width=12cm]{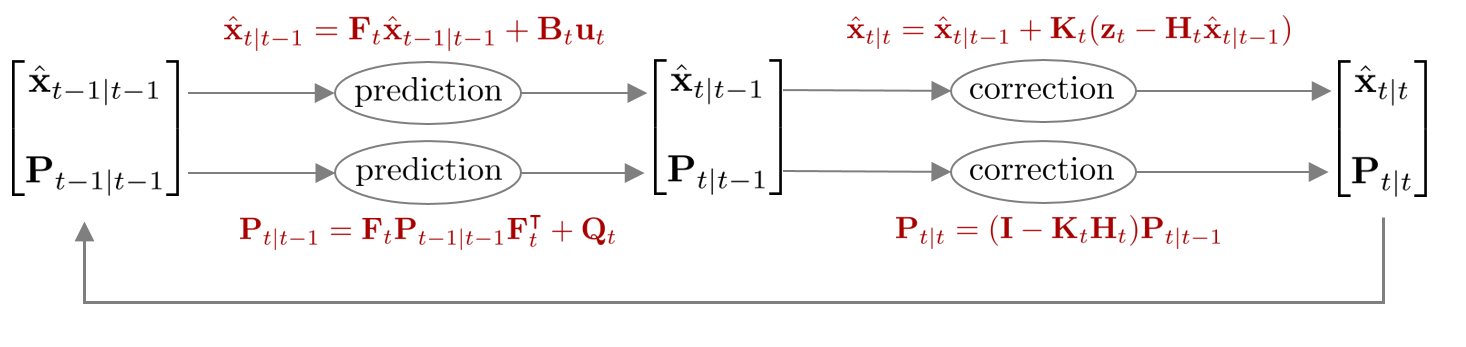}
	\end{figure}
	
	At time $t$, the robot's position is denoted by $\mathbf{x}_{t}$, the measurements from the robot's sensor by $\mathbf{z}_{t}$, and the robot's control input by $\mathbf{u}_{t}$. Using these, we can define the motion model and observation model. The models are constrained by the requirement that they must be linear.
	
	\begin{equation}
		\begin{aligned}
			& \text{Motion Model: } & \mathbf{x}_{t} = \mathbf{F}_{t}\mathbf{x}_{t-1} + \mathbf{B}_{t}\mathbf{u}_{t} + \mathbf{w}_{t} \\
			& \text{Observation Model: } & \mathbf{z}_{t} = \mathbf{H}_{t}\mathbf{x}_{t} + \mathbf{v}_{t} 
		\end{aligned}
	\end{equation}
	- $\mathbf{x}_{t}$: Model state variable\\
	- $\mathbf{u}_{t}$: Model input\\
	- $\mathbf{z}_{t}$: Model measurement\\
	- $\mathbf{F}_{t}$: Model state transition matrix\\
	- $\mathbf{B}_{t}$: Matrix that transforms input $\mathbf{u}_{t}$ to the state variable when given\\
	- $\mathbf{H}_{t}$: Model observation matrix\\
	- $\mathbf{w}_{t} \sim \mathcal{N}(0, \mathbf{Q}_{t})$: Motion model noise. $\mathbf{Q}_{t}$ denotes the covariance matrix of $\mathbf{w}_{t}$.\\
	- $\mathbf{v}_{t} \sim \mathcal{N}(0, \mathbf{R}_{t})$: Observation model noise. $\mathbf{R}_{t}$ denotes the covariance matrix of $\mathbf{v}_{t}$.\\
	
	Assuming all random variables follow Gaussian distributions, $p(\mathbf{x}_{t} \ | \ \mathbf{x}_{t-1}, \mathbf{u}_{t}), p(\mathbf{z}_{t} \ | \ \mathbf{x}_{t})$ can be represented as follows.
	\begin{equation}  \label{eq:kf1}
		\begin{aligned}
			& p(\mathbf{x}_{t} \ | \ \mathbf{x}_{t-1}, \mathbf{u}_{t}) && \sim \mathcal{N}(\mathbf{F}_{t}\mathbf{x}_{t-1} + \mathbf{B}_{t}\mathbf{u}_{t}, \mathbf{Q}_{t}) \\
			& && = \frac{1}{ \sqrt{\text{det}(2\pi \mathbf{Q}_t)}}\exp\bigg( -\frac{1}{2}(\mathbf{x}_{t} -\mathbf{F}_{t}\mathbf{x}_{t-1}-\mathbf{B}_{t}\mathbf{u}_t)^{\intercal}\mathbf{Q}_t^{-1}(\mathbf{x}_{t}-\mathbf{F}_{t}\mathbf{x}_{t-1}-\mathbf{B}_{t}\mathbf{u}_{t}) \bigg)\\ 
		\end{aligned}
	\end{equation}
	\begin{equation} \label{eq:kf2}
		\begin{aligned}
			& p(\mathbf{z}_{t} \ | \ \mathbf{x}_{t})  && \sim \mathcal{N}(\mathbf{H}_{t}\mathbf{x}_{t}, \mathbf{R}_{t}) \\
			& && = \frac{1}{ \sqrt{\text{det}(2\pi \mathbf{R}_t)}}\exp\bigg( -\frac{1}{2}(\mathbf{z}_{t}-\mathbf{H}_{t}\mathbf{x}_{t})^{\intercal}\mathbf{R}_{t}^{-1}(\mathbf{z}_{t}-\mathbf{H}_{t}\mathbf{x}_{t}) \bigg)
		\end{aligned}
	\end{equation}
	
	Next, the $\overline{\text{bel}}(\mathbf{x}_{t}), \text{bel}(\mathbf{x}_{t})$ that need to be computed through the Kalman filter can be represented as follows.
	\begin{equation} \label{eq:kf3}
		\begin{aligned}
			& \overline{\text{bel}}(\mathbf{x}_{t}) = \int p(\mathbf{x}_{t} \ | \ \mathbf{x}_{t-1}, \mathbf{u}_{t})\text{bel}(\mathbf{x}_{t-1}) d \mathbf{x}_{t-1} \sim \mathcal{N}(\hat{\mathbf{x}}_{t|t-1}, \mathbf{P}_{t|t-1}) \\
			& \text{bel}(\mathbf{x}_{t}) = \eta \cdot p( \mathbf{z}_{t} \ | \  \mathbf{x}_{t})\overline{\text{bel}}(\mathbf{x}_{t}) \sim \mathcal{N}(\hat{\mathbf{x}}_{t|t}, \mathbf{P}_{t|t})
		\end{aligned}
	\end{equation}
	
	As seen in (\ref{eq:kf3}), the Kalman filter operates by first computing the predicted value $\overline{\text{bel}}(\mathbf{x}_{t})$ using the previous step's value and motion model in the prediction phase, then obtaining the corrected value $\text{bel}(\mathbf{x}_{t})$ using the measurement and observation model in the correction phase. \textbfazure{By substituting (\ref{eq:kf1}) and (\ref{eq:kf2}) into (\ref{eq:kf3}), we can derive the means and covariances for the prediction and correction steps $(\hat{\mathbf{x}}_{t|t-1}, \mathbf{P}_{t|t-1}), (\hat{\mathbf{x}}_{t|t}, \mathbf{P}_{t|t})$ respectively.} For a detailed derivation process, refer to Section \ref{sec:derivkf}.
	
	The initial value $\text{bel}(\mathbf{x}_{0})$ is given as follows.
	\begin{equation}
		\begin{aligned}
			\text{bel}(\mathbf{x}_{0}) \sim \mathcal{N}(\hat{\mathbf{x}}_{0}, \mathbf{P}_{0})
		\end{aligned}
	\end{equation}
	- $\hat{\mathbf{x}}_{0}$: Typically set to 0 \\
	- $\mathbf{P}_{0}$ : Typically set to a small value (<1e-2).
	
	\subsection{Prediction step}
	Prediction involves computing $\overline{\text{bel}}(\mathbf{x}_{t})$. Since $\overline{\text{bel}}(\mathbf{x}_{t})$ follows a Gaussian distribution, we can compute the mean $\hat{\mathbf{x}}_{t|t-1}$ and variance $\mathbf{P}_{t|t-1}$ as follows.
	\begin{equation}
		\boxed{ \begin{aligned}
				& \hat{\mathbf{x}}_{t|t-1} = \mathbf{F}_{t}\hat{\mathbf{x}}_{t-1|t-1} + \mathbf{B}_{t}\mathbf{u}_{t} \\
				& \mathbf{P}_{t|t-1} = \mathbf{F}_{t}\mathbf{P}_{t-1|t-1}\mathbf{F}_{t}^{\intercal} + \mathbf{Q}_{t}
		\end{aligned} }
	\end{equation}
	
	\subsection{Correction step}
	Correction involves computing $\text{bel}(\mathbf{x}_{t})$. Since $\text{bel}(\mathbf{x}_{t})$ also follows a Gaussian distribution, we can compute the mean $\hat{\mathbf{x}}_{t|t}$ and variance $\mathbf{P}_{t|t}$ as follows. Here, $\mathbf{K}_{t}$ represents the Kalman gain.
	\begin{equation}
		\boxed{ \begin{aligned}
				&     \mathbf{K}_{t} = \mathbf{P}_{t|t-1}\mathbf{H}_{t}^{\intercal}(\mathbf{H}_{t}\mathbf{P}_{t|t-1}\mathbf{H}_{t}^{\intercal} + \mathbf{R}_{t})^{-1} \\
				& \hat{\mathbf{x}}_{t|t}  = \hat{\mathbf{x}}_{t|t-1} + \mathbf{K}_{t}( \mathbf{z}_{t} - \mathbf{H}_{t}\hat{\mathbf{x}}_{t|t-1}) \\
				&     \mathbf{P}_{t|t}    = (\mathbf{I} - \mathbf{K}_{t}\mathbf{H}_{t})\mathbf{P}_{t|t-1}
		\end{aligned} }
	\end{equation}
	
	\subsection{1D Kalman filter}
	The Kalman filter described so far was for the case where the state variable is a vector (=$\mathbf{x}_t$). Next, let's look at the 1D Kalman filter where the state variable is a scalar (=$x_t$). The 1D Kalman filter is identical to the existing nD Kalman filter in all respects except that the equations are composed of scalar values instead of matrices. The Belief expressed in the 1D Kalman filter version is as follows.
	\begin{equation} 
		\begin{aligned} 
			& \overline{\text{bel}}(x_{t}) \sim \mathcal{N}(\bar{\mu}_t, \bar{\sigma}^2_t) \\
			& \text{bel}(x_{t}) \sim \mathcal{N}(\mu_{t}, \sigma^2_{t})
		\end{aligned} 
	\end{equation}
	- $\bar{\mu}_t$: mean of the prediction step \\
	- $\bar{\sigma}^2_t$: variance of the prediction step \\
	- ${\mu}_t$: mean of the correction step \\
	- ${\sigma}^2_t$: variance of the correction step \\
	
	The motion model and observation model are as follows.
	\begin{equation} 
		\begin{aligned} 
			& \text{Motion Model: } & x_{t} =  x_{t-1} +  u_{t} + \sigma^2_{\text{motion},t} \\
			& \text{Observation Model: } & z_{t} =  x_{t} + \sigma^2_{\text{obs},t} 
		\end{aligned} 
	\end{equation}
	- $u_{t}$: control input of the motion model \\
	- $\sigma^2_{\text{motion},t}$: noise of the motion model \\
	- $\sigma^2_{\text{obs},t}$: noise of the observation model \\
	
	The prediction step of the 1D Kalman filter is as follows.
	\begin{equation}
		\boxed{ \begin{aligned}
				& \bar{\mu}_{t} = \mu_{t-1} + u_{t}  \\
				& \bar{\sigma}^2_{t} = \sigma^2_{t-1} + \sigma_{\text{motion},t}^2
		\end{aligned} }
	\end{equation}
	
	Next, the correction step of the 1D Kalman filter is as follows.
	\begin{equation}
		\boxed{ \begin{aligned}
				&     K_{t} = \frac{\bar{\sigma}^2_{t}}{\bar{\sigma}^2_{t} + {\sigma}^2_{\text{obs},t}} \\
				& \mu_{t}  = \bar{\mu}_t + K_t(\mu_{\text{obs,t}} - \bar{\mu}_t) = \frac{\mu_{\text{obs,t}}\bar{\sigma}^2_{t} + \bar{\mu}_t \sigma^2_{\text{obs,t}}}{\bar{\sigma}^2_{t} + {\sigma}^2_{\text{obs},t}}\\
				&     \sigma^2_{t}    = (1-K_t) \bar{\sigma}^2_{t} = \frac{\bar{\sigma}^2_{t}{\sigma}^2_{\text{obs},t}}{\bar{\sigma}^2_{t} + {\sigma}^2_{\text{obs},t}}
		\end{aligned} }
	\end{equation}
	- $\mu_{\text{obs,t}}$: mean of the observation model \\
	- ${\sigma}^2_{\text{obs},t}$: noise of the observation model \\
	
	Examining the equations in detail, it can be seen that they are structurally identical to the vector version of the Kalman filter mentioned earlier. $\mathbf{F}_t, \mathbf{B}_t,  \mathbf{H}_t$ become $1$ and are omitted, corresponding to $\mathbf{Q}_t \leftrightarrow \sigma_{\text{motion},t}$ and $\mathbf{R}_t \leftrightarrow \sigma_{\text{obs},t}$. To make the comparison easier to see at a glance, a figure is attached.
	\begin{figure}[h!]
		\centering
		\includegraphics[width=15cm]{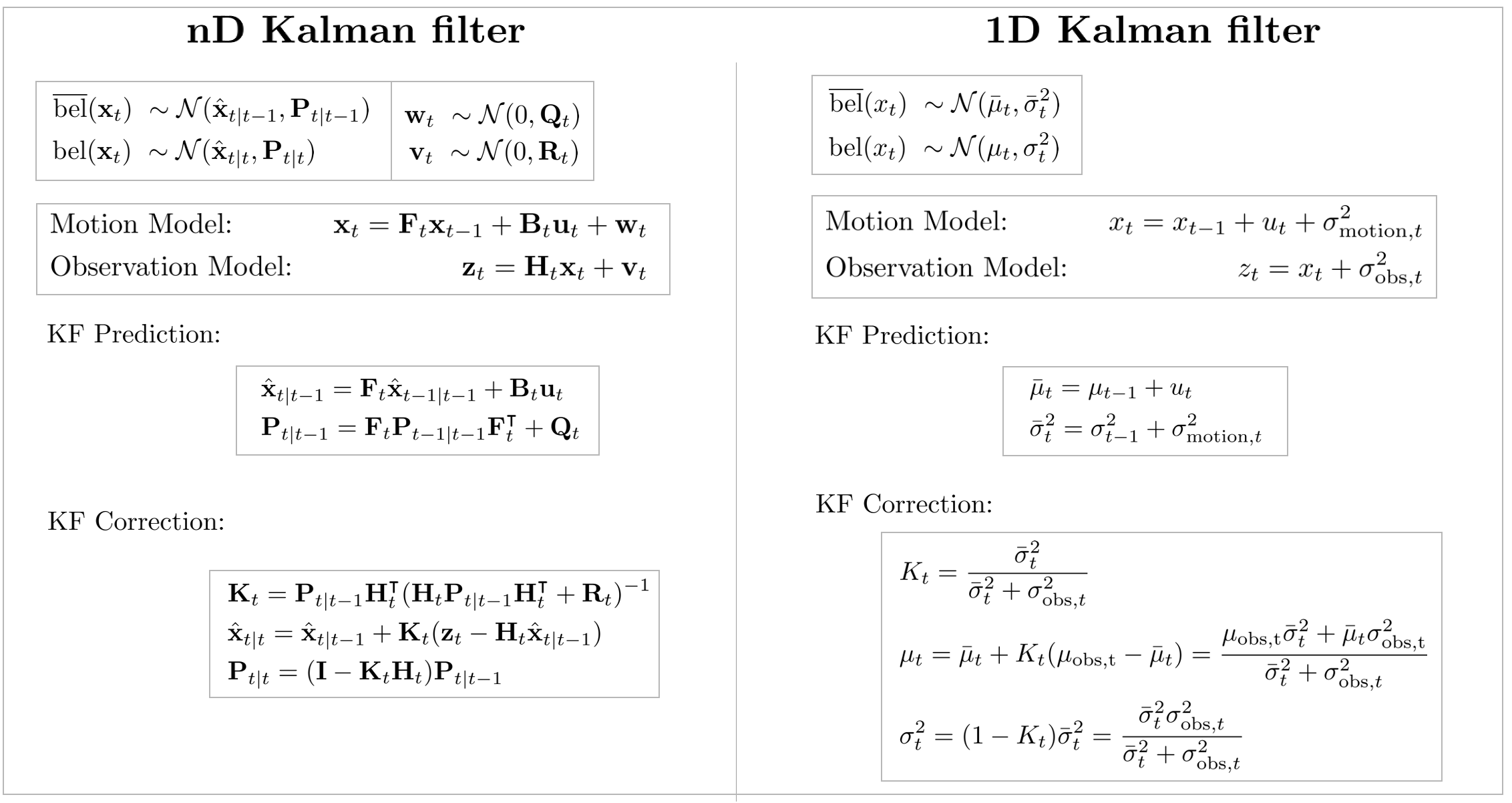}
	\end{figure}
	
	\subsection{Discussion}
	\subsubsection{Discussion about KF and posterior pdf}
	Let's assume a problem where the position $x_t$ of a robot is estimated by KF in a simple 1D space. The vertical axis represents the probability density function (pdf), the horizontal axis represents the 1D position $x$, and the state variable representing the robot's position at time $t$ is $x_t$.
	\begin{figure}[h!]
		\centering
		\includegraphics[width=16cm]{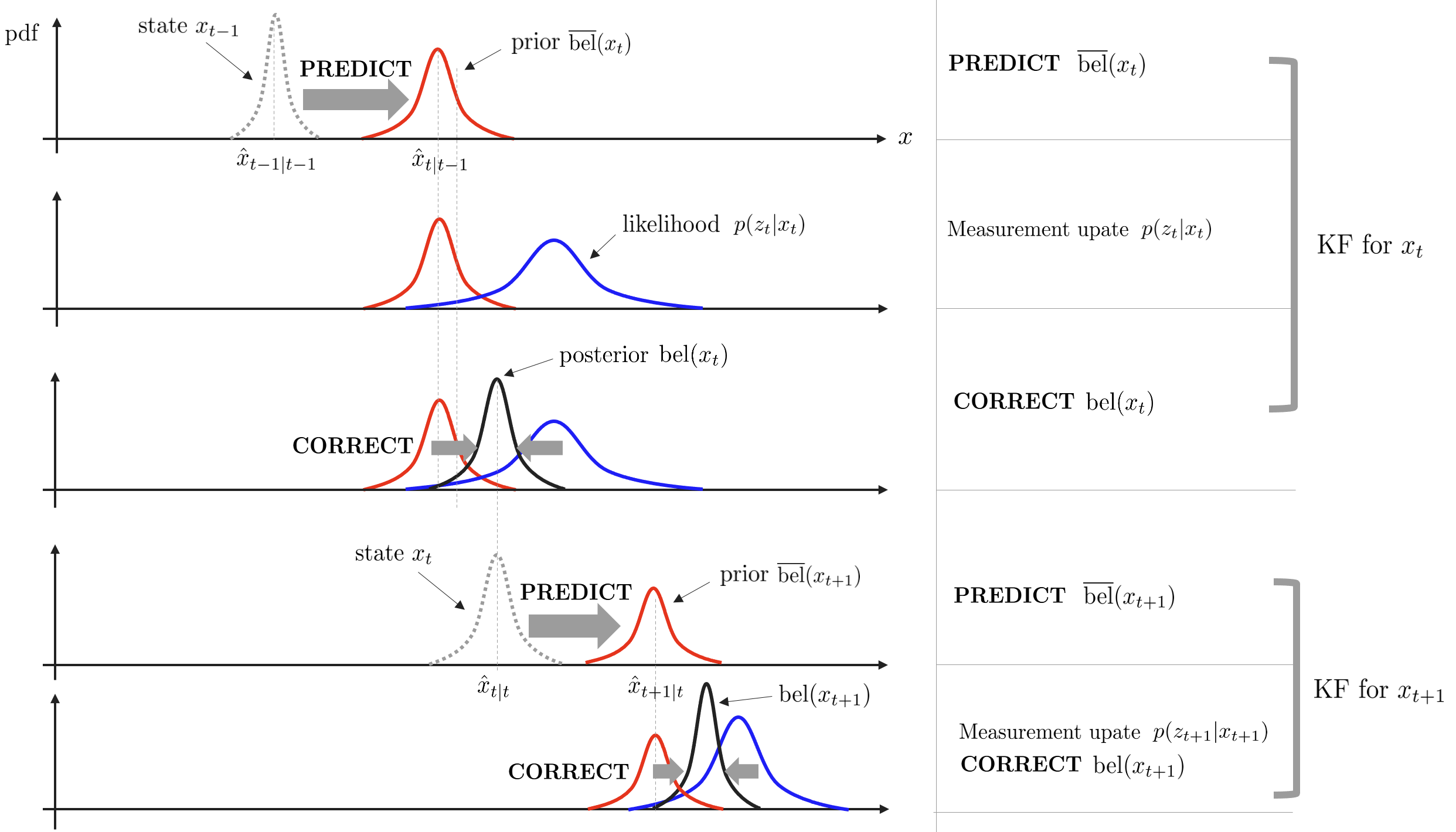}
	\end{figure}
	
	Looking at the figure above step by step:
	\begin{enumerate}
		\item The robot performs a prediction based on the motion model from the previous step's position $x_{t-1}$ to predict the prior $\overline{\text{bel}}(x_t)$. 
		\item Next, the robot measures the current position with its sensor to obtain the likelihood $p(z_t | x_t)$.
		\item Using the prior $\overline{\text{bel}}(x_t)$ and the likelihood $p(z_t | x_t)$, the correction step is performed to obtain the posterior $\text{bel}(x_t)$. 
		\item The process from 1 to 3 is repeated for the next step to obtain the posterior $\text{bel}(x_{t+1})$ for $x_{t+1}$. 
	\end{enumerate}
	
	Therefore, one step of the KF has the characteristics of a typical Bayesian filter, which \textbfazure{predicts the prior from the previous state variable (=prediction) and calculates the likelihood when the observation value is given, then finds the posterior pdf according to the Bayesian rule (=correction)}. Performing Bayesian filtering \textbfazure{recursively} on \textbfazure{a Belief that follows a Gaussian distribution} is the Kalman filter.
	
	\subsubsection{Discussion about Kalman gain}
	If we examine the correction step in detail, the mean $\hat{\mathbf{x}}_{t|t}$ is calculated with different weights depending on the Kalman gain $\mathbf{K}_t$.
	\begin{equation}
		\begin{aligned}
			\hat{\mathbf{x}}_{t|t}  = \underbrace{\hat{\mathbf{x}}_{t|t-1}}_{\text{prediction}} + \mathbf{K}_{t} \underbrace{ ( \mathbf{z}_{t} - \mathbf{H}_{t}\hat{\mathbf{x}}_{t|t-1}) }_{\text{innovation}}
		\end{aligned}
	\end{equation}
	- innovation: the difference between the observed value ($\mathbf{z}_t$) and the predicted value ($\mathbf{H}_t \hat{\mathbf{x}}_{t|t-1}$), that is, the error value \\
	
	If $\mathbf{K}_t \rightarrow \mathbf{0}$, it means that the sensor's observed value $\mathbf{z}_t$ is not trusted at all, and only the system's predicted value is reflected (the observed value $\mathbf{z}_t$ is not considered). On the other hand, if $\mathbf{K}_t \rightarrow \mathbf{1}$, it means that the sensor's observed value $\mathbf{z}_t$ is trusted 100
	
	Additionally, when the sensor noise ${\color{Red}\mathbf{R}}$ is large, $\mathbf{K}_t$ decreases, meaning the system model's predicted value is reflected more than the sensor's observed value. Conversely, when the system noise ${\color{Cyan} \mathbf{Q}}$ is large, $\mathbf{P}_{t|t-1}$ increases, leading to an increase in $\mathbf{K}_t$, which means the sensor's observed value is reflected more than the system model's predicted value.
	\begin{equation}
		\begin{aligned}
			& \mathbf{P}_{t|t-1} = \mathbf{F}_{t}\mathbf{P}_{t-1|t-1}\mathbf{F}_{t}^{\intercal} + {\color{Cyan} \mathbf{Q}_{t}} \\
			& \mathbf{K}_{t} = \mathbf{P}_{t|t-1}\mathbf{H}_{t}^{\intercal}(\mathbf{H}_{t}\mathbf{P}_{t|t-1}\mathbf{H}_{t}^{\intercal} + {\color{Red} \mathbf{R}_{t}})^{-1}  \\
		\end{aligned}
	\end{equation}
	- ${\color{Red}\mathbf{R}_{t}}(\uparrow) \Rightarrow (\ast + {\color{Red}\mathbf{R}_{t}})^{-1}(\downarrow)  \Rightarrow \mathbf{K}_{t}(\downarrow)$  \\
	- ${\color{Cyan}\mathbf{Q}_{t}}(\uparrow) \Rightarrow \mathbf{P}_{t|t-1}(\uparrow)  \Rightarrow \mathbf{K}_{t}(\uparrow)$  \\

	\subsection{Summary}
	The Kalman Filter can be expressed as a function as follows. 
	\begin{equation} 
		\boxed{ \begin{aligned} 
				& \text{KalmanFilter}(\hat{\mathbf{x}}_{t-1|t-1}, \mathbf{P}_{t-1|t-1}, \mathbf{u}_{t}, \mathbf{z}_{t}) \{ \\ 
				& \quad\quad \text{(Prediction Step)}\\ 
				& \quad\quad \hat{\mathbf{x}}_{t|t-1} = \mathbf{F}_{t}\hat{\mathbf{x}}_{t-1|t-1} + \mathbf{B}_{t}\mathbf{u}_{t} \\ 
				& \quad\quad \mathbf{P}_{t|t-1} = \mathbf{F}_{t}\mathbf{P}_{t-1|t-1}\mathbf{F}_{t}^{\intercal} + \mathbf{Q}_{t} \\ 
				& \\ 
				& \quad\quad \text{(Correction Step)} \\ 
				& \quad\quad \mathbf{K}_{t} = \mathbf{P}_{t|t-1}\mathbf{H}_{t}^{\intercal}(\mathbf{H}_{t}\mathbf{P}_{t|t-1}\mathbf{H}_{t}^{\intercal} + \mathbf{R}_{t})^{-1} \\ 
				& \quad\quad \hat{\mathbf{x}}_{t|t} = \hat{\mathbf{x}}_{t|t-1} + \mathbf{K}_{t}( \mathbf{z}_{t} - \mathbf{H}_{t}\hat{\mathbf{x}}_{t|t-1}) \\ 
				& \quad\quad \mathbf{P}_{t|t} = (\mathbf{I} - \mathbf{K}_{t}\mathbf{H}_{t})\mathbf{P}_{t|t-1} \\ 
				& \quad\quad \text{return} \ \ \hat{\mathbf{x}}_{t|t}, \mathbf{P}_{t|t} \\ 
				&   \} 
		\end{aligned} }
	\end{equation}

\section{Extended Kalman Filter (EKF)}

\begin{figure}[h!]
	\centering
	\includegraphics[width=12cm]{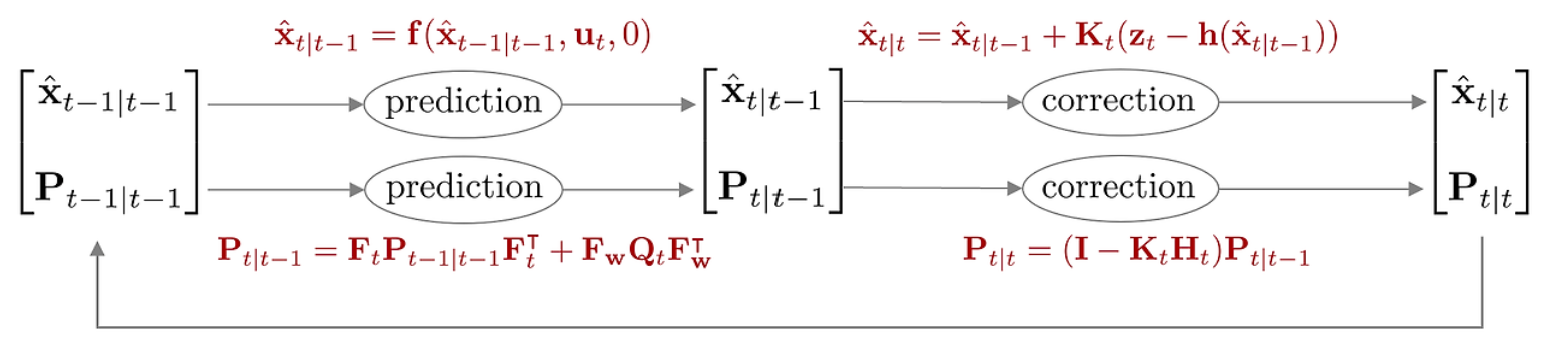}
\end{figure}
Kalman filters assume that both the motion model and the observation model are linear to estimate the state. However, since most phenomena in the real world are modeled non-linearly, applying the Kalman filter as previously defined would not work properly. The extended Kalman filter (EKF) was proposed to use the Kalman filter even in non-linear motion and observation models. \textbf{EKF uses the first-order Taylor approximation to approximate the non-linear model to a linear model before applying the Kalman filter.} The motion model and observation model of the EKF are as follows.
\begin{equation}
	\begin{aligned}
		& \text{Motion Model: } & \mathbf{x}_{t} = \mathbf{f}(\mathbf{x}_{t-1}, \mathbf{u}_{t} , \mathbf{w}_{t}) \\
		& \text{Observation Model: } &  \mathbf{z}_{t} = \mathbf{h}(\mathbf{x}_{t}, \mathbf{v}_{t})
	\end{aligned}
\end{equation}
- $\mathbf{x}_{t}$: state variable of the model\\
- $\mathbf{u}_{t}$: input of the model\\
- $\mathbf{z}_{t}$: measurement of the model\\
- $\mathbf{f}(\cdot)$: non-linear motion model function \\
- $\mathbf{h}(\cdot)$: non-linear observation model function\\
- $\mathbf{w}_{t} \sim \mathcal{N}(0, \mathbf{Q}_{t})$: noise in the motion model. $\mathbf{Q}_{t}$ represents the covariance matrix of $\mathbf{w}_{t}$\\
- $\mathbf{v}_{t} \sim \mathcal{N}(0, \mathbf{R}_{t})$: noise in the observation model. $\mathbf{R}_{t}$ represents the covariance matrix of $\mathbf{v}_{t}$\\

In the above formula, $\mathbf{f}(\cdot)$ means a non-linear motion model and $\mathbf{h}(\cdot)$ means a non-linear observation model. When the first-order Taylor approximation is performed on $\mathbf{f}(\cdot), \mathbf{h}(\cdot)$, it becomes as follows.
\begin{equation} \label{eq:20}
	\begin{aligned}
		& \mathbf{x}_{t} \approx \mathbf{f}(\hat{\mathbf{x}}_{t-1|t-1}, \mathbf{u}_{t} , 0 ) + \mathbf{F}_{t}(\mathbf{x}_{t-1} - \hat{\mathbf{x}}_{t-1|t-1}) + \mathbf{F}_{\mathbf{w}} \mathbf{w}_{t} \\
		&  \mathbf{z}_{t} \approx \mathbf{h}(\hat{\mathbf{x}}_{t|t-1}, 0) + \mathbf{H}_{t}(\mathbf{x}_{t} - \hat{\mathbf{x}}_{t|t-1}) + \mathbf{H}_{\mathbf{v}}\mathbf{v}_{t}
	\end{aligned}
\end{equation}

At this time, $\mathbf{F}_{t}$ represents the Jacobian matrix of the motion model calculated at $\hat{\mathbf{x}}_{t-1|t-1}$, and $\mathbf{H}_{t}$ represents the Jacobian matrix of the observation model calculated at $\hat{\mathbf{x}}_{t|t-1}$. And $\mathbf{F}_{\mathbf{w}}, \mathbf{H}_{\mathbf{v}}$ represent the Jacobian matrix of the noise when $\mathbf{w}_{t}=0, \mathbf{v}_{t}=0$ respectively. For more information on the Jacobian, refer to \href{https://alida.tistory.com/11#67.2.-jacobian-matrix}{this post}.
\begin{equation}
	\begin{aligned}
		 \mathbf{F}_{t} = \frac{\partial \mathbf{f}}{\partial \mathbf{x}_{t-1}}\bigg|_{\mathbf{x}_{t-1}=\hat{\mathbf{x}}_{t-1|t-1}} & \mathbf{F}_{\mathbf{w}} = \frac{\partial \mathbf{f}}{\partial \mathbf{w}_{t}} \bigg|_{\substack{  \mathbf{x}_{t-1}=\hat{\mathbf{x}}_{t-1|t-1} \\  \mathbf{w}_{t}=0}}   \\
	\end{aligned} 
\end{equation}
\begin{equation}
	\begin{aligned}
		 \mathbf{H}_{t} = \frac{\partial \mathbf{h}}{\partial \mathbf{x}_{t}}\bigg|_{\mathbf{x}_{t}=\hat{\mathbf{x}}_{t|t-1}}  & \mathbf{H}_{\mathbf{v}} = \frac{\partial \mathbf{h}}{\partial \mathbf{v}_{t}} \bigg|_{\substack{ \mathbf{x}_{t}=\hat{\mathbf{x}}_{t|t-1} \\   \mathbf{v}_{t} = 0}} \\
	\end{aligned}
\end{equation}

($\ref{eq:20})$ If the formula is expanded, it becomes as follows.
\begin{equation} \label{eq:ekf1}
	\boxed{  \begin{aligned}
			\mathbf{x}_{t} & = \mathbf{F}_{t}\mathbf{x}_{t-1} + 
			\mathbf{f}(\hat{\mathbf{x}}_{t-1|t-1}, \mathbf{u}_{t} , 0 ) - \mathbf{F}_{t} \hat{\mathbf{x}}_{t-1|t-1} + \mathbf{F}_{\mathbf{w}} \mathbf{w}_{t} \\
			& =  \mathbf{F}_{t}\mathbf{x}_{t-1} + \tilde{\mathbf{u}}_{t} + \tilde{\mathbf{w}}_{t} 
	\end{aligned} }
\end{equation}
- $\tilde{\mathbf{u}}_{t} = \mathbf{f}(\hat{\mathbf{x}}_{t-1|t-1}, \mathbf{u}_{t} , 0 ) - \mathbf{F}_{t} \hat{\mathbf{x}}_{t-1|t-1}$\\
- $\tilde{\mathbf{w}}_{t} = \mathbf{F}_{\mathbf{w}} \mathbf{w}_{t} \sim \mathcal{N}(0,  \mathbf{F}_{\mathbf{w}}\mathbf{Q}_{t} \mathbf{F}_{\mathbf{w}}^{\intercal})$\\

\begin{equation} \label{eq:ekf2}
	\boxed{ \begin{aligned}
			\mathbf{z}_{t} & = \mathbf{H}_{t}\mathbf{x}_{t} + \mathbf{h}(\hat{\mathbf{x}}_{t|t-1}, 0) - \mathbf{H}_{t}\hat{\mathbf{x}}_{t|t-1} + \mathbf{H}_{\mathbf{v}}\mathbf{v}_{t} \\ 
			& = \mathbf{H}_{t}\mathbf{x}_{t}  + \tilde{\mathbf{z}}_{t} + \tilde{\mathbf{v}}_{t}
	\end{aligned} }
\end{equation}
- $\tilde{\mathbf{z}}_{t} = \mathbf{h}(\hat{\mathbf{x}}_{t|t-1}, 0) - \mathbf{H}_{t}\hat{\mathbf{x}}_{t|t-1}$ \\
- $\tilde{\mathbf{v}}_{t} = \mathbf{H}_{\mathbf{v}}\mathbf{v}_{t} \sim \mathcal{N}(0,  \mathbf{H}_{\mathbf{v}}\mathbf{R}_{t} \mathbf{H}_{\mathbf{v}}^{\intercal})$ \\

	 Assuming all random variables follow a Gaussian distribution, $p(\mathbf{x}_{t} | \mathbf{x}_{t-1}, \mathbf{u}_{t}), p(\mathbf{z}_{t} | \mathbf{x}_{t})$ can be expressed as follows.
	\begin{equation}
		\begin{aligned}
			& p(\mathbf{x}_{t} | \mathbf{x}_{t-1}, \mathbf{u}_{t}) && \sim \mathcal{N}(\mathbf{F}_{t}\mathbf{x}_{t-1} + \tilde{\mathbf{u}}_{t}, \mathbf{F}_{\mathbf{w}}\mathbf{Q}_{t} \mathbf{F}_{\mathbf{w}}^{\intercal} ) \\
			& && = \frac{1}{ \sqrt{\text{det}(2\pi \mathbf{F}_{\mathbf{w}}\mathbf{Q}_{t} \mathbf{F}_{\mathbf{w}}^{\intercal} )}}\exp\bigg( -\frac{1}{2}(\mathbf{x}_{t} - \mathbf{F}_{t}\mathbf{x}_{t-1} - \tilde{\mathbf{u}}_{t})^{\intercal}( \mathbf{F}_{\mathbf{w}}\mathbf{Q}_{t} \mathbf{F}_{\mathbf{w}}^{\intercal} )^{-1}(\mathbf{x}_{t} - \mathbf{F}_{t}\mathbf{x}_{t-1} - \tilde{\mathbf{u}}_{t}) \bigg)\\ 
		\end{aligned}
	\end{equation}
	\begin{equation}
		\begin{aligned}
			& p(\mathbf{z}_{t} | \mathbf{x}_{t})  && \sim \mathcal{N}(\mathbf{H}_{t}\mathbf{x}_{t} + \tilde{\mathbf{z}}_{t},  \mathbf{H}_{\mathbf{v}}\mathbf{R}_{t} \mathbf{H}_{\mathbf{v}}^{\intercal} ) \\
			& && = \frac{1}{ \sqrt{\text{det}(2\pi \mathbf{H}_{\mathbf{v}}\mathbf{R}_{t} \mathbf{H}_{\mathbf{v}}^{\intercal} )}}\exp\bigg( -\frac{1}{2}(\mathbf{z}_{t} - \mathbf{H}_{t}\mathbf{x}_{t} - \tilde{\mathbf{z}}_{t})^{\intercal}(  \mathbf{H}_{\mathbf{v}}\mathbf{R}_{t} \mathbf{H}_{\mathbf{v}}^{\intercal} )^{-1}(\mathbf{z}_{t}- \mathbf{H}_{t}\mathbf{x}_{t} - \tilde{\mathbf{z}}_{t}) \bigg)
		\end{aligned}
	\end{equation}
	
	 $\mathbf{F}_{\mathbf{w}}\mathbf{Q}_{t} \mathbf{F}_{\mathbf{w}}^{\intercal}$ represents the noise in the linearized motion model, and $\mathbf{H}_{\mathbf{v}}\mathbf{R}_{t} \mathbf{H}_{\mathbf{v}}^{\intercal}$ represents the noise in the linearized observation model. Next, $\overline{\text{bel}}(\mathbf{x}_{t}), \text{bel}(\mathbf{x}_{t})$ that need to be found through the Kalman filter can be expressed as follows.
	\begin{equation}
		\begin{aligned}
			& \overline{\text{bel}}(\mathbf{x}_{t}) = \int p(\mathbf{x}_{t} | \mathbf{x}_{t-1}, \mathbf{u}_{t})\text{bel}(\mathbf{x}_{t-1}) d \mathbf{x}_{t-1} \sim \mathcal{N}(\hat{\mathbf{x}}_{t|t-1}, \mathbf{P}_{t|t-1}) \\
			& \text{bel}(\mathbf{x}_{t}) = \eta \cdot p( \mathbf{z}_{t} |  \mathbf{x}_{t})\overline{\text{bel}}(\mathbf{x}_{t}) \sim \mathcal{N}(\hat{\mathbf{x}}_{t|t}, \mathbf{P}_{t|t})
		\end{aligned}
	\end{equation}
	
	 EKF operates in a similar manner to KF, using the values from the previous step and the motion model in the prediction to first obtain the predicted value $\overline{\text{bel}}(\mathbf{x}_{t})$, and then using the observed values and observation model in the correction to obtain the corrected value $\text{bel}(\mathbf{x}_{t})$. \textbfazure{Substitute $p(\mathbf{x}_{t} | \mathbf{x}_{t-1}, \mathbf{u}_{t})$, $p(\mathbf{z}_{t} | \mathbf{x}_{t})$ into the above equation to obtain the mean and covariance of the prediction and correction steps $(\hat{\mathbf{x}}_{t|t-1}, \mathbf{P}_{t|t-1}), (\hat{\mathbf{x}}_{t|t}, \mathbf{P}_{t|t})$ respectively.} For a detailed derivation process, refer to Section \ref{sec:derivkf}.
	
	 The initial value $\text{bel}(\mathbf{x}_{0})$ is given as follows.
	\begin{equation}
		\begin{aligned}
			\text{bel}(\mathbf{x}_{0}) \sim \mathcal{N}(\hat{\mathbf{x}}_{0}, \mathbf{P}_{0})
		\end{aligned}
	\end{equation}
	- $\hat{\mathbf{x}}_{0}$: Typically set to 0\\
	- $\mathbf{P}_{0}$ : Typically set to a small value (<1e-2)\\
	
	\subsection{Prediction step}
	 Prediction involves calculating $\overline{\text{bel}}(\mathbf{x}_{t})$. The covariance matrix is obtained using the linearized Jacobian matrix $\mathbf{F}_{t}$.
	\begin{equation}
		\boxed{  \begin{aligned}
				& \hat{\mathbf{x}}_{t|t-1} = \mathbf{f}(\hat{\mathbf{x}}_{t-1|t-1}, \mathbf{u}_{t} , 0 ) \\
				& \mathbf{P}_{t|t-1} = \mathbf{F}_{t}\mathbf{P}_{t-1|t-1}\mathbf{F}_{t}^{\intercal} + \mathbf{F}_{\mathbf{w}}\mathbf{Q}_{t} \mathbf{F}_{\mathbf{w}}^{\intercal}
		\end{aligned} }
	\end{equation}
	
	\subsection{Correction step}
 Correction involves calculating $\text{bel}(\mathbf{x}_{t})$. The Kalman gain and covariance matrix are obtained using the linearized Jacobian matrix $\mathbf{H}_{t}$.
	\begin{equation} \label{eq:ekf3}
		\boxed{ \begin{aligned}
				&     \mathbf{K}_{t} = \mathbf{P}_{t|t-1}\mathbf{H}_{t}^{\intercal}(\mathbf{H}_{t}\mathbf{P}_{t|t-1}\mathbf{H}_{t}^{\intercal} + \mathbf{H}_{\mathbf{v}}\mathbf{R}_{t} \mathbf{H}_{\mathbf{v}}^{\intercal} )^{-1} \\
				& \hat{\mathbf{x}}_{t|t}  = \hat{\mathbf{x}}_{t|t-1} + \mathbf{K}_{t}( \mathbf{z}_{t} - \mathbf{h}(\hat{\mathbf{x}}_{t|t-1}, 0)) \\
				&     \mathbf{P}_{t|t}    = (\mathbf{I} - \mathbf{K}_{t}\mathbf{H}_{t})\mathbf{P}_{t|t-1}
		\end{aligned} } 
	\end{equation}
	
	\subsection{Summary}
	 The Extended Kalman Filter can be expressed as a function as follows.
	\begin{equation} 
		\boxed{ \begin{aligned} 
				& \text{ExtendedKalmanFilter}(\hat{\mathbf{x}}_{t-1|t-1}, \mathbf{P}_{t-1|t-1}, \mathbf{u}_{t}, \mathbf{z}_{t}) \{ \\ 
				& \quad\quad \text{(Prediction Step)}\\ 
				& \quad\quad \hat{\mathbf{x}}_{t|t-1} = \mathbf{f}_{t}(\hat{\mathbf{x}}_{t-1|t-1},  \mathbf{u}_{t} , 0  ) \\ 
				& \quad\quad \mathbf{P}_{t|t-1} = \mathbf{F}_{t}\mathbf{P}_{t-1|t-1}\mathbf{F}_{t}^{\intercal} + \mathbf{F}_{\mathbf{w}}\mathbf{Q}_{t} \mathbf{F}_{\mathbf{w}}^{\intercal} \\ 
				& \\ 
				& \quad\quad \text{(Correction Step)} \\ 
				& \quad\quad \mathbf{K}_{t} = \mathbf{P}_{t|t-1}\mathbf{H}_{t}^{\intercal}(\mathbf{H}_{t}\mathbf{P}_{t|t-1}\mathbf{H}_{t}^{\intercal} + \mathbf{H}_{\mathbf{v}}\mathbf{R}_{t} \mathbf{H}_{\mathbf{v}}^{\intercal} )^{-1} \\ 
				& \quad\quad \hat{\mathbf{x}}_{t|t} = \hat{\mathbf{x}}_{t|t-1} + \mathbf{K}_{t}( \mathbf{z}_{t} - \mathbf{h}_{t}(\hat{\mathbf{x}}_{t|t-1}, 0)) \\ 
				& \quad\quad \mathbf{P}_{t|t} = (\mathbf{I} - \mathbf{K}_{t}\mathbf{H}_{t})\mathbf{P}_{t|t-1} \\ 
				& \quad\quad \text{return} \ \ \hat{\mathbf{x}}_{t|t}, \mathbf{P}_{t|t} \\ 
				&   \} 
		\end{aligned} }
	\end{equation}
	
	\section{Error-state Kalman Filter (ESKF)}
	\textbf{NOMENCLATURE of Error-State Kalman Filter}
	\begin{itemize}
		\item prediction: $\overline{\text{bel}}(\delta \mathbf{x}_{t}) \sim \mathcal{N}(\delta \hat{\mathbf{x}}_{t|t-1}, {\mathbf{P}}_{t|t-1})$     
		\begin{itemize}
			\item $\delta \hat{\mathbf{x}}_{t|t-1}$ : The mean at step $t$ given the correction value at step $t-1$. Some literature also denotes this as $\delta \mathbf{x}^{-}_{t}$.
			\item $\hat{\mathbf{P}}_{t|t-1}$ : The covariance at step $t$ given the correction value at step $t-1$. Some literature also denotes this as $ \mathbf{P}^{-}_{t}$.
		\end{itemize}
		\item correction: $\text{bel}(\delta \mathbf{x}_{t}) \sim \mathcal{N}(\delta \hat{\mathbf{x}}_{t|t}, \mathbf{P}_{t|t})$  
		\begin{itemize}
			\item $\delta \hat{\mathbf{x}}_{t|t}$ : The mean at step $t$ given the prediction value at step $t$. Some literature also denotes this as $\delta \mathbf{x}^{+}_{t}$.
			\item $\hat{\mathbf{P}}_{t|t}$ : The covariance at step $t$ given the prediction value at step $t$. Some literature also denotes this as $\mathbf{P}^{+}_{t}$.
		\end{itemize}
	\end{itemize}
	
	\begin{figure}[h!]
		\centering
		\includegraphics[width=12cm]{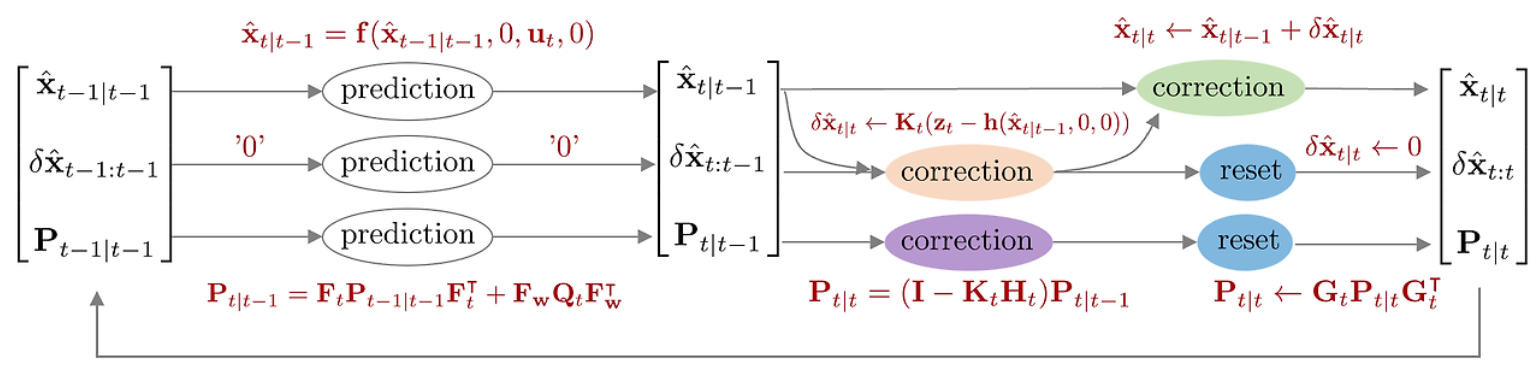}
	\end{figure}
	\textbf{The error-state Kalman filter (ESKF) is a Kalman filter algorithm that estimates the mean and variance of the error state variable $\delta \mathbf{x}_{t}$, unlike the EKF which estimates the mean and variance of the conventional state variable $\mathbf{x}_{t}$.} Since it estimates through the error state rather than directly estimating the state variable, it is also called an indirect Kalman filter. It is also known as the error-state extended Kalman filter (ES-EKF). In the ESKF, the conventional state variable is called the true state variable, and it is represented as a sum of the nominal (nominal) state and the error (error) state.
	\begin{equation} \label{eq:eskf33}
		\begin{aligned}
			\mathbf{x}_{\text{true},t} = \mathbf{x}_{t} + \delta \mathbf{x}_{t}
		\end{aligned}
	\end{equation}
	- $\mathbf{x}_{\text{true},t}$: True state variable updated in conventional KF, EKF at step $t$ \\
	- $\mathbf{x}_{t}$: Nominal state variable at step $t$. Represents a state without error \\
	- $\delta \mathbf{x}_{t}$: Error state variable at step $t$\\
	
	If we interpret the above formula, it means that the actual (true) state variable $\mathbf{x}_{t}$ we want to estimate can be represented as a sum of the general (or nominal) state $\hat{\mathbf{x}}_{t}$ without errors and the error state $\delta \mathbf{x}_{t}$ arising from model and sensor noise. Here, the nominal state has (relatively) large values and is nonlinear. In contrast, the error state has small values near zero and is linear. \textbf{While the conventional EKF linearizes the nonlinear true (nominal + error) state variable for filtering, resulting in slow speed and accumulated errors over time, the ESKF linearizes only the error state for filtering, thus having faster speed and accuracy.} The advantages of ESKF over conventional EKF are as follows (Madyastha et al., 2011):
	\begin{itemize}
		\item The representation of the orientation error state involves minimal parameters. That is, it has the minimum number of parameters as degrees of freedom, so phenomena like singularities due to over-parameterization do not occur.
		\item The error state system only operates near the origin, making it easy to linearize. Therefore, parameter singularities like gimbal lock do not occur, and it is always possible to perform linearization.
		
		\item Since error states typically have small values, terms of second order and higher can be ignored. This helps perform Jacobian operations quickly and easily. Some Jacobians are also used as constants.
	\end{itemize}
	
	However, while the prediction speed of the ESKF is fast, the correction step where the nominal state with generally nonlinear large values is processed is slow.
	~\\~\\
	
	The motion model and observation model of the ESKF are as follows.
	\begin{equation}
		\begin{aligned}
			& \text{Error-state Motion Model: } & \mathbf{x}_{t} + \delta \mathbf{x}_{t} = \mathbf{f}(\mathbf{x}_{t-1}, \delta \mathbf{x}_{t-1},\mathbf{u}_{t}, \mathbf{w}_{t})  \\  
			& \text{Error-state Observation Model: } &  \mathbf{z}_{t} = \mathbf{h}(\mathbf{x}_{t}, \delta \mathbf{x}_{t}, \mathbf{v}_{t})
		\end{aligned}
	\end{equation}
	- $\mathbf{x}_{t}$: Model's nominal state variable\\
	- $\delta \mathbf{x}_{t}$: Model's error state variable\\
	- $\mathbf{u}_{t}$: Model's input\\
	- $\mathbf{z}_{t}$: Model's measurement\\
	- $\mathbf{f}(\cdot)$: Nonlinear motion model function\\
	- $\mathbf{h}(\cdot)$: Nonlinear observation model function\\
	- $\mathbf{w}_{t} \sim \mathcal{N}(0, \mathbf{Q}_{t})$: Noise of the error state model. $\mathbf{Q}_{t}$ denotes the covariance matrix of $\mathbf{w}_{t}$\\
	- $\mathbf{v}_{t} \sim \mathcal{N}(0, \mathbf{R}_{t})$: Noise of the observation model. $\mathbf{R}_{t}$ denotes the covariance matrix of $\mathbf{v}_{t}$\\
	
	When first-order Taylor approximations are performed for $\mathbf{f}(\cdot), \mathbf{h}(\cdot)$, the following expansions result. This development refers to \ref{ref:5}.
	\begin{equation} \label{eq:33}
		\begin{aligned} 
			& \mathbf{x}_{t} + \delta \mathbf{x}_{t} \approx \mathbf{f} ( \mathbf{x}_{t-1|t-1}, \delta \hat{\mathbf{x}}_{t-1|t-1},\mathbf{u}_{t}, 0) + \mathbf{F}_{t}(\delta  \mathbf{x}_{t-1} - \delta 
			\hat{\mathbf{x}}_{t-1|t-1}) + \mathbf{F}_{\mathbf{w}} \mathbf{w}_{t} \\
			&  \mathbf{z}_{t} \approx \mathbf{h}( 
			\mathbf{x}_{t|t-1}, \delta \hat{\mathbf{x}}_{t|t-1}, 0) + \mathbf{H}_{t}(\delta \mathbf{x}_{t} - \delta \hat{\mathbf{x}}_{t|t-1}) + \mathbf{H}_{\mathbf{v}} \mathbf{v}_{t}
		\end{aligned} 
	\end{equation}
	
	\textbf{Note that both Jacobians, $\mathbf{F}_{t}, \mathbf{H}_{t}$, are for the error state $\delta \mathbf{x}_{t}$, not the true state $\mathbf{x}_{\text{true},t}$.} This is the most significant difference from EKF in the Jacobian part.
	\begin{equation} \label{eq:eskf36}
		\boxed { \begin{aligned}
				& \mathbf{F}_{t} = \frac{\partial \mathbf{f}} {\partial \delta \mathbf{x}_{t-1}}\bigg|_{\substack{ \delta \mathbf{x}_{t-1} = \delta \hat{\mathbf{x}}_{t-1|t-1} }} & \mathbf{F}_{\mathbf{w}}  = \frac{\partial \mathbf{f}} {\partial \mathbf{w}_{t}}\bigg|_{\substack{  \delta \mathbf{x}_{t-1} = \delta \hat{\mathbf{x}}_{t-1|t-1} \\ \mathbf{w}_{t}=0}}
		\end{aligned} }
	\end{equation}
	\begin{equation}
		\boxed{ \begin{aligned}
				& \mathbf{H}_{t} = \frac{\partial \mathbf{h}}{\partial \delta \mathbf{x}_{t}}\bigg|_{\delta \mathbf{x}_{t}= \delta  \hat{\mathbf{x}}_{t|t-1}}  & \mathbf{H}_{\mathbf{v}} = \frac{\partial \mathbf{h}}{\partial \mathbf{v}_{t}}\bigg|_{ \substack{  \delta \mathbf{x}_{t}= \delta \hat{\mathbf{x}}_{t|t-1} \\  \mathbf{v}_{t}=0 }}
		\end{aligned} }
	\end{equation}
	
	$\mathbf{H}_{t}$ can be expressed using the chain rule as follows.
	\begin{equation} \label{eq:eskf37}
		\begin{aligned} 
			\mathbf{H}_{t} = \frac{\partial \mathbf{h}}{\partial \delta \mathbf{x}_{t}} = \frac{\partial \mathbf{h}}{\partial \mathbf{x}_{\text{true},t}} \frac{\partial \mathbf{x}_{\text{true},t}}{\partial \delta \mathbf{x}_{t}} \\
		\end{aligned} 
	\end{equation}
	
	In this, the front part $\frac{\partial \mathbf{h}}{\partial \mathbf{x}_{\text{true},t}}$ is the same Jacobian obtained in EKF, but a Jacobian for the error state variable $\frac{\partial \mathbf{x}_{\text{true},t}}{\partial \delta \mathbf{x}_{t}}$ has been added. For more details on this, see the post \href{https://alida.tistory.com/63#6.1.1-jacobian-computation-for-the-filter-correction}{Quaternion kinematics for the error-state Kalman filter summary}.
	
	Expanding ($\ref{eq:33})$ results in the following.
	\begin{equation} 
		\begin{aligned} 
			\mathbf{x}_{t} + \delta \mathbf{x}_{t} & = \mathbf{F}_{t} \delta \mathbf{x}_{t-1} +  \mathbf{f} ( \mathbf{x}_{t-1|t-1}, \delta \hat{\mathbf{x}}_{t-1|t-1},\mathbf{u}_{t}, 0) - \mathbf{F}_{t} \delta  
			\hat{\mathbf{x}}_{t-1|t-1} + \mathbf{F}_{\mathbf{w}} \mathbf{w}_{t} 
		\end{aligned} 
	\end{equation}
	
	Since $\delta \hat{\mathbf{x}}_{t-1|t-1}=0$ is always initialized at the previous correction step, 0 is substituted for the related terms.
	\begin{equation} 
		\begin{aligned} 
			\mathbf{x}_{t} + \delta \mathbf{x}_{t} & = \mathbf{F}_{t} \delta \mathbf{x}_{t-1} +  \mathbf{f} ( \mathbf{x}_{t-1|t-1}, 0,\mathbf{u}_{t}, 0) + \mathbf{F}_{\mathbf{w}} \mathbf{w}_{t} 
		\end{aligned} 
	\end{equation}
	
	The nominal state variable $\mathbf{x}_{t}$ is identical to $\mathbf{f}(\mathbf{x}_{t-1}, 0, \mathbf{u}_{t}, 0)$ by definition, so they cancel each other out.
	\begin{equation} 
		\boxed { \begin{aligned} 
				\delta \mathbf{x}_{t} & = \mathbf{F}_{t} \delta \mathbf{x}_{t-1}  + \mathbf{F}_{\mathbf{w}} \mathbf{w}_{t} \\ 
				& = \mathbf{F}_{t} \delta \mathbf{x}_{t-1} +  \tilde{\mathbf{w}}_{t} \\ 
				& = 0 + \tilde{\mathbf{w}}_{t}
		\end{aligned}  }
	\end{equation}
	- $\tilde{\mathbf{w}}_{t} = \mathbf{F}_{\mathbf{w}} \mathbf{w}_{t} \sim \mathcal{N}(0, \mathbf{F}_{\mathbf{w}}\mathbf{Q}_{t} \mathbf{F}_{\mathbf{w}}^{\intercal})$
	
	The observation model function is similarly developed by substituting the error state variable values with 0.
	\begin{equation}  \label{eq:eskf-z}
		\boxed{  \begin{aligned} 
				\mathbf{z}_{t} &= \mathbf{H}_{t} \delta \mathbf{x}_{t} + \mathbf{h}(  
				\mathbf{x}_{t|t-1}, \delta \hat{\mathbf{x}}_{t-1|t-1}, 0) - \mathbf{H}_{t} \delta \hat{\mathbf{x}}_{t|t-1} + \mathbf{H}_{\mathbf{v}} \mathbf{v}_{t} \\
				& = \mathbf{h}( \mathbf{x}_{t|t-1}, 0, 0) + \mathbf{H}_{\mathbf{v}} \mathbf{v}_{t} \\
				& =  \tilde{\mathbf{z}}_{t} + \tilde{\mathbf{v}}_{t}
		\end{aligned} }
	\end{equation}
	- $\tilde{\mathbf{z}}_{t} = \mathbf{h}( \mathbf{x}_{t|t-1}, 0, 0)$\\
	- $\tilde{\mathbf{v}}_{t} = \mathbf{H}_{\mathbf{v}} \mathbf{v}_{t} \sim \mathcal{N}(0, \mathbf{H}_{\mathbf{v}}\mathbf{R}_{t} \mathbf{H}_{\mathbf{v}}^{\intercal})$\\

	Assuming all random variables follow a Gaussian distribution, $p(\delta \mathbf{x}_{t} \ | \ \delta \mathbf{x}_{t-1}, \mathbf{u}_{t}), p(\mathbf{z}_{t} \ | \ \delta \mathbf{x}_{t})$ can be represented as follows. 
	\begin{equation}
	\begin{aligned}
		& p(\delta \mathbf{x}_{t} \ | \ \delta \mathbf{x}_{t-1}, \mathbf{u}_{t}) && \sim \mathcal{N}(0, \mathbf{F}_{\mathbf{w}}\mathbf{Q}_{t} \mathbf{F}_{\mathbf{w}}^{\intercal} ) \\
		& && = \frac{1}{ \sqrt{\text{det}(2\pi \mathbf{F}_{\mathbf{w}}\mathbf{Q}_{t} \mathbf{F}_{\mathbf{w}}^{\intercal} )}}\exp\bigg( -\frac{1}{2}(\delta \mathbf{x}_{t} )^{\intercal} (\mathbf{F}_{\mathbf{w}}\mathbf{Q}_{t} \mathbf{F}_{\mathbf{w}}^{\intercal})^{-1}(\delta \mathbf{x}_{t} ) \bigg)\\ 
	\end{aligned}
	\end{equation}
	\begin{equation} \label{eq:eskf2}
	\begin{aligned}
		& p(\mathbf{z}_{t} \ | \ \delta \mathbf{x}_{t})  && \sim \mathcal{N}( \tilde{\mathbf{z}}_{t}, \mathbf{H}_{\mathbf{v}}\mathbf{R}_{t}\mathbf{H}_{\mathbf{v}}^{\intercal}) \\
		& && = \frac{1}{ \sqrt{\text{det}(2\pi \mathbf{H}_{\mathbf{v}}\mathbf{R}_{t}\mathbf{H}_{\mathbf{v}}^{\intercal})}}\exp\bigg( -\frac{1}{2}(\mathbf{z}_{t} - \tilde{\mathbf{z}}_{t})^{\intercal} (\mathbf{H}_{\mathbf{v}}\mathbf{R}_{t}\mathbf{H}_{\mathbf{v}}^{\intercal})^{-1}(\mathbf{z}_{t} - \tilde{\mathbf{z}}_{t}) \bigg)
	\end{aligned}
	\end{equation}
	
	$\mathbf{F}_{\mathbf{w}}\mathbf{Q}_{t} \mathbf{F}_{\mathbf{w}}^{\intercal}$ denotes the noise of the linearized error state motion model, and $\mathbf{H}_{\mathbf{v}}\mathbf{R}_{t}\mathbf{H}_{\mathbf{v}}^{\intercal}$ denotes the noise of the linearized observation model. Next, the Kalman filter needs to derive $\overline{\text{bel}}(\delta \mathbf{x}_{t}), \text{bel}(\delta \mathbf{x}_{t})$ as follows.
	\begin{equation}
	\begin{aligned}
		& \overline{\text{bel}}(\delta \mathbf{x}_{t}) = \int p(\delta \mathbf{x}_{t} \ | \ \delta \mathbf{x}_{t-1}, \mathbf{u}_{t})\text{bel}(\delta \mathbf{x}_{t-1}) d \mathbf{x}_{t-1}  \sim \mathcal{N}(\delta \hat{\mathbf{x}}_{t|t-1}, \mathbf{P}_{t|t-1}) \\
		& \text{bel}(\delta \mathbf{x}_{t}) = \eta \cdot p( \mathbf{z}_{t} \ | \  \delta \mathbf{x}_{t})\overline{\text{bel}}(\delta \mathbf{x}_{t}) \sim \mathcal{N}(\delta \hat{\mathbf{x}}_{t|t}, \mathbf{P}_{t|t})
	\end{aligned}
	\end{equation}
	
	ESKF also operates in the same way as EKF, where the prediction first derives the predicted value $\overline{\text{bel}}(\delta \mathbf{x}_{t})$ using the previous step's value and motion model, and then the correction derives the corrected value $\text{bel}(\delta \mathbf{x}_{t})$ using the measurement value and observation model. \textbfazure{Substituting $p(\delta \mathbf{x}_{t} | \delta \mathbf{x}_{t-1}, \mathbf{u}_{t}), p(\mathbf{z}_{t} | \delta \mathbf{x}_{t})$ into the above formula allows you to derive the mean and covariance for the prediction and correction steps, $(\delta \hat{\mathbf{x}}_{t|t-1}, \mathbf{P}_{t|t-1}), (\delta \hat{\mathbf{x}}_{t|t}, \mathbf{P}_{t|t})$, respectively.} For a detailed derivation process, refer to Section \ref{sec:derivkf}.

	The initial value $\text{bel}(\delta \mathbf{x}_{0})$ is given as follows.
	\begin{equation}
	\begin{aligned}
		\text{bel}(\delta \mathbf{x}_{0}) \sim \mathcal{N}(0, \mathbf{P}_{0})
	\end{aligned}
	\end{equation}
	- $\delta \hat{\mathbf{x}}_{0} = 0$ : Always has a value of 0\\
	- $\mathbf{P}_{0}$ : Typically set to small values (<1e-2).

	\subsection{Prediction step}
	Prediction is the process of deriving $\overline{\text{bel}}(\delta \mathbf{x}_{t})$. The linearized Jacobian matrix $\mathbf{F}_{t}$ is used to derive the covariance matrix.
	\begin{equation} \label{eq:eskf-pred}
	\boxed{  \begin{aligned}
			& \delta \hat{\mathbf{x}}_{t|t-1} = \mathbf{F}_{t}\delta \hat{\mathbf{x}}_{t-1|t-1} = 0 {\color{Mahogany} \quad \leftarrow \text{Always 0} } \\
			& \hat{\mathbf{x}}_{t|t-1} = \mathbf{f}(\hat{\mathbf{x}}_{t-1|t-1}, 0,\mathbf{u}_{t}, 0) \\ 
			& \mathbf{P}_{t|t-1} = \mathbf{F}_{t}\mathbf{P}_{t-1|t-1}\mathbf{F}_{t}^{\intercal} + \mathbf{F}_{\mathbf{w}}\mathbf{Q}_{t}  \mathbf{F}_{\mathbf{w}}^{\intercal}
	\end{aligned} }
	\end{equation}
	
	Note that $\mathbf{F}_{t}$ is the Jacobian for the error state. Since the error state variable $\delta \hat{\mathbf{x}}$ is reset to 0 at every correction step, multiplying it by the linear Jacobian $\mathbf{F}_{t}$ still results in 0. \textbf{Therefore, the value of $\delta \hat{\mathbf{x}}$ always remains 0 in the prediction step.} Thus, the error state $\delta \hat{\mathbf{x}}_{t|t-1}$ remains unchanged in the prediction step, while only the nominal state $\hat{\mathbf{x}}_{t|t-1}$ and the error state covariance $\mathbf{P}_{t|t-1}$ are updated.
	
	\subsection{Correction step}
	Correction is the process of deriving $\text{bel}(\delta \mathbf{x}_{t})$. The Kalman gain and covariance matrix are derived using the linearized Jacobian matrix $\mathbf{H}_{t}$.
	\begin{equation} 
	\boxed { \begin{aligned}
			&     \mathbf{K}_{t} = \mathbf{P}_{t|t-1}\mathbf{H}_{t}^{\intercal}(\mathbf{H}_{t}\mathbf{P}_{t|t-1}\mathbf{H}_{t}^{\intercal} + \mathbf{H}_{\mathbf{v}} \mathbf{R}_{t} \mathbf{H}_{\mathbf{v}})^{-1} \\
			& {\color{Mahogany} \delta \hat{\mathbf{x}}_{t|t} = \mathbf{K}_{t}( \mathbf{z}_{t} - \mathbf{h}(\hat{\mathbf{x}}_{t|t-1}, 0,0)) } \\ 
			& \hat{\mathbf{x}}_{t|t}  = \hat{\mathbf{x}}_{t|t-1} +  \delta \hat{\mathbf{x}}_{t|t} \\
			&     \mathbf{P}_{t|t}    = (\mathbf{I} - \mathbf{K}_{t}\mathbf{H}_{t})\mathbf{P}_{t|t-1} \\
			& \text{reset } \delta \hat{\mathbf{x}}_{t|t}  \\
			& \quad \quad \quad \delta \hat{\mathbf{x}}_{t|t} \leftarrow 0 \\
			&  \quad \quad \quad \mathbf{P}_{t|t} \leftarrow \mathbf{G}\mathbf{P}_{t|t}\mathbf{G}^{\intercal}
	\end{aligned} } \label{eq:eskf1}
	\end{equation}
	
	\textbf{Note that in the above formula, $\mathbf{H}_{t}$ is the Jacobian for the observation model of the error state $\delta \mathbf{x}_{t}$, not the true state $\mathbf{x}_{\text{true},t}$. The symbols $\mathbf{P}_{t|t-1}, \mathbf{K}_{t}$ are only similar to those in EKF, but the actual values are different.} Thus, while the overall formulas are the same as in EKF, the matrices $\mathbf{F},\mathbf{H},\mathbf{P},\mathbf{K}$ denote values for the error state $\delta \mathbf{x}_{t}$, which is a key difference.
	
	\subsubsection{Reset}
	Once the nominal state is updated, the next step is to reset the value of the error state to 0. The reset is necessary because it is necessary to represent a new error for the new nominal state. The reset updates the covariance of the error state $\mathbf{P}_{t|t}$.
	
	If the reset function is denoted as $\mathbf{g}(\cdot)$, it can be represented as follows. For more details on this, see the chapter 6 content in the post \href{https://alida.tistory.com/63#6.3-eskf-reset}{Quaternion kinematics for the error-state Kalman filter summary}.
	\begin{equation}
	\begin{aligned}
		& \delta \mathbf{x} \leftarrow \mathbf{g}(\delta \mathbf{x}) =  \delta \mathbf{x} - \delta \hat{\mathbf{x}}_{t|t-1}
	\end{aligned}
	\end{equation}
	
	The reset process in ESKF is as follows.
	\begin{equation}
	\begin{aligned}
		& \delta \hat{\mathbf{x}}_{t|t} \leftarrow 0 \\
		& \mathbf{P}_{t|t} \leftarrow \mathbf{G}\mathbf{P}_{t|t}\mathbf{G}^{\intercal}
	\end{aligned}
	\end{equation}
	
	$\mathbf{G}$ denotes the Jacobian for the reset defined as follows.
	\begin{equation}
	\begin{aligned}
		\mathbf{G} = \left. \frac{\partial \mathbf{g}}{\partial \delta \mathbf{x}}\right\vert_{\delta \mathbf{x}_{t}=\delta \hat{\mathbf{x}}_{t|t}}
	\end{aligned}
	\end{equation}

	\subsection{Summary}
	\begin{equation}  
		\boxed{ \begin{aligned}  
				& \text{ErrorStateKalmanFilter}( \hat{\mathbf{x}}_{t-1|t-1} , \delta \hat{\mathbf{x}}_{t-1|t-1}, \mathbf{P}_{t-1|t-1}, \mathbf{u}_{t}, \mathbf{z}_{t}) \{ \\  
				& \quad\quad \text{(Prediction Step)}\\  
				& \quad\quad \delta \hat{\mathbf{x}}_{t|t-1} = \mathbf{F}_{t}\hat{\mathbf{x}}_{t-1|t-1}= 0 {\color{Mahogany} \quad \leftarrow \text{Always 0} } \\  
				& \quad\quad  \hat{\mathbf{x}}_{t|t-1} = \mathbf{f}(\hat{\mathbf{x}}_{t-1|t-1}, 0,\mathbf{u}_{t}, 0) \\ 
				& \quad\quad \mathbf{P}_{t|t-1} = \mathbf{F}_{t}\mathbf{P}_{t-1|t-1}\mathbf{F}_{t}^{\intercal} + \mathbf{F}_{\mathbf{w}}\mathbf{Q}_{t} \mathbf{F}_{\mathbf{w}}^{\intercal}\\  
				& \\  
				& \quad\quad \text{(Correction Step)} \\  
				& \quad\quad \mathbf{K}_{t} = \mathbf{P}_{t|t-1}\mathbf{H}_{t}^{\intercal}(\mathbf{H}_{t}\mathbf{P}_{t|t-1}\mathbf{H}_{t}^{\intercal} + \mathbf{H}_{\mathbf{v}}\mathbf{R}_{t} \mathbf{H}_{\mathbf{v}}^{\intercal})^{-1} \\  
				& \quad\quad \delta \hat{\mathbf{x}}_{t|t} =  \mathbf{K}_{t}( \mathbf{z}_{t} - \mathbf{h}_{t}(\hat{\mathbf{x}}_{t|t-1},0,0)) \\  
				& \quad\quad \hat{\mathbf{x}}_{t|t} = \hat{\mathbf{x}}_{t|t-1} + \delta \hat{\mathbf{x}}_{t|t} \\  
				& \quad\quad \mathbf{P}_{t|t} = (\mathbf{I} - \mathbf{K}_{t}\mathbf{H}_{t})\mathbf{P}_{t|t-1} \\  
				& \quad \quad \text{reset } \delta \hat{\mathbf{x}}_{t|t}  \\
				& \quad \quad \quad  \delta \hat{\mathbf{x}}_{t|t} \leftarrow 0 \\
				& \quad \quad \quad \mathbf{P}_{t|t} \leftarrow \mathbf{G}\mathbf{P}_{t|t}\mathbf{P}^{\intercal} \\
				& \quad\quad \text{return} \ \ \delta \hat{\mathbf{x}}_{t|t}, \mathbf{P}_{t|t} \\  
				&   \}  
		\end{aligned} } 
	\end{equation}
	
	\section{Iterated Extended Kalman Filter (IEKF)}
	\begin{figure}[h!]
		\centering
		\includegraphics[width=12cm]{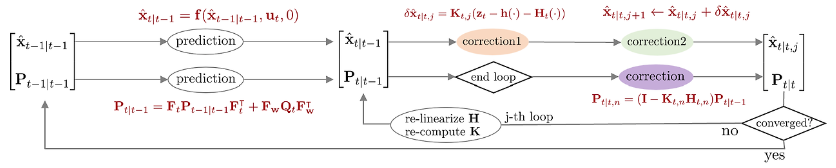}
	\end{figure}
	The Iterated Extended Kalman Filter (IEKF) is an algorithm that repetitively performs the correction step of the Extended Kalman Filter (EKF). Since EKF linearizes non-linear functions to estimate state variables, errors inevitably occur during the linearization process. IEKF reduces these linearization errors by repeatedly performing the linearization if the update change $\delta \hat{\mathbf{x}}_{t|t,j}$ after the correction step is sufficiently large.

	In this context, $\delta \hat{\mathbf{x}}$ is interpreted as the update change, not an error state variable. That is, $\hat{\mathbf{x}}_{t|t,j+1} \leftarrow \hat{\mathbf{x}}_{t|t,j} + \delta \hat{\mathbf{x}}_{t|t,j}$ is used solely to update the j-th posterior value to the j+1 posterior value. In other words, it is not the target of state estimation.
	
	\subsection{Compare to EKF}
	\subsubsection{Commonality 1}
	In IEKF, the motion and observation models are as follows, which are exactly the same as in EKF.
	
	\begin{equation}
		\begin{aligned}
			& \text{Motion Model: } & \mathbf{x}_{t} = \mathbf{f}(\mathbf{x}_{t-1}, \mathbf{u}_{t} , \mathbf{w}_{t}) \\
			& \text{Observation Model: } &  \mathbf{z}_{t} = \mathbf{h}(\mathbf{x}_{t}, \mathbf{v}_{t})
		\end{aligned}
	\end{equation}
	- $\mathbf{x}_{t}$: state variable of the model
	- $\mathbf{u}_{t}$: input of the model
	- $\mathbf{z}_{t}$: measurement of the model
	- $\mathbf{f}(\cdot)$: non-linear motion model function 
	- $\mathbf{h}(\cdot)$: non-linear observation model function
	- $\mathbf{w}_{t} \sim \mathcal{N}(0, \mathbf{Q}_{t})$: noise in the motion model. $\mathbf{Q}_{t}$ represents the covariance matrix of $\mathbf{w}_{t}$
	- $\mathbf{v}_{t} \sim \mathcal{N}(0, \mathbf{R}_{t})$: noise in the observation model. $\mathbf{R}_{t}$ represents the covariance matrix of $\mathbf{v}_{t}$

	\subsubsection{Commonality 2}
	Next, the linearization process is also exactly the same as EKF's equations ($\ref{eq:ekf1}$), ($\ref{eq:ekf2}$).
	
	\begin{equation} 
		\boxed{  \begin{aligned}
				\mathbf{x}_{t} & = \mathbf{F}_{t}\mathbf{x}_{t-1} + 
				\mathbf{f}(\hat{\mathbf{x}}_{t-1|t-1}, \mathbf{u}_{t} , 0 ) - \mathbf{F}_{t} \hat{\mathbf{x}}_{t-1|t-1} + \mathbf{F}_{\mathbf{w}} \mathbf{w}_{t} \\
				& =  \mathbf{F}_{t}\mathbf{x}_{t-1} + \tilde{\mathbf{u}}_{t} + \tilde{\mathbf{w}}_{t} 
		\end{aligned} }
	\end{equation}
	- $\tilde{\mathbf{u}}_{t} = \mathbf{f}(\hat{\mathbf{x}}_{t-1|t-1}, \mathbf{u}_{t} , 0 ) - \mathbf{F}_{t} \hat{\mathbf{x}}_{t-1|t-1}$
	- $\tilde{\mathbf{w}}_{t} = \mathbf{F}_{\mathbf{w}} \mathbf{w}_{t} \sim \mathcal{N}(0,  \mathbf{F}_{\mathbf{w}}\mathbf{Q}_{t} \mathbf{F}_{\mathbf{w}}^{\intercal})$
	
	\begin{equation} \label{eq:iekf1}
		\boxed{ \begin{aligned}
				\mathbf{z}_{t} & = \mathbf{H}_{t}\mathbf{x}_{t} + \mathbf{h}(\hat{\mathbf{x}}_{t|t-1}, 0) - \mathbf{H}_{t}\hat{\mathbf{x}}_{t|t-1} + \mathbf{H}_{\mathbf{v}}\mathbf{v}_{t} \\ 
				& = \mathbf{H}_{t}\mathbf{x}_{t}  + \tilde{\mathbf{z}}_{t} + \tilde{\mathbf{v}}_{t}
		\end{aligned} }
	\end{equation}
	- $\tilde{\mathbf{z}}_{t} = \mathbf{h}(\hat{\mathbf{x}}_{t|t-1}, 0) - \mathbf{H}_{t}\hat{\mathbf{x}}_{t|t-1}$
	- $\tilde{\mathbf{v}}_{t} = \mathbf{H}_{\mathbf{v}}\mathbf{v}_{t} \sim \mathcal{N}(0,  \mathbf{H}_{\mathbf{v}}\mathbf{R}_{t} \mathbf{H}_{\mathbf{v}}^{\intercal})$

	\subsubsection{Commonality 3}
	IEKF, like EKF, uses the matrices $\mathbf{F}, \mathbf{H}, \mathbf{K}$ true to the state variables $\mathbf{x}_{\text{true}}$.

	\subsubsection{Difference 1}
	\textbfazure{EKF}: The correction value is obtained from the prediction value in one step. \\
	\textbfazure{IEKF}: The correction value becomes the prediction value again and the correction step is repeated iteratively.
	
	\begin{equation}
		\boxed{ \begin{aligned}
				& \text{EKF Correction : } \quad \hat{\mathbf{x}}_{t|t-1} \rightarrow \hat{\mathbf{x}}_{t|t} \\
				& \text{IEKF Correction : } \quad \hat{\mathbf{x}}_{t|t,j} \leftrightarrows \hat{\mathbf{x}}_{t|t,j+1} \\
		\end{aligned} }
	\end{equation}
	
	- $j$: j-th iteration

	\subsubsection{Difference 2}
	Expanding equation ($\ref{eq:iekf1}$) to create the innovation term $\mathbf{r}_{t}$ gives the following.
	
	\begin{equation} 
		\begin{aligned}
			\mathbf{r}_{t} = \mathbf{z}_{t} - \mathbf{h} (\hat{\mathbf{x}}_{t|t-1}, 0) - \mathbf{H}_{t}(\mathbf{x}_{t} -\hat{\mathbf{x}}_{t|t-1}) 
		\end{aligned} 
	\end{equation}

	In EKF, looking at the second line of correction step ($\ref{eq:ekf3}$), it can be seen that the Kalman gain $\mathbf{K}_{t}$ is multiplied by $\mathbf{r}_{t}$ to compute the posterior.
	
	\begin{equation}
		\begin{aligned}
			\text{EKF correction : } \quad & \hat{\mathbf{x}}_{t|t}  && = \hat{\mathbf{x}}_{t|t-1} + \mathbf{K}_{t}( \mathbf{r}_{t}) \\
			& &&  = \hat{\mathbf{x}}_{t|t-1} + \mathbf{K}_{t}( \mathbf{z}_{t} - \mathbf{h}(\hat{\mathbf{x}}_{t|t-1}, 0))
		\end{aligned} 
	\end{equation}

	\textbfazure{EKF}: The $\mathbf{H}_{t}(\mathbf{x}_{t} -\hat{\mathbf{x}}_{t|t-1})$ part is eliminated when $\mathbf{x}_{t} = \hat{\mathbf{x}}_{t|t-1}$ is substituted, and only the remaining parts are used.
	
	\textbfazure{IEKF}: Linearization for a new state (new operating point) is performed at each moment of the correction step, so this part is not eliminated.
	
	\begin{equation} 
		\boxed{ \begin{aligned}
				& \text{EKF innovation term : } \quad     \mathbf{r}_{t} = \mathbf{z}_{t} - \mathbf{h}(\hat{\mathbf{x}}_{t|t-1}, 0)  \\ 
				& \text{IEKF innovation term : } \quad     \mathbf{r}_{t,j} = \mathbf{z}_{t} - \mathbf{h}(\hat{\mathbf{x}}_{t|t,j}, 0) - \mathbf{H}_{t,j}(\hat{\mathbf{x}}_{t|t-1} - \hat{\mathbf{x}}_{t|t,j})  \\ 
		\end{aligned} }
	\end{equation}

	If it is the first iteration $j=0$, $\hat{\mathbf{x}}_{t|t,0} = \hat{\mathbf{x}}_{t|t-1}$ so it is eliminated resulting in a formula similar to EKF, but from $j=1$ onwards, different values are used so it is not eliminated.

	\subsection{Prediction step}
	Prediction involves the process of determining $\overline{\text{bel}}(\mathbf{x}_{t})$. The linearized Jacobian matrix $\mathbf{F}_{t}$ is used when computing the covariance matrix. This is identical to the prediction step in EKF.
	\begin{equation}
		\boxed{  \begin{aligned}
				& \hat{\mathbf{x}}_{t|t-1} = \mathbf{f}(\hat{\mathbf{x}}_{t-1|t-1}, \mathbf{u}_{t} , 0 ) \\
				& \mathbf{P}_{t|t-1} = \mathbf{F}_{t}\mathbf{P}_{t-1|t-1}\mathbf{F}_{t}^{\intercal} + \mathbf{F}_{\mathbf{w}}\mathbf{Q}_{t} \mathbf{F}_{\mathbf{w}}^{\intercal}
		\end{aligned} }
	\end{equation}

	\subsection{Correction step}
	Correction involves the process of determining $\text{bel}(\mathbf{x}_{t})$. The Kalman gain and covariance matrix are computed using the linearized Jacobian matrix $\mathbf{H}_{t}$. The IEKF's correction step is performed iteratively until the update change $\delta \hat{\mathbf{x}}_{t|t,j}$ is sufficiently small.
	
	\begin{equation} \label{eq:iekf-cor}
		\boxed {\begin{aligned}
				& \text{set } \epsilon \\
				&  \text{start j-th loop }  \\
				&  \quad   \mathbf{K}_{t,j} = \mathbf{P}_{t|t-1} \mathbf{H}_{t,j}^{\intercal}(\mathbf{H}_{t,j}\mathbf{P}_{t|t-1} \mathbf{H}_{t,j}^{\intercal} + \mathbf{H}_{\mathbf{v},j}\mathbf{R}_{t}\mathbf{H}_{\mathbf{v},j}^{\intercal})^{-1} \\
				& {\color{Mahogany} \quad \delta \hat{\mathbf{x}}_{t|t,j} = \mathbf{K}_{t,j} (\mathbf{z}_{t} - \mathbf{h}(\hat{\mathbf{x}}_{t|t,j}, 0) - \mathbf{H}_{t}(\hat{\mathbf{x}}_{t|t-1} - \hat{\mathbf{x}}_{t|t,j})) } \\
				& \quad \hat{\mathbf{x}}_{t|t, j+1}  = \hat{\mathbf{x}}_{t|t, j} + \delta \hat{\mathbf{x}}_{t|t,j} \\
				& \quad \text{iterate until } \delta \hat{\mathbf{x}}_{t|t,j} < \epsilon.   \\ 
				& \text{end loop} \\ 
				&     \mathbf{P}_{t|t,n}    = (\mathbf{I} - \mathbf{K}_{t,n}\mathbf{H}_{t,n}) \mathbf{P}_{t|t-1}  \\
		\end{aligned} }
	\end{equation}

	\subsection{Summary}	
	\begin{equation} 
		\boxed{ \begin{aligned} 
				& \text{IteratedExtendedKalmanFilter}(\hat{\mathbf{x}}_{t-1|t-1}, \mathbf{P}_{t-1|t-1}, \mathbf{u}_{t}, \mathbf{z}_{t}) \{ \\ 
				& \quad\quad \text{(Prediction Step)}\\ 
				& \quad\quad \hat{\mathbf{x}}_{t|t-1} = \mathbf{f}_{t}(\hat{\mathbf{x}}_{t-1|t-1},  \mathbf{u}_{t} , 0  ) \\ 
				& \quad\quad \mathbf{P}_{t|t-1} = \mathbf{F}_{t}\mathbf{P}_{t-1|t-1}\mathbf{F}_{t}^{\intercal} + \mathbf{F}_{\mathbf{w}}\mathbf{Q}_{t} \mathbf{F}_{\mathbf{w}}^{\intercal} \\ 
				& \\ 
				& \quad\quad \text{(Correction Step)} \\ 
				& \quad\quad \text{set } \epsilon \\
				&  \quad\quad\text{start j-th loop }  \\
				&  \quad\quad \quad   \mathbf{K}_{t,j} = \mathbf{P}_{t|t-1} \mathbf{H}_{t,j}^{\intercal}(\mathbf{H}_{t,j}\mathbf{P}_{t|t-1} \mathbf{H}_{t,j}^{\intercal} + \mathbf{H}_{\mathbf{v},j}\mathbf{R}_{t}\mathbf{H}_{\mathbf{v},j}^{\intercal})^{-1} \\
				& \quad\quad  \quad \delta \hat{\mathbf{x}}_{t|t,j} = \mathbf{K}_{t,j} (\mathbf{z}_{t} - \mathbf{h}(\hat{\mathbf{x}}_{t|t,j}, 0) - \mathbf{H}_{t}(\hat{\mathbf{x}}_{t|t-1} - \hat{\mathbf{x}}_{t|t,j})) \\
				& \quad\quad \quad \hat{\mathbf{x}}_{t|t, j+1}  = \hat{\mathbf{x}}_{t|t, j} + \delta \hat{\mathbf{x}}_{t|t,j} \\
				& \quad\quad \quad \text{iterate until } \delta \hat{\mathbf{x}}_{t|t,j} < \epsilon.   \\ 
				& \quad\quad \text{end loop} \\ 
				&    \quad\quad \mathbf{P}_{t|t,n}    = (\mathbf{I} - \mathbf{K}_{t,n}\mathbf{H}_{t,n}) \mathbf{P}_{t|t-1}  \\ 
				& \quad\quad \text{return} \ \ \hat{\mathbf{x}}_{t|t}, \mathbf{P}_{t|t} \\ 
				&   \} 
		\end{aligned} }
	\end{equation}
	
\section{Iterated Error-State Kalman Filter (IESKF)}
~\\
\begin{figure}[h!]
	\centering
	\includegraphics[width=12cm]{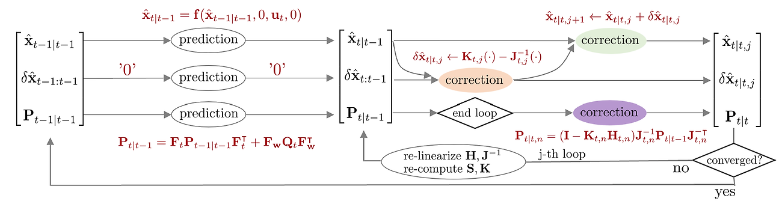}
\end{figure}

While IEKF estimates the true state variable $\mathbf{x}_{\text{true}}$ iteratively in the correction step, IESKF is an algorithm that iteratively estimates the error state variable $\delta \mathbf{x}_{t}$. The relationship between these state variables is as follows.

\begin{equation} 
	\begin{aligned}
		\mathbf{x}_{\text{true},t} = \mathbf{x}_{t} + \delta \mathbf{x}_{t}
	\end{aligned}
\end{equation}
- $\mathbf{x}_{\text{true},t}$: true state variable at step $t$ updated in the conventional KF, EKF \\
- $\mathbf{x}_{t}$: nominal state variable at step $t$. Represents the state without error \\
- $\delta \mathbf{x}_{t}$: error state variable at step $t$\\ 

The iterative update process in IESKF involves understanding $\mathbf{x}_{\text{true},t}$ as the post-update state, nominal  $\mathbf{x}_{t}$ as the pre-update state, and the error state variable $\delta \mathbf{x}_{t}$ as the update amount.

\subsection{Compare to ESKF}
\subsubsection{Commonality 1}
The motion and observation models of IESKF are as follows, which are identical to those of ESKF.
\begin{equation}
	\begin{aligned}
		& \text{Error-state Motion Model: } & \mathbf{x}_{t} + \delta \mathbf{x}_{t} = \mathbf{f}(\mathbf{x}_{t-1}, \delta \mathbf{x}_{t-1},\mathbf{u}_{t}, \mathbf{w}_{t})  \\  
		& \text{Error-state Observation Model: } &  \mathbf{z}_{t} = \mathbf{h}(\mathbf{x}_{t}, \delta \mathbf{x}_{t}, \mathbf{v}_{t})
	\end{aligned}
\end{equation}
- $\mathbf{x}_{t}$: model's nominal state variable\\
- $\delta \mathbf{x}_{t}$: model's error state variable\\
- $\mathbf{u}_{t}$: model's input\\
- $\mathbf{z}_{t}$: model's observation value\\
- $\mathbf{f}(\cdot)$: nonlinear motion model function\\
- $\mathbf{h}(\cdot)$: nonlinear observation model function\\
- $\mathbf{w}_{t} \sim \mathcal{N}(0, \mathbf{Q}_{t})$: noise in the error state model. $\mathbf{Q}_{t}$ represents the covariance matrix of $\mathbf{w}_{t}$\\
- $\mathbf{v}_{t} \sim \mathcal{N}(0, \mathbf{R}_{t})$: noise in the observation model. $\mathbf{R}_{t}$ is the covariance matrix of $\mathbf{v}_{t}$\\

\subsubsection{Commonality 2}
Both Jacobians $\mathbf{F}_{t}, \mathbf{H}_{t}$ of IESKF refer to the Jacobians for the error state $\delta \mathbf{x}_{t}$, similar to ESKF.
\begin{equation}   \boxed { \begin{aligned}
			& \mathbf{F}_{t} = \frac{\partial \mathbf{f}} {\partial \delta \mathbf{x}_{t-1}}\bigg|_{\substack{ \delta \mathbf{x}_{t-1} = \delta \hat{\mathbf{x}}_{t-1|t-1} }} & \mathbf{F}_{\mathbf{w}}  = \frac{\partial \mathbf{f}} {\partial \mathbf{w}_{t}}\bigg|_{\substack{ \delta \mathbf{x}_{t-1} = \delta \hat{\mathbf{x}}_{t-1|t-1} \\  \mathbf{w}_{t}=0}}
	\end{aligned} }
\end{equation}
\begin{equation}
	\boxed{ \begin{aligned}
			& \mathbf{H}_{t} = \frac{\partial \mathbf{h}}{\partial \delta \mathbf{x}_{t}}\bigg|_{\delta \mathbf{x}_{t}= \delta  \hat{\mathbf{x}}_{t|t-1}}  & \mathbf{H}_{\mathbf{v}} = \frac{\partial \mathbf{h}}{\partial \mathbf{v}_{t}}\bigg|_{ \substack{  \delta \mathbf{x}_{t}= \delta \hat{\mathbf{x}}_{t|t-1} \\  \mathbf{v}_{t}=0 }}
	\end{aligned} }
\end{equation}

$\mathbf{H}_{t}$ can be expressed through the chain rule as follows.
\begin{equation} 
	\begin{aligned} 
		\mathbf{H}_{t} = \frac{\partial \mathbf{h}}{\partial \delta \mathbf{x}_{t}} = \frac{\partial \mathbf{h}}{\partial \mathbf{x}_{\text{true},t}} \frac{\partial \mathbf{x}_{\text{true},t}}{\partial \delta \mathbf{x}_{t}} \\
	\end{aligned} 
\end{equation}

\subsubsection{Commonality 3}
The linearized error state variable $\hat{\mathbf{x}}_{t}$ can also be derived identically to ESKF.

\begin{equation} 
	\boxed { \begin{aligned} 
			\delta \mathbf{x}_{t} & = \mathbf{F}_{t} \delta \mathbf{x}_{t-1}  + \mathbf{F}_{\mathbf{w}} \mathbf{w}_{t} \\ 
			& = \mathbf{F}_{t} \delta \mathbf{x}_{t-1} +  \tilde{\mathbf{w}}_{t} \\ 
			& = 0 + \tilde{\mathbf{w}}_{t}
	\end{aligned}  }
\end{equation}
- $\tilde{\mathbf{w}}_{t} = \mathbf{F}_{\mathbf{w}} \mathbf{w}_{t} \sim \mathcal{N}(0, \mathbf{F}_{\mathbf{w}}\mathbf{Q}_{t} \mathbf{F}_{\mathbf{w}}^{\intercal})$\\

\subsubsection{Difference (MAP-based derivation)}
\textbfazure{ESKF derivation:}

The ESKF correction step unfolds the following equation to derive the mean $\delta \hat{\mathbf{x}}_{t|t}$ and covariance $\mathbf{P}_{t|t}$.

\begin{equation} \label{eq:ieskf1}
	\begin{aligned} 
		\text{bel}(\delta \mathbf{x}_{t}) = \eta \cdot p(\mathbf{z}_{t} | \delta  \mathbf{x}_{t}) \overline{\text{bel}}(\delta \mathbf{x}_{t}) \sim \mathcal{N}(\delta \hat{\mathbf{x}}_{t|t}, \mathbf{P}_{t|t})
	\end{aligned} 
\end{equation}

- $p(\mathbf{z}_{t} | \delta  \mathbf{x}_{t}) \sim \mathcal{N}(\tilde{\mathbf{z}}_{t}, \mathbf{H}_{\mathbf{v}}\mathbf{R}_{t}\mathbf{H}_{\mathbf{v}}^{\intercal})$ : (see \ref{eq:eskf2})

- $\overline{\text{bel}}(\delta \mathbf{x}_{t}) \sim \mathcal{N}(\delta \hat{\mathbf{x}}_{t|t-1}, \mathbf{P}_{t|t-1})$

\begin{equation} 
	\boxed{  \begin{aligned} 
			& \delta \hat{\mathbf{x}}_{t|t} = \mathbf{K}_{t} (\mathbf{z}_{t} - \mathbf{h}(\hat{\mathbf{x}}_{t|t-1}, 0,0)) \\
			& \mathbf{P}_{t|t} = (\mathbf{I} - \mathbf{K}_{t}\mathbf{H}_{t})\mathbf{P}_{t|t-1}
	\end{aligned} }
\end{equation}

\textbfazure{IESKF derivation:}

IESKF uses maximum a posteriori (MAP) for deriving the correction step. For MAP estimation, the Gauss-Newton optimization method is utilized. For more details, refer to section \ref{sec:map}. This derivation process is based on references \ref{ref:6}, \ref{ref:7}, \ref{ref:8}, \ref{ref:9}, \ref{ref:10}, and \ref{ref:11}, among which \ref{ref:6} provides the most detailed explanation of the IESKF derivation process.

If EKF is derived using the MAP approach, the final equation to optimize is as follows:

\begin{equation} \label{eq:ieskf2}
	\begin{aligned} 
		\arg\min_{\mathbf{x}_{t}} \quad \left\| \mathbf{z}_{t} - \mathbf{h}(\mathbf{x}_{\text{true}, t}, 0) \right\|_{\mathbf{H}_{\mathbf{v}}\mathbf{R}^{-1}\mathbf{H}_{\mathbf{v}}^{\intercal}} + \left\| \mathbf{x}_{\text{true},t} - \hat{\mathbf{x}}_{t|t-1} \right\|_{\mathbf{P}_{t|t-1}^{-1}} 
	\end{aligned} 
\end{equation}

When expressed in terms of the nominal state $\mathbf{x}_{t}$ and error state $\delta \mathbf{x}_{t}$, it becomes the following:

\begin{equation} \label{eq:ieskf7}
	\begin{aligned} 
		\arg\min_{\delta \hat{\mathbf{x}}_{t|t}} \quad \left\| \mathbf{z}_{t} - \mathbf{h}(\mathbf{x}_{t}, \delta \mathbf{x}_{t}, 0) \right\|_{\mathbf{H}_{\mathbf{v}}\mathbf{R}^{-1}\mathbf{H}_{\mathbf{v}}^{\intercal}} + \left\| \delta \hat{\mathbf{x}}_{t|t-1} \right\|_{\mathbf{P}_{t|t-1}^{-1}} 
	\end{aligned} 
\end{equation}

\mybox{Tip}{gray!40}{gray!10}{
	Prediction error state variable $\delta \hat{\mathbf{x}}_{t|t-1}$ is defined as follows:
	\begin{equation} \begin{aligned}  \delta \hat{\mathbf{x}}_{t|t-1} & = \mathbf{x}_{\text{true}, t} - \hat{\mathbf{x}}_{t|t-1} \\ & \sim \mathcal{N}(0, \mathbf{P}_{t|t-1})
	\end{aligned} \end{equation}
	Posterior (correction) error state variable $\delta \hat{\mathbf{x}}_{t|t}$ is defined as follows:
	\begin{equation} \begin{aligned} & \delta \hat{\mathbf{x}}_{t|t} = \mathbf{x}_{\text{true}, t|t} - \hat{\mathbf{x}}_{t|t} \end{aligned} \end{equation}
	If the equation is rearranged, the following formula is established:
	\begin{equation} \label{eq:ieskf3} \begin{aligned} & \mathbf{x}_{\text{true}, t|t} = \hat{\mathbf{x}}_{t|t} + \delta \hat{\mathbf{x}}_{t|t} \\ \end{aligned} \end{equation}
}

The first part of (\ref{eq:ieskf7}) should be linearized similarly to (\ref{eq:eskf-z}).

\begin{equation} \label{eq:ieskf5}
	\begin{aligned} 
		\mathbf{z}_{t} - \mathbf{h}(\mathbf{x}_{t|t-1}, \delta \hat{\mathbf{x}}_{t|t-1}, 0)  \approx \mathbf{z}_{t} -  \mathbf{h}(\mathbf{x}_{t|t-1}, 0, 0) - \mathbf{H}_{t} \delta \mathbf{x}_{t}
	\end{aligned} 
\end{equation}

In ESKF, $\delta \mathbf{x}_{t}$ is always 0, thus it was removed, but in IESKF it retains a non-zero value, so it is not removed. The next part of (\ref{eq:ieskf7}) with (\ref{eq:ieskf3}) substituted in unfolds as follows:

\begin{equation}  \label{eq:ieskf4}
	\begin{aligned} 
		\delta \hat{\mathbf{x}}_{t|t-1} & = \mathbf{x}_{\text{true},t} - \hat{\mathbf{x}}_{t|t-1} \\ &= (\hat{\mathbf{x}}_{t|t} + \delta \hat{\mathbf{x}}_{t|t}) - \hat{\mathbf{x}}_{t|t-1} \\ 
		& \approx \hat{\mathbf{x}}_{t|t} - \hat{\mathbf{x}}_{t|t-1} + \mathbf{J}_{t} \delta \hat{\mathbf{x}}_{t|t} \\ 
		& \sim \mathcal{N}(0, \mathbf{P}_{t|t-1})
	\end{aligned} 
\end{equation}

Here, $\mathbf{J}_{t}$ is defined as follows:

\begin{equation} 
	\begin{aligned} 
		\mathbf{J}_{t} = \frac{\partial }{\partial \delta \hat{\mathbf{x}}_{t|t}} \bigg( (\hat{\mathbf{x}}_{t|t} + \delta \hat{\mathbf{x}}_{t|t}) - \hat{\mathbf{x}}_{t|t-1} \bigg) \bigg|_{\delta \hat{\mathbf{x}}_{t|t}=0}
	\end{aligned} 
\end{equation}

If the third and fourth lines of (\ref{eq:ieskf4}) are rearranged and simplified, the following formula is obtained:

\begin{equation}  
	\begin{aligned} 
		\delta \hat{\mathbf{x}}_{t|t} \sim \mathcal{N}(-\mathbf{J}_{t}^{-1}(\hat{\mathbf{x}}_{t|t} - \hat{\mathbf{x}}_{t|t-1}), \ \mathbf{J}_{t}^{-1}\mathbf{P}_{t|t-1}\mathbf{J}_{t}^{-\intercal})
	\end{aligned} 
\end{equation}

Having calculated (\ref{eq:ieskf5}) and (\ref{eq:ieskf4}), the MAP estimation problem becomes as follows:

\begin{equation} 
	\begin{aligned} 
		\arg\min_{\delta \hat{\mathbf{x}}_{t|t}} \quad & \left\| \mathbf{z}_{t} - \mathbf{h}(\hat{\mathbf{x}}_{t|t-1}, 0, 0) - \mathbf{H}_{t} \delta\hat{\mathbf{x}}_{t|t} \right\|_{\mathbf{H}_{\mathbf{v}}\mathbf{R}^{-1}\mathbf{H}_{\mathbf{v}}^{\intercal}} \\
		& + \left\| \hat{\mathbf{x}}_{t|t} - \hat{\mathbf{x}}_{t|t-1} + \mathbf{J}_{t}\delta\hat{\mathbf{x}}_{t|t} \right\|_{\mathbf{P}_{t|t-1}^{-1}} 
	\end{aligned} 
\end{equation}

IESKF iteratively estimates the error state variable $\delta \hat{\mathbf{x}}_{t|t}$ until it converges to a value less than a specific threshold $\epsilon$. The expression for the $j$-th iteration is as follows:

\begin{equation}  
	\boxed{ \begin{aligned} 
			\delta \hat{\mathbf{x}}_{t|t,j} \sim \mathcal{N}(-\mathbf{J}_{t,j}^{-1}(\hat{\mathbf{x}}_{t|t,j} - \hat{\mathbf{x}}_{t|t-1}), \ \mathbf{J}_{t,j}^{-1}\mathbf{P}_{t|t-1}\mathbf{J}_{t,j}^{-\intercal})
	\end{aligned} }
\end{equation}

\begin{equation}  \label{eq:ieskf6}
	\boxed{ \begin{aligned} 
			\arg\min_{\delta \hat{\mathbf{x}}_{t|t,j}} \quad & \left\| \mathbf{z}_{t} - \mathbf{h}(\hat{\mathbf{x}}_{t|t-1}, 0, 0) - \mathbf{H}_{t} \delta\hat{\mathbf{x}}_{t|t,j} \right\|_{\mathbf{H}_{\mathbf{v}}\mathbf{R}^{-1}\mathbf{H}_{\mathbf{v}}^{\intercal}} \\ 
			& + \left\| \hat{\mathbf{x}}_{t|t,j} - \hat{\mathbf{x}}_{t|t-1} + \mathbf{J}_{t,j}\delta\hat{\mathbf{x}}_{t|t,j} \right\|_{\mathbf{P}_{t|t-1}^{-1}} 
	\end{aligned} }
\end{equation}

Refer to Section \ref{sec:deriv} for the derivation process of the update equation from the above equation.

\subsection{Prediction step}
Prediction involves the process of determining $\overline{\text{bel}}(\delta \mathbf{x}_{t})$. The linearized Jacobian matrix $\mathbf{F}_{t}$ is used when calculating the covariance matrix. This is completely identical to ESKF.
\begin{equation} 
	\boxed{  \begin{aligned}
			& \delta \hat{\mathbf{x}}_{t|t-1} = \mathbf{F}_{t}\delta \hat{\mathbf{x}}_{t-1|t-1} = 0 {\color{Mahogany} \quad \leftarrow \text{Always 0} } \\
			& \hat{\mathbf{x}}_{t|t-1} = \mathbf{f}(\hat{\mathbf{x}}_{t-1|t-1}, 0,\mathbf{u}_{t}, 0) \\ 
			& \mathbf{P}_{t|t-1} = \mathbf{F}_{t}\mathbf{P}_{t-1|t-1}\mathbf{F}_{t}^{\intercal} + \mathbf{F}_{\mathbf{w}}\mathbf{Q}_{t}  \mathbf{F}_{\mathbf{w}}^{\intercal}
	\end{aligned} }
\end{equation}

\subsection{Correction step}
Correction involves the process of determining $\text{bel}(\mathbf{x}_{t})$. The Kalman gain and covariance matrix are calculated using the linearized Jacobian matrix $\mathbf{H}_{t}$. The correction step of IESKF is performed iteratively until the update amount $\delta \hat{\mathbf{x}}_{t|t,j}$ becomes sufficiently small. The derived formula (\ref{eq:ieskf6}) is differentiated and set to zero as follows: 

\begin{equation}  \label{eq:ieskf8}
	\boxed{ \begin{aligned}
			& \text{set } \epsilon \\
			&  \text{start j-th loop }  \\
			& \quad \mathbf{S}_{t,j} = \mathbf{H}_{t,j}\mathbf{J}^{-1}_{t,j} \mathbf{P}_{t|t-1} \mathbf{J}^{-\intercal}_{t,j}\mathbf{H}_{t,j}^{\intercal} + \mathbf{H}_{\mathbf{v},j}\mathbf{R}_{t}\mathbf{H}_{\mathbf{v},j}^{\intercal} \\ 
			&  \quad   \mathbf{K}_{t,j} = \mathbf{J}^{-1}_{t,j} \mathbf{P}_{t|t-1} \mathbf{J}^{-\intercal}_{t,j}\mathbf{H}_{t,j}^{\intercal}\mathbf{S}_{t,j}^{-1} \\
			& {\color{Mahogany} \quad \delta \hat{\mathbf{x}}_{t|t,j} = \mathbf{K}_{t,j} \bigg(  \mathbf{z}_{t} - \mathbf{h}(\hat{\mathbf{x}}_{t|t,j}, 0 ,0) + \mathbf{H}_{t,j}\mathbf{J}^{-1}_{t,j}(\hat{\mathbf{x}}_{t|t,j} - \hat{\mathbf{x}}_{t|t-1}) \bigg) - \mathbf{J}^{-1}_{t,j}(\hat{\mathbf{x}}_{t|t,j} - \hat{\mathbf{x}}_{t|t-1}) }\\
			& \quad \hat{\mathbf{x}}_{t|t, j+1}  = \hat{\mathbf{x}}_{t|t, j} \boxplus \delta \hat{\mathbf{x}}_{t|t,j} \\
			& \quad \text{iterate until } \delta \hat{\mathbf{x}}_{t|t,j} < \epsilon.   \\ 
			& \text{end loop} \\ 
			&     \mathbf{P}_{t|t,n}    = (\mathbf{I} - \mathbf{K}_{t,n}\mathbf{H}_{t,n}) \mathbf{J}^{-1}_{t,n}\mathbf{P}_{t|t-1} \mathbf{J}_{t,n}^{-\intercal} \\
	\end{aligned} }
\end{equation}

For a more detailed derivation process, refer to Section \ref{sec:deriv}.

\subsection{Summary}
IESKF can be expressed as a function as follows.
	
	\begin{equation} 
		\boxed{ \begin{aligned} 
				& \text{IteratedErrorStateKalmanFilter}(\hat{\mathbf{x}}_{t-1|t-1}, \delta \hat{\mathbf{x}}_{t-1|t-1}, \mathbf{P}_{t-1|t-1}, \mathbf{u}_{t}, \mathbf{z}_{t}) \{ \\ 
				& \quad\quad \text{(Prediction Step)}\\  
				& \quad\quad \delta \hat{\mathbf{x}}_{t|t-1} = \mathbf{F}_{t}\hat{\mathbf{x}}_{t-1|t-1}= 0 {\color{Mahogany} \quad \leftarrow \text{Always 0} } \\  
				& \quad\quad  \hat{\mathbf{x}}_{t|t-1} = \mathbf{f}(\hat{\mathbf{x}}_{t-1|t-1}, 0,\mathbf{u}_{t}, 0) \\ 
				& \quad\quad \mathbf{P}_{t|t-1} = \mathbf{F}_{t}\mathbf{P}_{t-1|t-1}\mathbf{F}_{t}^{\intercal} + \mathbf{F}_{\mathbf{w}}\mathbf{Q}_{t} \mathbf{F}_{\mathbf{w}}^{\intercal}\\  
				& \\  
				& \quad\quad \text{(Correction Step)} \\ 
				& \quad\quad \text{set } \epsilon \\
				& \quad\quad  \text{start j-th loop }  \\
				& \quad\quad \quad \mathbf{S}_{t,j} = \mathbf{H}_{t,j}\mathbf{J}^{-1}_{t,j} \mathbf{P}_{t|t-1} \mathbf{J}^{-\intercal}_{t,j}\mathbf{H}_{t,j}^{\intercal} + \mathbf{H}_{\mathbf{v},j}\mathbf{R}_{t}\mathbf{H}_{\mathbf{v},j}^{\intercal} \\ 
				&  \quad\quad \quad   \mathbf{K}_{t,j} = \mathbf{J}^{-1}_{t,j} \mathbf{P}_{t|t-1} \mathbf{J}^{-\intercal}_{t,j}\mathbf{H}_{t,j}^{\intercal}\mathbf{S}_{t,j}^{-1} \\
				& \quad\quad \quad \delta \hat{\mathbf{x}}_{t|t,j} = \mathbf{K}_{t,j} \bigg(  \mathbf{z}_{t} - \mathbf{h}(\hat{\mathbf{x}}_{t|t,j}, 0 ,0) + \mathbf{H}_{t,j}\mathbf{J}^{-1}_{t,j}(\hat{\mathbf{x}}_{t|t,j} - \hat{\mathbf{x}}_{t|t-1}) \bigg) - \mathbf{J}^{-1}_{t,j}(\hat{\mathbf{x}}_{t|t,j} - \hat{\mathbf{x}}_{t|t-1})\\
				& \quad\quad \quad \hat{\mathbf{x}}_{t|t, j+1}  = \hat{\mathbf{x}}_{t|t, j} \boxplus \delta \hat{\mathbf{x}}_{t|t,j} \\
				& \quad\quad \quad \text{iterate until } \delta \hat{\mathbf{x}}_{t|t,j} < \epsilon.   \\ 
				& \quad\quad \text{end loop} \\ 
				& \quad\quad     \mathbf{P}_{t|t,n}    = (\mathbf{I} - \mathbf{K}_{t,n}\mathbf{H}_{t,n}) \mathbf{J}^{-1}_{t,n}\mathbf{P}_{t|t-1} \mathbf{J}_{t,n}^{-\intercal} \\ 
				& \quad\quad \text{return} \ \ \hat{\mathbf{x}}_{t|t}, \mathbf{P}_{t|t} \\ 
				&   \} 
		\end{aligned} }
	\end{equation}

\section{Derivation of Kalman filter} \label{sec:derivkf}
In this section, we explain the derivation of the equations for the prediction and update steps of the Kalman filter. Most of the content is based on Section 3.2.4 "Mathematical Derivation of the KF" from \ref{ref:13}.

The Kalman filter assumes that $\text{bel}(\mathbf{x}_{t})$ and $\overline{\text{bel}}(\mathbf{x}_{t})$ both follow Gaussian distributions, allowing us to compute the mean $\hat{\mathbf{x}}_t$ and the covariance $\mathbf{P}$ for each.
\begin{equation} 
	\begin{aligned} 
		& \overline{\text{bel}}(\mathbf{x}_{t}) = \int {\color{Red} p(\mathbf{x}_{t} \ | \ \mathbf{x}_{t-1}, \mathbf{u}_{t}) } \text{bel}(\mathbf{x}_{t-1}) d \mathbf{x}_{t-1} \sim \mathcal{N}(\hat{\mathbf{x}}_{t|t-1}, \mathbf{P}_{t|t-1}) \\ 
		& \text{bel}(\mathbf{x}_{t}) = \eta \cdot {\color{Cyan} p( \mathbf{z}_{t} \ | \  \mathbf{x}_{t}) } \overline{\text{bel}}(\mathbf{x}_{t}) \sim \mathcal{N}(\hat{\mathbf{x}}_{t|t}, \mathbf{P}_{t|t})
	\end{aligned} 
\end{equation}

From the motion model and observation model in (\ref{eq:kf0}), we can express ${\color{Red} p(\mathbf{x}_{t} \ | \ \mathbf{x}_{t-1}, \mathbf{u}_{t}) }$ and ${\color{Cyan} p(\mathbf{z}_{t} \ | \ \mathbf{x}_{t})}$ as shown in (\ref{eq:kf1}) and (\ref{eq:kf2}).
\begin{equation} 
	\begin{aligned} 
		& {\color{Red} p(\mathbf{x}_{t} \ | \ \mathbf{x}_{t-1}, \mathbf{u}_{t}) }&& \sim \mathcal{N}(\mathbf{F}_{t}\mathbf{x}_{t-1} + \mathbf{B}_{t}\mathbf{u}_{t}, \mathbf{Q}_{t}) \\ 
		& && = \frac{1}{ \sqrt{\text{det}(2\pi \mathbf{Q}_t)}}\exp\bigg( -\frac{1}{2}(\mathbf{x}_{t} -\mathbf{F}_{t}\mathbf{x}_{t-1}-\mathbf{B}_{t}\mathbf{u}_t)^{\intercal}\mathbf{Q}_t^{-1}(\mathbf{x}_{t}-\mathbf{F}_{t}\mathbf{x}_{t-1}-\mathbf{B}_{t}\mathbf{u}_{t}) \bigg)\\  
	\end{aligned} 
\end{equation}
\begin{equation} 
	\begin{aligned} 
		& {\color{Cyan} p(\mathbf{z}_{t} \ | \ \mathbf{x}_{t}) } && \sim \mathcal{N}(\mathbf{H}_{t}\mathbf{x}_{t}, \mathbf{R}_{t}) \\ 
		& && = \frac{1}{ \sqrt{\text{det}(2\pi \mathbf{R}_t)}}\exp\bigg( -\frac{1}{2}(\mathbf{z}_{t}-\mathbf{H}_{t}\mathbf{x}_{t})^{\intercal}\mathbf{R}_{t}^{-1}(\mathbf{z}_{t}-\mathbf{H}_{t}\mathbf{x}_{t}) \bigg) 
	\end{aligned} 
\end{equation}

\subsection{Derivation of KF prediction step}
First, let's derive the equation for $\overline{\text{bel}}(\mathbf{x}_{t})$. The mean and variance we aim to compute for $\overline{\text{bel}}(\mathbf{x}_{t})$ are as follows:
\begin{equation} 
	\begin{aligned} 
		& \underbrace{\overline{\text{bel}}(\mathbf{x}_{t})}_{\sim \mathcal{N}(\hat{\mathbf{x}}_{t|t-1}, \mathbf{P}_{t|t-1})  }  = \int \underbrace{  p(\mathbf{x}_{t} \ | \ \mathbf{x}_{t-1}, \mathbf{u}_{t})  }_{\sim \mathcal{N}(\mathbf{F}_t\mathbf{x}_{t-1}+\mathbf{B}_t \mathbf{u}_t, \mathbf{R}_{t})} \underbrace{\text{bel}(\mathbf{x}_{t-1})}_{\quad \sim \mathcal{N}(\hat{\mathbf{x}}_{t-1}, \mathbf{P}_{t-1})} d \mathbf{x}_{t-1}
	\end{aligned} 
\end{equation}

Expanding the above equation in Gaussian form, we get:
\begin{equation} 
	\begin{aligned} 
		& \overline{\text{bel}}(\mathbf{x}_{t}) = \eta \int \exp \Big(-\frac{1}{2}(\mathbf{x}_t - \mathbf{F}_t\mathbf{x}_{t-1}-\mathbf{B}_t\mathbf{u}_t)^{\intercal} \mathbf{R}_t^{-1}(\mathbf{x}_t - \mathbf{F}_t\mathbf{x}_{t-1}-\mathbf{B}_t\mathbf{u}_t) \Big) \\
		& \qquad\qquad\qquad\qquad \cdot \exp \Big( -\frac{1}{2}(\mathbf{x}_{t-1} - \hat{\mathbf{x}}_{t-1})^{\intercal} \mathbf{P}_{t-1}^{-1}(\mathbf{x}_{t-1} - \hat{\mathbf{x}}_{t-1}) \Big) d \mathbf{x}_{t-1}
	\end{aligned} 
\end{equation}

This equation can be simplified as follows:
\begin{equation} 
	\begin{aligned} 
		& \overline{\text{bel}}(\mathbf{x}_{t}) = \eta \int \exp(- {\color{Mahogany} \mathbf{L}_{t} } ) d \mathbf{x}_{t-1}
	\end{aligned} 
\end{equation}
\begin{equation}  \label{eq:derivkf1}
	\boxed{ \begin{aligned} 
			& \text{where, } {\color{Mahogany} \mathbf{L}_t }= \frac{1}{2}(\mathbf{x}_t - \mathbf{F}_t\mathbf{x}_{t-1}-\mathbf{B}_t\mathbf{u}_t)^{\intercal} \mathbf{R}_t^{-1}(\mathbf{x}_t - \mathbf{F}_t\mathbf{x}_{t-1}-\mathbf{B}_t\mathbf{u}_t) \\ 
			& \qquad\qquad\qquad + \frac{1}{2}(\mathbf{x}_{t-1} - \hat{\mathbf{x}}_{t-1})^{\intercal} \mathbf{P}_{t-1}^{-1}(\mathbf{x}_{t-1} - \hat{\mathbf{x}}_{t-1})
	\end{aligned}  }
\end{equation}

Examining $\mathbf{L}_t$, we can see it has a quadratic form in terms of $\mathbf{x}_t$ and $\mathbf{x}_{t-1}$. Since $\overline{\text{bel}}(\mathbf{x}_{t})$ includes an integral $\int$, we need to factor out $\mathbf{L}_t$ from the integral to obtain a closed-form solution. To do this, we separate $\mathbf{L}_t$ into $\mathbf{L}_t(\mathbf{x}_{t-1}, \mathbf{x}_t)$ and $\mathbf{L}_t(\mathbf{x}_t)$ terms.
\begin{equation}
	\begin{aligned}
		\mathbf{L}_t = \mathbf{L}_t(\mathbf{x}_{t-1}, \mathbf{x}_t) + \mathbf{L}_t(\mathbf{x}_t)
	\end{aligned}
\end{equation}

$\mathbf{L}_t(\mathbf{x}_{t-1}, \mathbf{x}_t)$ includes all terms containing $\mathbf{x}_{t-1}$, while $\mathbf{L}_t(\mathbf{x}_t)$ includes only terms with $\mathbf{x}_t$. This allows $\mathbf{L}_t(\mathbf{x}_t)$ to be independent of $\mathbf{x}_{t-1}$ and factored out of the integral.
\begin{equation}  \label{eq:derivkf4}
	\begin{aligned} 
		\overline{\text{bel}}(\mathbf{x}_{t}) & = \eta \int \exp(-\mathbf{L}_{t}) d \mathbf{x}_{t-1}  \\
		& = \eta \int \exp(-\mathbf{L}_t(\mathbf{x}_{t-1}, \mathbf{x}_t) - {\color{Mahogany}  \mathbf{L}_t(\mathbf{x}_t) } ) d \mathbf{x}_{t-1} \\
		& = \eta \exp(-{\color{Mahogany} \mathbf{L}_t(\mathbf{x}_t) } ) \int \exp(-\mathbf{L}_t(\mathbf{x}_{t-1}, \mathbf{x}_t)) d \mathbf{x}_{t-1}
	\end{aligned} 
\end{equation}

Next, we set $\mathbf{L}_t(\mathbf{x}_{t-1}, \mathbf{x}_t)$ to be independent of $\mathbf{x}_t$ so that the integral inside becomes a constant. To decompose $\mathbf{L}_t(\mathbf{x}_{t-1}, \mathbf{x}_t)$, we first take the partial derivative of $\mathbf{L}_t$ with respect to $\mathbf{x}_{t-1}$.
\begin{equation}  \label{eq:derivkf2}
	\begin{aligned} 
		\frac{\partial \mathbf{L}_t}{\partial \mathbf{x}_{t-1}} = -\mathbf{F}_t^{\intercal}\mathbf{R}_t^{-1}(\mathbf{x}_t - \mathbf{F}_t\mathbf{x}_{t-1} - \mathbf{B}_t \mathbf{u}_t) + \mathbf{P}_{t-1}^{-1}(\mathbf{x}_{t-1} - \hat{\mathbf{x}}_{t-1})
	\end{aligned} 
\end{equation}
\begin{equation}   \label{eq:derivkf3}
	\begin{aligned} 
		\frac{\partial^2 \mathbf{L}_t}{\partial \mathbf{x}^2_{t-1}} = \mathbf{F}^{\intercal}_t\mathbf{R}_t^{-1} \mathbf{F}_t+ \mathbf{P}_{t-1}^{-1} := \Psi_t^{-1}
	\end{aligned} 
\end{equation}
- $\Psi_t^{-1}$: represents the curvature of $\mathbf{L}_t$ and is the inverse of the covariance.\\

Since $\mathbf{L}_t$ is quadratic and positive semi-definite due to the covariance matrix, the value where the first derivative with respect to $\mathbf{x}_{t-1}$ is zero is the mean for $\mathbf{x}_{t-1}$, and the second derivative is the inverse covariance matrix.
\begin{figure}[h!]
	\centering
	\includegraphics[width=6cm]{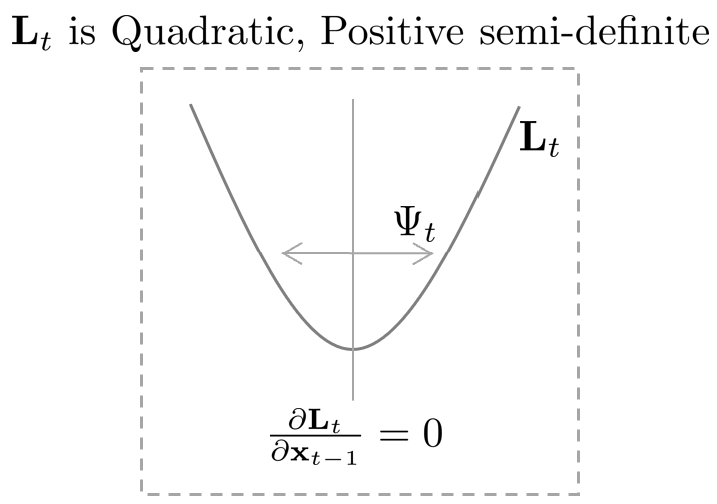}
\end{figure}

Setting the first derivative (\ref{eq:derivkf2}) to zero and solving for ${\color{Cyan} \mathbf{x}_{t-1} }$, we get:
\begin{equation} 
	\begin{aligned} 
		\mathbf{F}_t^{\intercal}\mathbf{R}_t^{-1}(\mathbf{x}_t - \mathbf{F}_t {\color{Cyan} \mathbf{x}_{t-1} } - \mathbf{B}_t \mathbf{u}_t) = \mathbf{P}_{t-1}^{-1}({\color{Cyan} \mathbf{x}_{t-1} } - \hat{\mathbf{x}}_{t-1})
	\end{aligned} 
\end{equation}

Solving for ${\color{Cyan} \mathbf{x}_{t-1} }$, we get:
\begin{equation}
	\begin{aligned}
		& \Leftrightarrow \mathbf{F}_t^{\intercal}\mathbf{R}_t^{-1}(\mathbf{x}_t - \mathbf{B}_t \mathbf{u}_t) - \mathbf{F}_t^{\intercal}\mathbf{R}_t^{-1}\mathbf{F}_t {\color{Cyan} \mathbf{x}_{t-1} } = \mathbf{P}_{t-1}^{-1}{\color{Cyan} \mathbf{x}_{t-1} } - \mathbf{P}_{t-1}^{-1}\hat{\mathbf{x}}_{t-1} \\
		& \Leftrightarrow \mathbf{F}_t^{\intercal}\mathbf{R}_t^{-1}\mathbf{F}_t {\color{Cyan} \mathbf{x}_{t-1} }+ \mathbf{P}_{t-1}^{-1}{\color{Cyan} \mathbf{x}_{t-1} } = \mathbf{F}_t^{\intercal}\mathbf{R}_t^{-1}(\mathbf{x}_t - \mathbf{B}_t \mathbf{u}_t) + \mathbf{P}_{t-1}^{-1}\hat{\mathbf{x}}_{t-1} \\
		& \Leftrightarrow (\mathbf{F}_t^{\intercal}\mathbf{R}_t^{-1}\mathbf{F}_t + \mathbf{P}_{t-1}^{-1}){\color{Cyan} \mathbf{x}_{t-1} } = \mathbf{F}_t^{\intercal}\mathbf{R}_t^{-1}(\mathbf{x}_t - \mathbf{B}_t \mathbf{u}_t) + \mathbf{P}_{t-1}^{-1}\hat{\mathbf{x}}_{t-1}\\
		& \Leftrightarrow \Psi_t^{-1}{\color{Cyan} \mathbf{x}_{t-1} } = \mathbf{F}_t^{\intercal}\mathbf{R}_t^{-1}(\mathbf{x}_t - \mathbf{B}_t \mathbf{u}_t) + \mathbf{P}_{t-1}^{-1}\hat{\mathbf{x}}_{t-1} \\
		& \Leftrightarrow  {\color{Cyan} \mathbf{x}_{t-1} } =   \Psi_t[ \mathbf{F}_t^{\intercal}\mathbf{R}_t^{-1}(\mathbf{x}_t - \mathbf{B}_t \mathbf{u}_t) + \mathbf{P}_{t-1}^{-1}\hat{\mathbf{x}}_{t-1} ] 
	\end{aligned}
\end{equation}

Using the obtained ${\color{Cyan} \mathbf{x}_{t-1}}$, we can decompose $\mathbf{L}_t$ in terms of $\mathbf{x}_{t-1}$ and $\mathbf{x}_t$ as follows:
\begin{equation} 
	\boxed{ \begin{aligned} 
			&\mathbf{L}_t(\mathbf{x}_{t-1}, \mathbf{x}_t) = \frac{1}{2}(\mathbf{x}_{t-1} - { \Psi_t[ \mathbf{F}_t^{\intercal}\mathbf{R}_t^{-1}(\mathbf{x}_t - \mathbf{B}_t \mathbf{u}_t) + \mathbf{P}_{t-1}^{-1}\hat{\mathbf{x}}_{t-1} ] })^{\intercal} \Psi^{-1} \\
			&\qquad\qquad\qquad\qquad (\mathbf{x}_{t-1} - { \Psi_t[ \mathbf{F}_t^{\intercal}\mathbf{R}_t^{-1}(\mathbf{x}_t - \mathbf{B}_t \mathbf{u}_t) + \mathbf{P}_{t-1}^{-1}\hat{\mathbf{x}}_{t-1} ] } )
	\end{aligned}  }
\end{equation}

Note that this is only one way to decompose $\mathbf{L}_t(\mathbf{x}_{t-1}, \mathbf{x}_t)$. By putting $\mathbf{L}_t(\mathbf{x}_{t-1}, \mathbf{x}_t)$ inside the $\exp(\cdot)$, it can be defined as a Gaussian distribution.
\begin{equation} 
	\begin{aligned} 
		\det(2\pi \Psi)^{-\frac{1}{2}} \exp(-\mathbf{L}_t(\mathbf{x}_{t-1}, \mathbf{x}_t))
	\end{aligned} 
\end{equation}

Since it is a Gaussian distribution, the sum of the area over all regions is 1.
\begin{equation} 
	\begin{aligned} 
		\int \det(2\pi \Psi)^{-\frac{1}{2}} \exp(-\mathbf{L}_t(\mathbf{x}_{t-1}, \mathbf{x}_t)) d\mathbf{x}_{t-1} = 1
	\end{aligned} 
\end{equation}

Simplifying the above equation, we get:
\begin{equation} 
	\begin{aligned} 
		{\color{Mahogany}  \int  \exp(-\mathbf{L}_t(\mathbf{x}_{t-1}, \mathbf{x}_t)) d\mathbf{x}_{t-1} } = \det(2\pi \Psi)^{\frac{1}{2}}
	\end{aligned} 
\end{equation}

This confirms that the integral term in (\ref{eq:derivkf4}) becomes a constant.
\begin{equation} 
	\begin{aligned} 
		\overline{\text{bel}}(\mathbf{x}_{t}) & = \eta \int \exp(-\mathbf{L}_{t}) d \mathbf{x}_{t-1}  \\
		& = \eta \int \exp(-\mathbf{L}_t(\mathbf{x}_{t-1}, \mathbf{x}_t) - \mathbf{L}_t(\mathbf{x}_t)) d \mathbf{x}_{t-1} \\
		& = \eta \exp(-\mathbf{L}_t(\mathbf{x}_t)) {\color{Mahogany} \int \exp(-\mathbf{L}_t(\mathbf{x}_{t-1}, \mathbf{x}_t)) d \mathbf{x}_{t-1} } \\
		& = \eta' \exp(-\mathbf{L}_t(\mathbf{x}_t))
	\end{aligned} 
\end{equation}

Although $\overline{\text{bel}}(\mathbf{x}_{t})$ has been simplified, we still need to derive the exact expression for $\mathbf{L}_t(\mathbf{x}_t)$. This can be obtained by subtracting the decomposed part from the total $\mathbf{L}_t$.
\begin{equation}
	\begin{aligned}
		\mathbf{L}_t(\mathbf{x}_t) & = {\color{Red}  \mathbf{L}_t }- {\color{Cyan} \mathbf{L}_t(\mathbf{x}_{t-1}, \mathbf{x}_t) } \\
		& = {\color{Red}  \frac{1}{2}(\mathbf{x}_t - \mathbf{F}_t\mathbf{x}_{t-1}-\mathbf{B}_t\mathbf{u}_t)^{\intercal} \mathbf{R}_t^{-1}(\mathbf{x}_t - \mathbf{F}_t\mathbf{x}_{t-1}-\mathbf{B}_t\mathbf{u}_t) } \\
		& \quad{\color{Red}  +  \frac{1}{2}(\mathbf{x}_{t-1} - \hat{\mathbf{x}}_{t-1})^{\intercal} \mathbf{P}_{t-1}^{-1}(\mathbf{x}_{t-1} - \hat{\mathbf{x}}_{t-1}) } \\
		& \quad - {\color{Cyan}  \frac{1}{2}(\mathbf{x}_{t-1} - \Psi_t[ \mathbf{F}_t^{\intercal}\mathbf{R}_t^{-1}(\mathbf{x}_t - \mathbf{B}_t \mathbf{u}_t) + \mathbf{P}_{t-1}^{-1}\hat{\mathbf{x}}_{t-1} ])^{\intercal} \Psi^{-1} } \\
		& \quad {\color{Cyan} (\mathbf{x}_{t-1} - \Psi_t[ \mathbf{F}_t^{\intercal}\mathbf{R}_t^{-1}(\mathbf{x}_t - \mathbf{B}_t \mathbf{u}_t) + \mathbf{P}_{t-1}^{-1}\hat{\mathbf{x}}_{t-1} ]) }
	\end{aligned}
\end{equation}

Expanding the equation to obtain the exact expression for $\mathbf{L}_t(\mathbf{x}_t)$ and restoring the symbol $\Psi = (\mathbf{F}^{\intercal}_t\mathbf{R}_t^{-1} \mathbf{F}_t+ \mathbf{P}_{t-1}^{-1})^{-1}$, we get:
\begin{equation}
	\begin{aligned}
		\mathbf{L}_t(\mathbf{x}_t) & = \mathbf{L}_t - \mathbf{L}_t(\mathbf{x}_{t-1}, \mathbf{x}_t) \\
		& = \underline{\underline{\frac{1}{2}\mathbf{x}_{t-1}^{\intercal}\mathbf{F}_t^{\intercal}\mathbf{R}_t^{-1}\mathbf{F}_t\mathbf{x}_{t-1} }} - \underline{ \mathbf{x}_{t-1}^{\intercal}\mathbf{F}_t^{\intercal}\mathbf{R}_t^{-1}(\mathbf{x}_t - \mathbf{B}_t \mathbf{u}_t) } \\
		& \quad + \frac{1}{2}(\mathbf{x}_t - \mathbf{B}_t\mathbf{u}_t)^{\intercal}\mathbf{R}_t^{-1}(\mathbf{x}_t - \mathbf{B}_t \mathbf{u}_t) \\
		& \quad + \underline{\underline{ \frac{1}{2} \mathbf{x}_{t-1}^{\intercal}\mathbf{P}_{t-1}^{-1}\mathbf{x}_{t-1} }} - \underline{ \mathbf{x}_{t-1}^{\intercal} \mathbf{P}_{t-1}^{-1}\hat{\mathbf{x}}_{t-1} } + \frac{1}{2}\hat{\mathbf{x}}_{t-1}^{\intercal} \mathbf{P}_{t-1}^{-1} \hat{\mathbf{x}}_{t-1} \\
		& \quad - \underline{\underline{ \frac{1}{2}\mathbf{x}_{t-1}^{\intercal} (\mathbf{F}_t^{\intercal}\mathbf{R}_t^{-1}\mathbf{F}_t + \mathbf{P}_{t-1}^{-1})\mathbf{x}_{t-1} }} \\
		& \quad + \underline{ \mathbf{x}_{t-1}^{\intercal}[\mathbf{F}_t^{\intercal}\mathbf{R}_t^{-1}(\mathbf{x}_t - \mathbf{B}_t \mathbf{u}_t) + \mathbf{P}_{t-1}^{-1}\hat{\mathbf{x}}_{t-1}] } \\
		& \quad -\frac{1}{2} [ \mathbf{F}_t^{\intercal}\mathbf{R}_t^{-1}(\mathbf{x}_t - \mathbf{B}_t \mathbf{u}_t) + \mathbf{P}_{t-1}^{-1}\hat{\mathbf{x}}_{t-1}]^{\intercal} (\mathbf{F}_t^{\intercal} \mathbf{R}_t^{-1}\mathbf{F}_t + \mathbf{P}_{t-1}^{-1})^{-1} \\
		& \quad [ \mathbf{F}_t^{\intercal}\mathbf{R}_t^{-1}(\mathbf{x}_t - \mathbf{B}_t \mathbf{u}_t) + \mathbf{P}_{t-1}^{-1}\hat{\mathbf{x}}_{t-1}]
	\end{aligned}
\end{equation}
- $\underline{\underline{(\cdot)}}$: quadratic terms in $\mathbf{x}_{t-1}$\\
- $\underline{(\cdot)}$: linear terms in $\mathbf{x}_{t-1}$\\

The terms involving $\mathbf{x}_{t-1}$ are cancelled out, as we have already excluded them in $\mathbf{L}_t(\mathbf{x}_{t-1}, \mathbf{x}_t)$. Removing the cancelled terms, we get $\mathbf{L}_t(\mathbf{x}_t)$.
\begin{equation}
	\boxed{	\begin{aligned}
			\mathbf{L}_t(\mathbf{x}_t) & = \frac{1}{2}(\mathbf{x}_t - \mathbf{B}_t\mathbf{u}_t)^{\intercal}\mathbf{R}_t^{-1}(\mathbf{x}_t - \mathbf{B}_t \mathbf{u}_t) + \frac{1}{2}\hat{\mathbf{x}}_{t-1}^{\intercal} \mathbf{P}_{t-1}^{-1} \hat{\mathbf{x}}_{t-1} \\
			& \quad - \frac{1}{2} [ \mathbf{F}_t^{\intercal}\mathbf{R}_t^{-1}(\mathbf{x}_t - \mathbf{B}_t \mathbf{u}_t) + \mathbf{P}_{t-1}^{-1}\hat{\mathbf{x}}_{t-1}]^{\intercal} (\mathbf{F}_t^{\intercal} \mathbf{R}_t^{-1}\mathbf{F}_t + \mathbf{P}_{t-1}^{-1})^{-1} \\
			& \quad [ \mathbf{F}_t^{\intercal}\mathbf{R}_t^{-1}(\mathbf{x}_t - \mathbf{B}_t \mathbf{u}_t) + \mathbf{P}_{t-1}^{-1}\hat{\mathbf{x}}_{t-1}]
	\end{aligned} }
\end{equation}

Since $\overline{\text{bel}}(\mathbf{x}_{t})$ follows a Gaussian distribution, $\mathbf{L}_t(\mathbf{x}_t)$ also has a quadratic form. Similar to $\mathbf{L}_t(\mathbf{x}_{t-1}, \mathbf{x}_t)$, we find the mean and covariance by taking the derivatives.

First, we find the mean of $\mathbf{x}_t$ by taking the first derivative. Using the matrix inversion lemma, we simplify the equation:
\begin{equation}
	\begin{aligned}
		\frac{\partial \mathbf{L}_t(\mathbf{x}_t)}{\partial \mathbf{x}_t} & = \mathbf{R}_t^{-1}(\mathbf{x}_t - \mathbf{B}_t \mathbf{u}_t) - \mathbf{R}_t^{-1}\mathbf{F}_t(\mathbf{F}_t^{\intercal}\mathbf{R}_t^{-1}\mathbf{F}_t + \mathbf{P}_{t-1}^{-1})^{-1} [\mathbf{F}_t^{\intercal}\mathbf{R}_t^{-1}(\mathbf{x}_t - \mathbf{B}_t\mathbf{u}_t) + \mathbf{P}_{t-1}^{-1}\hat{\mathbf{x}}_{t-1}] \\
		& = {\color{Cyan} [\mathbf{R}_t^{-1} - \mathbf{R}_t^{-1}\mathbf{F}_t(\mathbf{F}_t^{\intercal}\mathbf{R}_t^{-1}\mathbf{F}_t + \mathbf{P}_{t-1}^{-1})^{-1} \mathbf{F}_t^{\intercal}\mathbf{R}_t^{-1}] } (\mathbf{x}_t - \mathbf{B}_t\mathbf{u}_t) - \mathbf{R}_t^{-1}\mathbf{F}_t(\mathbf{F}_t^{\intercal}\mathbf{R}_t^{-1}\mathbf{F}_t + \mathbf{P}_{t-1}^{-1})^{-1}\mathbf{P}_{t-1}^{-1}\hat{\mathbf{x}}_{t-1} \\
		& = {\color{Cyan} (\mathbf{R}_t + \mathbf{F}_t\mathbf{P}_{t-1}\mathbf{F}_t^{\intercal})^{-1} } (\mathbf{x}_t - \mathbf{B}_t \mathbf{u}_t) - \mathbf{R}_t^{-1}\mathbf{F}_t(\mathbf{F}_t^{\intercal}\mathbf{R}_t^{-1}\mathbf{F}_t + \mathbf{P}_{t-1}^{-1})^{-1}\mathbf{P}_{t-1}^{-1}\hat{\mathbf{x}}_{t-1}
	\end{aligned}
\end{equation}

Setting $\frac{\partial \mathbf{L}_t(\mathbf{x}_t)}{\partial \mathbf{x}_t}=0$, we find the mean:
\begin{equation} 
	\begin{aligned} 
		{\color{Cyan} (\mathbf{R}_t + \mathbf{F}_t\mathbf{P}_{t-1}\mathbf{F}_t^{\intercal})^{-1}  }(\mathbf{x}_t - \mathbf{B}_t \mathbf{u}_t) =  \mathbf{R}_t^{-1}\mathbf{F}_t(\mathbf{F}_t^{\intercal}\mathbf{R}_t^{-1}\mathbf{F}_t + \mathbf{P}_{t-1}^{-1})^{-1}\mathbf{P}_{t-1}^{-1}\hat{\mathbf{x}}_{t-1}
	\end{aligned} 
\end{equation}

Solving for $\mathbf{x}_t$, we get a simplified equation:
\begin{equation}
	\begin{aligned}
		\mathbf{x}_t & = \mathbf{B}_t \mathbf{u}_t + \underbrace{ (\mathbf{R}_t + \mathbf{F}_t\mathbf{P}_{t-1}\mathbf{F}_t^{\intercal}) \mathbf{R}_t^{-1}\mathbf{F}_t }_{\mathbf{F}_t + \mathbf{F}_t\mathbf{P}_{t-1}\mathbf{F}_t^{\intercal}\mathbf{R}_t^{-1}\mathbf{F}_t} \underbrace{ (\mathbf{F}_t^{\intercal}\mathbf{R}_t^{-1}\mathbf{F}_t + \mathbf{P}_{t-1}^{-1})^{-1} \mathbf{P}_{t-1}^{-1}}_{(\mathbf{P}_{t-1}\mathbf{F}_t^{\intercal}\mathbf{R}_t^{-1}\mathbf{F}_t + \mathbf{I})^{-1}} \hat{\mathbf{x}}_{t-1} \\
		& = \mathbf{B}_t \mathbf{u}_t + \mathbf{F}_t \underbrace{ (\mathbf{I} + \mathbf{P}_{t-1}\mathbf{F}_t^{\intercal}\mathbf{R}_t^{-1}\mathbf{F}_t)(\mathbf{P}_{t-1}\mathbf{F}_t^{\intercal}\mathbf{R}_t^{-1}\mathbf{F}_t + \mathbf{I})^{-1}}_{\mathbf{I}} \hat{\mathbf{x}}_{t-1} \\
		& = {\color{Mahogany}  \mathbf{B}_t \mathbf{u}_t + \mathbf{F}_t \hat{\mathbf{x}}_{t-1} }
	\end{aligned}
\end{equation}

Next, we find the inverse covariance matrix by taking the second derivative.
\begin{equation}
	\begin{aligned}
		\frac{\partial^2 \mathbf{L}_t(\mathbf{x}_t)}{\partial \mathbf{x}^2_t} & = {\color{Mahogany}  (\mathbf{R}_t + \mathbf{F}_t\mathbf{P}_{t-1}\mathbf{F}_t^{\intercal})^{-1} }
	\end{aligned}
\end{equation}

Finally, we can summarize $\overline{\text{bel}}(\mathbf{x}_{t})$ as follows:
\begin{equation} 
	\begin{aligned} 
		\underbrace{\overline{\text{bel}}(\mathbf{x}_{t})}_{\sim \mathcal{N}(\hat{\mathbf{x}}_{t|t-1}, \mathbf{P}_{t|t-1})  } & = \int \underbrace{p(\mathbf{x}_{t} \ | \ \mathbf{x}_{t-1}, \mathbf{u}_{t})}_{\sim \mathcal{N}(\mathbf{F}_t\mathbf{x}_{t-1}+\mathbf{B}_t \mathbf{u}_t, \mathbf{R}_{t})} \underbrace{\text{bel}(\mathbf{x}_{t-1})}_{\quad \sim \mathcal{N}(\hat{\mathbf{x}}_{t-1}, \mathbf{P}_{t-1})} d \mathbf{x}_{t-1}   \\
		& = \eta \int \exp(-\mathbf{L}_{t}) d \mathbf{x}_{t-1}  \\
		& = \eta \int \exp(-\mathbf{L}_t(\mathbf{x}_{t-1}, \mathbf{x}_t) - \mathbf{L}_t(\mathbf{x}_t)) d \mathbf{x}_{t-1} \\
		& = \eta \exp(-\mathbf{L}_t(\mathbf{x}_t)) \int \exp(-\mathbf{L}_t(\mathbf{x}_{t-1}, \mathbf{x}_t)) d \mathbf{x}_{t-1} \\
		& = \eta' \exp(-\mathbf{L}_t(\mathbf{x}_t)) 
	\end{aligned} 
\end{equation}
\begin{equation} 
	\boxed{ \begin{aligned} 
			& \text{mean } : \hat{\mathbf{x}}_{t|t-1} =  {\color{Mahogany}  \mathbf{F}_t\hat{\mathbf{x}}_{t-1|t-1}+\mathbf{B}_t \mathbf{u}_t }\\
			& \text{covariance }: \mathbf{P}_{t|t-1} = {\color{Mahogany}  \mathbf{F}_t\mathbf{P}_{t-1}\mathbf{F}_t^{\intercal} + \mathbf{R}_t }
	\end{aligned} }
\end{equation}

These equations are used to calculate the mean and covariance in the prediction step.

\subsection{Derivation of KF update step (ver. 1)}
Next, let's derive the equation for ${\text{bel}}(\mathbf{x}_{t})$. The mean and variance we aim to compute for ${\text{bel}}(\mathbf{x}_{t})$ are as follows:
\begin{equation}  
	\begin{aligned} 
		& \underbrace{{\text{bel}}(\mathbf{x}_{t})}_{\sim \mathcal{N}(\hat{\mathbf{x}}_{t|t}, \mathbf{P}_{t|t})  }  = \eta \underbrace{p(\mathbf{z}_{t} \ | \ \mathbf{x}_{t})}_{\sim \mathcal{N}(\mathbf{H}_t\mathbf{x}_{t}, \mathbf{Q}_{t})} \underbrace{\overline{\text{bel}(\mathbf{x}}_{t})}_{\quad \sim \mathcal{N}(\hat{\mathbf{x}}_{t|t-1}, \mathbf{P}_{t|t-1})}   
	\end{aligned} 
\end{equation}

Expanding the above equation in Gaussian form, we get:
\begin{equation} 
	\begin{aligned} 
		& {\text{bel}}(\mathbf{x}_{t}) = \eta \exp \Big(-\frac{1}{2}(\mathbf{z}_t - \mathbf{H}_t\mathbf{x}_{t})^{\intercal} \mathbf{Q}_t^{-1}(\mathbf{z}_t - \mathbf{H}_t\mathbf{x}_{t}) \Big) \cdot \exp \Big( -\frac{1}{2}(\mathbf{x}_{t} - \hat{\mathbf{x}}_{t|t-1})^{\intercal} {\mathbf{P}}_{t|t-1}^{-1}(\mathbf{x}_{t} - \hat{\mathbf{x}}_{t|t-1}) \Big)
	\end{aligned} 
\end{equation}

This equation can be simplified as follows:
\begin{equation} 
	\begin{aligned} 
		& {\text{bel}}(\mathbf{x}_{t}) = \eta \exp(-{\color{Mahogany} \mathbf{J}_{t} })
	\end{aligned} 
\end{equation}
\begin{equation}
	\boxed{	\begin{aligned}
			\text{where, } {\color{Mahogany} \mathbf{J}_t } = \frac{1}{2}(\mathbf{z}_t - \mathbf{H}_t\mathbf{x}_{t})^{\intercal} \mathbf{Q}_t^{-1}(\mathbf{z}_t - \mathbf{H}_t\mathbf{x}_{t}) + \frac{1}{2}(\mathbf{x}_{t} - \hat{\mathbf{x}}_{t|t-1})^{\intercal} {\mathbf{P}}_{t|t-1}^{-1}(\mathbf{x}_{t} - \hat{\mathbf{x}}_{t|t-1})
	\end{aligned} }
\end{equation}

Since ${\text{bel}}(\mathbf{x}_{t})$ follows a Gaussian distribution, we can find the mean and covariance by taking derivatives of $\mathbf{J}_t$.

\begin{equation}
	\begin{aligned}
		\frac{\partial \mathbf{J}_t}{\partial \mathbf{x}_t}
		= -\mathbf{H}_t^{\intercal}\mathbf{Q}_t^{-1}(\mathbf{z}_t - \mathbf{H}_t\mathbf{x}_{t}) + \mathbf{P}_{t|t-1}^{-1}(\mathbf{x}_{t} - \hat{\mathbf{x}}_{t|t-1})
	\end{aligned}
\end{equation}
\begin{equation}
	\begin{aligned}
		\frac{\partial^2 \mathbf{J}_t}{\partial \mathbf{x}^2_t}
		= \mathbf{H}_t^{\intercal}\mathbf{Q}_t^{-1}\mathbf{H}_t + \mathbf{P}_{t|t-1}^{-1} := \mathbf{P}_{t|t}^{-1} \quad \cdots {\color{Mahogany} \text{covariance}^{-1} }
	\end{aligned}
\end{equation}

Setting the first derivative $\frac{\partial \mathbf{J}_t}{\partial \mathbf{x}_{t}}$ to 0, we obtain the following formula:
\begin{equation}
	\begin{aligned}
		\mathbf{H}_t^{\intercal}\mathbf{Q}_t^{-1}(\mathbf{z}_t - \mathbf{H}_t {\color{Red} \mathbf{x}_{t} }) = \mathbf{P}_{t|t-1}^{-1}({\color{Red}\mathbf{x}_{t} } - \hat{\mathbf{x}}_{t|t-1})
	\end{aligned}
\end{equation}

Solving for ${\color{Red} \mathbf{x}_{t} }$, we get:
\begin{equation} \label{eq:derivkf5}
	\begin{aligned}
		& \Leftrightarrow \mathbf{H}_t^{\intercal}\mathbf{Q}_t^{-1}(\mathbf{z}_t - \mathbf{H}_t {\color{Red} \mathbf{x}_{t} } + \mathbf{H}_t\hat{\mathbf{x}}_{t|t-1} - \mathbf{H}_t \hat{\mathbf{x}}_{t|t-1}) = \mathbf{P}_{t|t-1}^{-1}({\color{Red}\mathbf{x}_{t} } - \hat{\mathbf{x}}_{t|t-1}) \\
		& \Leftrightarrow \mathbf{H}_t^{\intercal}\mathbf{Q}_t^{-1}(\mathbf{z}_t -   \mathbf{H}_t \hat{\mathbf{x}}_{t|t-1}) - \mathbf{H}_t^{\intercal}\mathbf{Q}_t^{-1}\mathbf{H}_t({\color{Red}\mathbf{x}_t } - \hat{\mathbf{x}}_{t|t-1}) = \mathbf{P}_{t|t-1}^{-1}({\color{Red}\mathbf{x}_{t} } - \hat{\mathbf{x}}_{t|t-1}) \\ 
		& \Leftrightarrow \mathbf{H}_t^{\intercal}\mathbf{Q}_t^{-1}(\mathbf{z}_t -   \mathbf{H}_t \hat{\mathbf{x}}_{t|t-1})  = (\mathbf{H}_t^{\intercal}\mathbf{Q}_t^{-1}\mathbf{H}_t + \mathbf{P}_{t|t-1}^{-1})({\color{Red}\mathbf{x}_{t} } - \hat{\mathbf{x}}_{t|t-1}) \\
		& \Leftrightarrow \mathbf{H}_t^{\intercal}\mathbf{Q}_t^{-1}(\mathbf{z}_t -   \mathbf{H}_t \hat{\mathbf{x}}_{t|t-1})  = \mathbf{P}_{t|t}^{-1}({\color{Red}\mathbf{x}_{t} } - \hat{\mathbf{x}}_{t|t-1}) \\
		& \Leftrightarrow \underline{ \mathbf{P}_{t|t}\mathbf{H}_t^{\intercal}\mathbf{Q}_t^{-1} } (\mathbf{z}_t -   \mathbf{H}_t \hat{\mathbf{x}}_{t|t-1})  = {\color{Red}\mathbf{x}_{t} } - \hat{\mathbf{x}}_{t|t-1} \\
		& \Leftrightarrow \underline{ \mathbf{K}_t } (\mathbf{z}_t -   \mathbf{H}_t \hat{\mathbf{x}}_{t|t-1})  = {\color{Red}\mathbf{x}_{t} } - \hat{\mathbf{x}}_{t|t-1} \\
		& \therefore {\color{Red}\mathbf{x}_{t} } =  \hat{\mathbf{x}}_{t|t-1} + \mathbf{K}_t(\mathbf{z}_t -   \mathbf{H}_t \hat{\mathbf{x}}_{t|t-1})  \quad \cdots {\color{Mahogany} \text{mean} }
	\end{aligned}
\end{equation}

Thus, the mean and variance of $\text{bel}(\mathbf{x}_{t})$ can be obtained as follows:
\begin{equation}
	\boxed{ \begin{aligned}
			& \text{kalman gain }: \mathbf{K}_t = \mathbf{P}_{t|t} \mathbf{H}_t^{\intercal}\mathbf{Q}_t^{-1} \\
			& \text{mean }: \hat{\mathbf{x}}_{t|t} = \hat{\mathbf{x}}_{t|t-1} + \mathbf{K}_t(\mathbf{z}_t -   \mathbf{H}_t \hat{\mathbf{x}}_{t|t-1}) \\
			& \text{covariance }: \mathbf{P}_{t|t} = (\mathbf{H}_t^{\intercal}\mathbf{Q}_t^{-1}\mathbf{H}_t + \mathbf{P}_{t|t-1}^{-1})^{-1}
	\end{aligned} }
\end{equation}

However, calculating the inverse matrix for the variance $\mathbf{P}_{t|t}$ of $\text{bel}(\mathbf{x}_{t})$ can be time-consuming. We can use the matrix inversion lemma to transform this equation.
\begin{equation} \label{eq:derivkf6}
	\begin{aligned} 
		\mathbf{P}_{t|t} & = (\mathbf{H}_t^{\intercal}\mathbf{Q}_t^{-1}\mathbf{H}_t + \mathbf{P}_{t|t-1}^{-1})^{-1} \\
		& = (\mathbf{P}_{t|t-1}^{-1} + \mathbf{H}_t^{\intercal}\mathbf{Q}_t^{-1}\mathbf{H}_t)^{-1} \\
		& = \mathbf{P}_{t|t-1} - \underline{\underline{ \mathbf{P}_{t|t-1}\mathbf{H}_t^{\intercal}(\mathbf{Q}_t + \mathbf{H}_t\mathbf{P}_{t|t-1}\mathbf{H}_t^{\intercal})^{-1} }} \mathbf{H}_t \mathbf{P}_{t|t-1} \\ 
		& = \mathbf{P}_{t|t-1} - \underline{\underline{ \mathbf{K}_t }} \mathbf{H}_t \mathbf{P}_{t|t-1} \\
		& = (\mathbf{I} - \mathbf{K}_t\mathbf{H}_t) \mathbf{P}_{t|t-1}
	\end{aligned} 
\end{equation}

There is a transformation between the Kalman gain $\mathbf{K}_t$ in (\ref{eq:derivkf5}) and (\ref{eq:derivkf6}) as follows:
\begin{equation} 
	\begin{aligned} 
		\mathbf{K}_t & = \underline{ \mathbf{P}_{t|t} \mathbf{H}_t^{\intercal}\mathbf{Q}_t^{-1} } \\
		& = \mathbf{P}_{t|t} \mathbf{H}_t^{\intercal}\mathbf{Q}_t^{-1} \underbrace{(\mathbf{H}_t\mathbf{P}_{t|t-1}\mathbf{H}_t^{\intercal} + \mathbf{Q}_t)(\mathbf{H}_t\mathbf{P}_{t|t-1}\mathbf{H}_t^{\intercal} + \mathbf{Q}_t)^{-1}}_{\mathbf{I}} \\
		& = \mathbf{P}_{t|t}(\mathbf{H}_t^{\intercal}\mathbf{Q}_t^{-1}\mathbf{H}_t \mathbf{P}_{t|t-1}\mathbf{H}_t^{\intercal} + \mathbf{H}_t^{\intercal} \underbrace{\mathbf{Q}_t^{-1}\mathbf{Q}_t}_{\mathbf{I}})(\mathbf{H}_t\mathbf{P}_{t|t-1}\mathbf{H}_t^{\intercal} + \mathbf{Q}_t)^{-1} \\
		& = \mathbf{P}_{t|t}(\mathbf{H}_t^{\intercal}\mathbf{Q}_t^{-1}\mathbf{H}_t \mathbf{P}_{t|t-1}\mathbf{H}_t^{\intercal} + \mathbf{H}_t^{\intercal}) (\mathbf{H}_t\mathbf{P}_{t|t-1}\mathbf{H}_t^{\intercal} + \mathbf{Q}_t)^{-1} \\
		& = \mathbf{P}_{t|t}(\mathbf{H}_t^{\intercal}\mathbf{Q}_t^{-1}\mathbf{H}_t \mathbf{P}_{t|t-1}\mathbf{H}_t^{\intercal} + \underbrace{\mathbf{P}_{t|t-1}^{-1}\mathbf{P}_{t|t-1}}_{\mathbf{I}} \mathbf{H}_t^{\intercal} )(\mathbf{H}_t\mathbf{P}_{t|t-1}\mathbf{H}_t^{\intercal} + \mathbf{Q}_t)^{-1} \\
		& = \mathbf{P}_{t|t} \underbrace{ (\mathbf{H}_t^{\intercal}\mathbf{Q}_t^{-1}\mathbf{H}_t + \mathbf{P}_{t|t-1}^{-1})}_{\mathbf{P}_{t|t}^{-1}}\mathbf{P}_{t|t-1}\mathbf{H}_t^{\intercal} (\mathbf{H}_t\mathbf{P}_{t|t-1}\mathbf{H}_t^{\intercal} + \mathbf{Q}_t)^{-1} \\ 
		& =  \underline{\underline{ \mathbf{P}_{t|t-1}\mathbf{H}_t^{\intercal} (\mathbf{H}_t\mathbf{P}_{t|t-1}\mathbf{H}_t^{\intercal} + \mathbf{Q}_t)^{-1}  }}
	\end{aligned} 
\end{equation}

Finally, ${\text{bel}}(\mathbf{x}_{t})$ can be summarized as follows:
\begin{equation}
	\begin{aligned} 
		\underbrace{{\text{bel}}(\mathbf{x}_{t})}_{\sim \mathcal{N}(\hat{\mathbf{x}}_{t|t}, \mathbf{P}_{t|t})  }  & = \eta \underbrace{p(\mathbf{z}_{t} \ | \ \mathbf{x}_{t})}_{\sim \mathcal{N}(\mathbf{H}_t\mathbf{x}_{t}, \mathbf{Q}_{t})}  \underbrace{\overline{\text{bel}(\mathbf{x}}_{t})}_{\quad \sim \mathcal{N}(\hat{\mathbf{x}}_{t|t-1}, \mathbf{P}_{t|t-1})}    \\
		& = \eta \exp(-\mathbf{J}_{t})
	\end{aligned} 
\end{equation}
\begin{equation}
	\boxed{ \begin{aligned}
			& \text{kalman gain }: \mathbf{K}_t = \hat{\mathbf{x}}_{t|t-1} + \mathbf{K}_{t}( \mathbf{z}_{t} - \mathbf{H}_{t}\hat{\mathbf{x}}_{t|t-1}) \\
			& \text{mean }: \hat{\mathbf{x}}_{t|t} = \hat{\mathbf{x}}_{t|t-1} + \mathbf{K}_t(\mathbf{z}_t -   \mathbf{H}_t \hat{\mathbf{x}}_{t|t-1}) \\
			& \text{covariance }: \mathbf{P}_{t|t} = (\mathbf{I} - \mathbf{K}_{t}\mathbf{H}_{t})\mathbf{P}_{t|t-1}
	\end{aligned} }
\end{equation}

\subsection{Derivation of KF update step (ver. 2)}
Since ${\text{bel}}(\mathbf{x}_{t})$ consists of the product of likelihood and prior, we can find the mean and variance of the posterior more simply by calculating the conditional probability (conditional pdf).
\begin{equation}  
	\begin{aligned} 
		\underbrace{{\text{bel}}(\mathbf{x}_{t})}_{\sim \mathcal{N}(\hat{\mathbf{x}}_{t|t}, \mathbf{P}_{t|t})  }  & = \eta \underbrace{p(\mathbf{z}_{t} \ | \ \mathbf{x}_{t})}_{\sim \mathcal{N}(\mathbf{H}_t\mathbf{x}_{t}, \mathbf{Q}_{t})}  \underbrace{\overline{\text{bel}(\mathbf{x}}_{t})}_{\quad \sim \mathcal{N}(\hat{\mathbf{x}}_{t|t-1}, \mathbf{P}_{t|t-1})}   \\
		\text{posterior} & = \text{ likelihood } \times \text{ prior }
	\end{aligned} 
\end{equation}

\mybox{Tip}{gray!40}{gray!10}{
	\textbfazure{Conditional gaussian distribution}
	~\\
	Given two vector random variables $\mathbf{x}$ and $\mathbf{z}$, if the conditional probability distribution $p(\mathbf{x}|\mathbf{z})$ follows a Gaussian distribution, then
	\begin{equation}
		\begin{aligned}
			p(\mathbf{x}| \mathbf{z}) & = \frac{p(\mathbf{x},\mathbf{z})}{p(\mathbf{z})} = \frac{p(\mathbf{z} | \mathbf{x})p(\mathbf{x})}{p(\mathbf{z})} = \eta \cdot p(\mathbf{z}|\mathbf{x})p(\mathbf{x}) \\
			& \sim \mathcal{N}(\mathbb{E}(\mathbf{x}|\mathbf{z}), \mathbf{C}_{x|z})
		\end{aligned}
	\end{equation}
	
	The mean $\mathbb{E}(\mathbf{x}|\mathbf{z})$ and variance $\mathbf{C}_{x|z}$ are as follows:
	\begin{equation}
		\boxed{ \begin{aligned}
				& \mathbb{E}(\mathbf{x}|\mathbf{z})= \mathbb{E}(\mathbf{x}) + \mathbf{C}_{xz}\mathbf{C}_{zz}^{inverted}(\mathbf{z}  - \mathbb{E}(\mathbf{z})) \\
				& \mathbf{C}_{x|z} = \mathbf{C}_{xx} - \mathbf{C}_{xz}\mathbf{C}_{zz}^{inverted}\mathbf{C}_{xz}^{\intercal}
		\end{aligned} }
	\end{equation}
}

This can be explained graphically as follows:
\begin{figure}[h!]
	\centering
	\includegraphics[width=13cm]{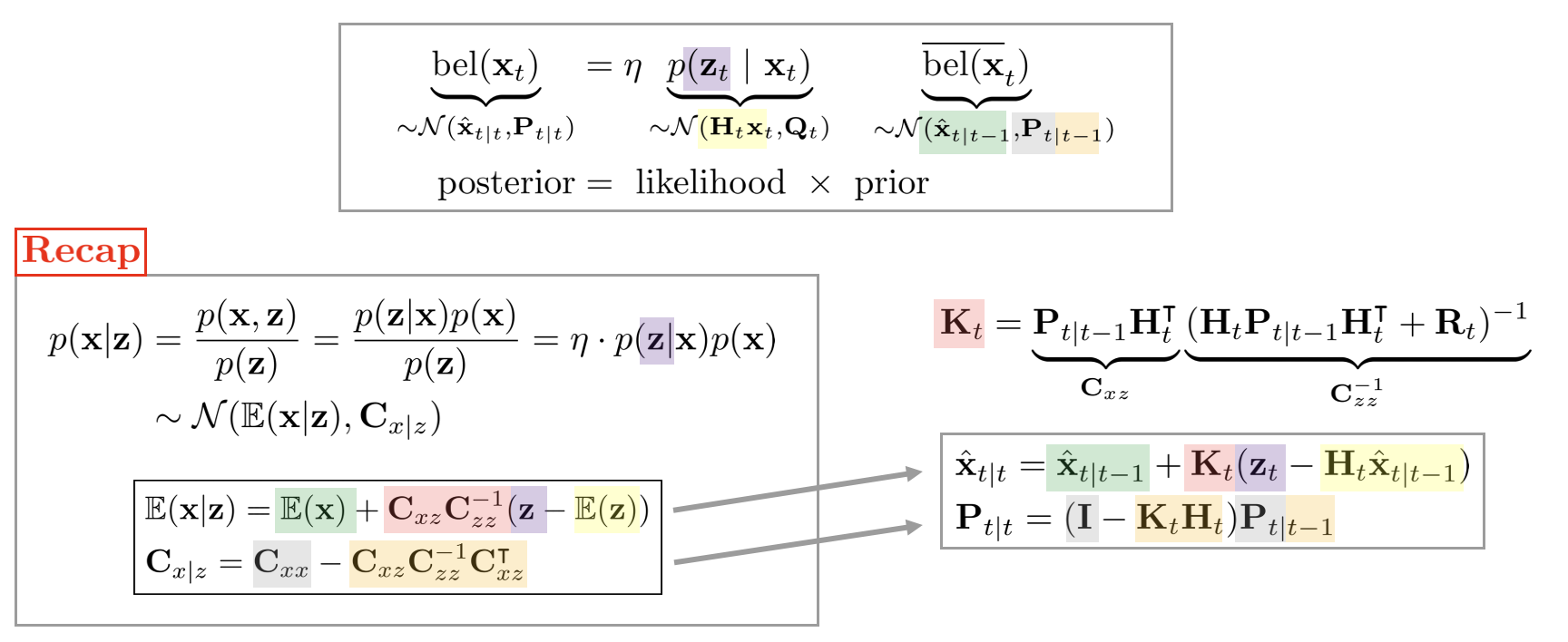}
\end{figure}

\newpage

\section{MAP, GN, and EKF relationship} \label{sec:map}
\subsection{Traditional EKF derivation}

Let's consider the observation model function for the EKF given as follows. For the convenience of development, the observation noise $\mathbf{v}_{t}$ is positioned outside.

\begin{equation}
	\begin{aligned}
		& \text{Observation Model: } &  \mathbf{z}_{t} = \mathbf{h}(\mathbf{x}_{t}) + \mathbf{v}_{t}
	\end{aligned}
\end{equation}

The correction step of the EKF unfolds the following formula to derive the mean $\hat{\mathbf{x}}_{t|t}$ and the covariance $\mathbf{P}_{t|t}$.

\begin{equation}
	\begin{aligned} 
		\text{bel}( \mathbf{x}_{t}) = \eta \cdot p(\mathbf{z}_{t} \ | \ \mathbf{x}_{t}) \overline{\text{bel}}( \mathbf{x}_{t}) \sim \mathcal{N}(\hat{\mathbf{x}}_{t|t}, \mathbf{P}_{t|t})
	\end{aligned} 
\end{equation}

- $p(\mathbf{z}_{t} \ | \ \mathbf{x}_{t}) \sim \mathcal{N}( \mathbf{h}_{t}(\mathbf{x}_{t}) , \mathbf{R}_{t})$ \\
- $\overline{\text{bel}}( \mathbf{x}_{t}) \sim \mathcal{N}( \hat{\mathbf{x}}_{t|t-1}, \mathbf{P}_{t|t-1})$\\

\begin{equation} 
	\boxed{  \begin{aligned} 
			& \hat{\mathbf{x}}_{t|t} = \mathbf{K}_{t} ( \mathbf{z}_{t} - \mathbf{h}(\hat{\mathbf{x}}_{t|t-1})) \\
			& \mathbf{P}_{t|t} = (\mathbf{I} - \mathbf{K}_{t}\mathbf{H}_{t})\mathbf{P}_{t|t-1}
	\end{aligned} }
\end{equation}

\subsection{MAP-based EKF derivation}

\subsubsection{Start from MAP estimator}

As a method for deriving the correction step, one can use the maximum a posteriori (MAP) estimation that maximizes the probability of the posterior. The details were written with reference to \ref{ref:12}.

\begin{equation} \label{eq:map1}
	\begin{aligned} 
		\hat{\mathbf{x}}_{t|t} & = \arg\max_{ \mathbf{x}_{t}} \text{bel}(\mathbf{x}_{t}) \\
		& = \arg\max_{ \mathbf{x}_{t}} p(\mathbf{x}_{t} | \mathbf{z}_{1:t}, \mathbf{u}_{1:t})  \quad \cdots \text{posterior} \\
		& \propto \arg\max_{ \mathbf{x}_{t}} p(\mathbf{z}_{t} | \mathbf{x}_{t}) \overline{\text{bel}}(\mathbf{x}_{t}) \quad \cdots \text{likelihood} \cdot \text{prior} \\
		& \propto \arg\max_{\mathbf{x}_{t}}   \exp  \bigg( -\frac{1}{2}  \bigg[ (\mathbf{z}_{t} - \mathbf{h}(\mathbf{x}_{t}))^{\intercal} \mathbf{R}_{t}^{-1} (\mathbf{z}_{t} - \mathbf{h}(\mathbf{x}_{t})) \\
		& \quad\quad\quad \quad \quad \quad + ( \mathbf{x}_{t} - \hat{\mathbf{x}}_{t|t-1})^{\intercal} \mathbf{P}_{t|t-1}^{-1} ( \mathbf{x}_{t}  - \hat{\mathbf{x}}_{t|t-1}) \bigg] \bigg)
	\end{aligned} 
\end{equation}
- $p(\mathbf{z}_{t} \ | \ \mathbf{x}_{t}) \sim \mathcal{N}( \mathbf{h}_{t}(\mathbf{x}_{t}) , \mathbf{R}_{t})$ \\
- $\overline{\text{bel}}( \mathbf{x}_{t}) \sim \mathcal{N}( \hat{\mathbf{x}}_{t|t-1}, \mathbf{P}_{t|t-1})$\\

Removing the negative sign turns the maximization problem into a minimization problem, and it can be organized into the following optimization formula.

\begin{equation} \label{eq:map2}
	\begin{aligned} 
		\hat{\mathbf{x}}_{t|t} & \propto \arg\min_{\mathbf{x}_{t}}   \exp  \bigg( (\mathbf{z}_{t} - \mathbf{h}(\mathbf{x}_{t}))^{\intercal} \mathbf{R}_{t}^{-1} (\mathbf{z}_{t} - \mathbf{h}(\mathbf{x}_{t})) \\
		& \quad\quad\quad\quad\quad\quad + ( \mathbf{x}_{t}  - \hat{\mathbf{x}}_{t|t-1} )^{\intercal} \mathbf{P}^{-1_{t|t-1}} ( \mathbf{x}_{t}  - \hat{\mathbf{x}}_{t|t-1} ) \bigg) \end{aligned} 
\end{equation}

\begin{equation} \label{eq:map3}
	\boxed{ \begin{aligned} 
			\hat{\mathbf{x}}_{t|t} & = \arg\min_{\mathbf{x}_{t}} \left\| \mathbf{z}_{t} - \mathbf{h}(\mathbf{x}_{t}) \right\|_{\mathbf{R}_{t}^{-1}} + \left\| \mathbf{x}_{t}   - \hat{\mathbf{x}}_{t|t-1} \right\|_{\mathbf{P}_{t|t-1}^{-1}} 
	\end{aligned} }
\end{equation}
- $\left\| \mathbf{a} \right\|_{\mathbf{B}} = \mathbf{a}^{\intercal}\mathbf{B}\mathbf{a}$ : Mahalanobis norm\\

Expanding the formula within (\ref{eq:map3}) and defining it as a cost function $\mathbf{C}_{t}$ results in the following. Note that switching the order of $\mathbf{x}_{t}  - \hat{\mathbf{x}}_{t|t-1}$ does not affect the overall value.

\begin{equation} 
	\begin{aligned} 
		\mathbf{C}_{t} = (\mathbf{z}_{t} - \mathbf{h}(\mathbf{x}_{t}))^{\intercal} \mathbf{R}_{t}^{-1} (\mathbf{z}_{t} - \mathbf{h}(\mathbf{x}_{t})) + ( \hat{\mathbf{x}}_{t|t-1}  - \mathbf{x}_{t} )^{\intercal} \mathbf{P}^{-1}_{t|t-1} ( \hat{\mathbf{x}}_{t|t-1}  - \mathbf{x}_{t} )
	\end{aligned} 
\end{equation}

This can be expressed in matrix form as follows.

\begin{equation} \label{eq:map4}
	\begin{aligned} 
		\mathbf{C}_{t} = \begin{bmatrix}   \hat{\mathbf{x}}_{t|t-1} - \mathbf{x}_{t}  \\ \mathbf{z}_{t} - \mathbf{h}(\mathbf{x}_{t})\end{bmatrix}^{\intercal} \begin{bmatrix} \mathbf{P}^{-1}_{t|t-1} & \mathbf{0} \\ \mathbf{0} & \mathbf{R}^{-1}_{t} \end{bmatrix} \begin{bmatrix}   \hat{\mathbf{x}}_{t|t-1}  - \mathbf{x}_{t}  \\ \mathbf{z}_{t} - \mathbf{h}(\mathbf{x}_{t})\end{bmatrix}
	\end{aligned} 
\end{equation}

\subsubsection{MLE of new observation function}
A new observation function satisfying the above equation can be defined as follows.

\begin{equation} 
	\boxed{ \begin{aligned} 
			\mathbf{y}_{t} & = \mathbf{g}(\mathbf{x}_{t}) + \mathbf{e}_{t} \\ 
			& \sim \mathcal{N}(\mathbf{g}(\mathbf{x}_{t}), \mathbf{P}_{\mathbf{e}})
	\end{aligned} }
\end{equation}
- $\mathbf{y}_{t} = \begin{bmatrix} \hat{\mathbf{x}}_{t|t-1} \\ \mathbf{z}_{t} \end{bmatrix}$\\
- $\mathbf{g}(\mathbf{x}_{t}) = \begin{bmatrix} \mathbf{x}_{t} \\ \mathbf{h}(\mathbf{x}_{t}) \end{bmatrix}$\\
- $\mathbf{e}_{t} \sim \mathcal{N}(0, \mathbf{P}_{\mathbf{e}})$\\
- $\mathbf{P}_{\mathbf{e}} = \begin{bmatrix} \mathbf{P}_{t|t-1} & \mathbf{0} \\ \mathbf{0} & \mathbf{R}_{t} \end{bmatrix}$\\

The nonlinear function $\mathbf{g}(\mathbf{x}_{t})$ can be linearized as follows.

\begin{equation} 
	\begin{aligned} 
		\mathbf{g}(\mathbf{x}_{t}) & \approx \mathbf{g}(\hat{\mathbf{x}}_{t|t-1}) + \mathbf{J}_{t}(\mathbf{x}_{t} - \hat{\mathbf{x}}_{t|t-1})  \\
		& = \mathbf{g}(\hat{\mathbf{x}}_{t|t-1}) + \mathbf{J}_{t}\delta \hat{\mathbf{x}}_{t|t-1} 
	\end{aligned} 
\end{equation}

The Jacobian $\mathbf{J}_{t}$ is as follows.

\begin{equation} 
	\begin{aligned} 
		\mathbf{J}_{t} & = \frac{\partial \mathbf{g}}{\partial \mathbf{x}_{t}}\bigg|_{\mathbf{x}_{t} = \hat{\mathbf{x}}_{t|t-1}} \\ 
		& \frac{\partial \begin{bmatrix} \mathbf{x}_{t} \\ \mathbf{h}(\mathbf{x}_{t}) \end{bmatrix} }{\partial \mathbf{x}_{t}}\bigg|_{\mathbf{x}_{t} = \hat{\mathbf{x}}_{t|t-1}} \\
		& \frac{\partial \begin{bmatrix} \mathbf{x}_{t} \\ \mathbf{h}(\hat{\mathbf{x}}_{t|t-1}) + \mathbf{H}_{t} (\mathbf{x}_{t} - \hat{\mathbf{x}}_{t|t-1}) \end{bmatrix} }{\partial \mathbf{x}_{t}}\bigg|_{\mathbf{x}_{t} = \hat{\mathbf{x}}_{t|t-1}} \\
		& = \begin{bmatrix} \mathbf{I} \\ \mathbf{H}_{t} \end{bmatrix}
	\end{aligned} 
\end{equation}

Therefore, the likelihood for $\mathbf{y}_{t}$ can be developed as follows.

\begin{equation}  \label{eq:map5}
	\begin{aligned} 
		p(\mathbf{y}_{t} | \mathbf{x}_{t}) & \sim \mathcal{N}(\mathbf{g}(\mathbf{x}_{t}), \mathbf{P}_{\mathbf{e}}) \\
		& = \eta \cdot \exp\bigg( -\frac{1}{2}(\mathbf{y}_{t} - \mathbf{g}(\mathbf{x}_{t}))^{\intercal} \mathbf{P}_{\mathbf{e}}^{-1}(\mathbf{y}_{t} - \mathbf{g}(\mathbf{x}_{t})) \bigg)
	\end{aligned} 
\end{equation}

Thus, the maximum a posteriori (MAP) problem for the existing $\text{bel}(\mathbf{x}_{t})$ resolves into solving the maximum likelihood estimation (MLE) problem for $p(\mathbf{y}_{t} | \mathbf{x}_{t})$. Solving equation (\ref{eq:map5}) with MLE leads to the following.

\begin{equation}  \label{eq:map6}
	\begin{aligned} 
		\hat{\mathbf{x}}_{t|t} & = \arg\max_{\mathbf{x}_{t}} p(\mathbf{y}_{t} | \mathbf{x}_{t}) \\ 
		& \propto \arg\min_{\mathbf{x}_{t}} -\ln p(\mathbf{y}_{t} | \mathbf{x}_{t}) \\ 
		& \propto \arg\min_{\mathbf{x}_{t}} \frac{1}{2} (\mathbf{y}_{t} - \mathbf{g}(\mathbf{x}_{t}))^{\intercal} \mathbf{P}_{\mathbf{e}}^{-1}(\mathbf{y}_{t} - \mathbf{g}(\mathbf{x}_{t})) \\
		& \propto \arg\min_{\mathbf{x}_{t}} \left\| \mathbf{y}_{t} - \mathbf{g}(\mathbf{x}_{t}) \right\|_{\mathbf{P}_{\mathbf{e}}^{-1}}
	\end{aligned} 
\end{equation}

\subsubsection{Gauss-Newton Optimization}
Equation (\ref{eq:map6}) has the form of a least squares method. Specifically, it is also called weighted least squares (WLS) since the weight $\mathbf{P}_{\mathbf{e}}^{-1}$ is multiplied in between. After linearizing and rearranging the formula, it becomes as follows.

\begin{equation}  
	\begin{aligned} 
		\hat{\mathbf{x}}_{t|t} & = \arg\min_{\mathbf{x}_{t}} \left\| \mathbf{y}_{t} - \mathbf{g}(\mathbf{x}_{t}) \right\|_{\mathbf{P}_{\mathbf{e}}^{-1}} \\ 
		&  = \arg\min_{\mathbf{x}_{t}} \left\| \mathbf{y}_{t} - \mathbf{g}(\hat{\mathbf{x}}_{t|t-1})  - \mathbf{J}_{t} \delta \hat{\mathbf{x}}_{t|t-1} \right\|_{\mathbf{P}_{\mathbf{e}}^{-1}} \\ 
		&  = \arg\min_{\mathbf{x}_{t}} \left\|  \mathbf{J}_{t} \delta \hat{\mathbf{x}}_{t|t-1} - (\mathbf{y}_{t} - \mathbf{g}(\hat{\mathbf{x}}_{t|t-1}) ) \right\|_{\mathbf{P}_{\mathbf{e}}^{-1}} \\ 
		&  = \arg\min_{\mathbf{x}_{t}} \left\|  \mathbf{J}_{t} \delta \hat{\mathbf{x}}_{t|t-1} - \mathbf{r}_{t} \right\|_{\mathbf{P}_{\mathbf{e}}^{-1}} \\ 
	\end{aligned} 
\end{equation}
- $\delta \hat{\mathbf{x}}_{t|t-1} = \mathbf{x}_{t} - \hat{\mathbf{x}}_{t|t-1}$ : For ease of expression, $\mathbf{x}_{t}$ is represented as the true state \\

The linearized residual term $\mathbf{r}_{t}$ is as follows.

\begin{equation}
	\begin{aligned}
		\mathbf{r}_{t} & = \mathbf{y}_{t} -  \mathbf{g}(\hat{\mathbf{x}}_{t|t-1}) \\
		& = \mathbf{J}_{t} \delta \hat{\mathbf{x}}_{t|t-1}  + \mathbf{e} \\ & \sim \mathcal{N}(0, \mathbf{P}_{\mathbf{e}}) 
	\end{aligned}
\end{equation}

By solving the normal equations of Gauss-Newton, the solution is as follows.

\begin{equation}  
	\begin{aligned} 
		& (\mathbf{J}_{t}^{\intercal} \mathbf{P}_{\mathbf{e}}^{-1} \mathbf{J}_{t}) \delta \hat{\mathbf{x}}_{t|t-1} = \mathbf{J}_{t}^{\intercal} \mathbf{P}_{\mathbf{e}}^{-1} \mathbf{r}_{t} \\
	\end{aligned}  
\end{equation}

\begin{equation}  
	\boxed{ \begin{aligned} & \therefore \delta \hat{\mathbf{x}}_{t|t-1} = (\mathbf{J}_{t}^{\intercal} \mathbf{P}_{\mathbf{e}}^{-1}\mathbf{J}_{t})^{-1} \mathbf{J}_{t}^{\intercal} \mathbf{P}_{\mathbf{e}}^{-1} \mathbf{r}_{t} 
	\end{aligned}  }
\end{equation}

In the above equation, the part $(\mathbf{J}_{t}^{\intercal} \mathbf{P}_{\mathbf{e}}^{-1} \mathbf{J}_{t})$ is generally called the approximate Hessian matrix $\tilde{\mathbf{H}}$.

~\\
\textbf{Posterior covariance matrix $\mathbf{P}_{t|t}$:}
~\\ \\
$\mathbf{P}_{t|t}$ can be obtained as follows.

	\begin{equation}  
		\boxed{	\begin{aligned}
				\mathbf{P}_{t|t}
				=& \mathbb{E}[(\mathbf{x}_{t} - \hat{\mathbf{x}}_{t|t-1})(\mathbf{x}_{t} - \hat{\mathbf{x}}_{t|t-1})^{\intercal}] \\
				=& \mathbb{E}(\delta \hat{\mathbf{x}}_{t|t-1}\delta \hat{\mathbf{x}}_{t|t-1}^{\intercal}) \\
				=& \mathbb{E}\bigg[
				(\mathbf{J}_{t}^{\intercal} \mathbf{P}_{\mathbf{e}}^{-1} \mathbf{J}_{t})^{-1} \mathbf{J}_{t}^{\intercal} \mathbf{P}_{\mathbf{e}}^{-1} \mathbf{r}_{t}
				\mathbf{r}_{t}^{\intercal} \mathbf{P}_{\mathbf{e}}^{-\intercal} \mathbf{J}_{t} (\mathbf{J}_{t}^{\intercal} \mathbf{P}_{\mathbf{e}}^{-1} \mathbf{J}_{t})^{-\intercal} \bigg] \\
				=&
				(\mathbf{J}_{t}^{\intercal} \mathbf{P}_{\mathbf{e}}^{-1} \mathbf{J}_{t})^{-1} \mathbf{J}_{t}^{\intercal} \mathbf{P}_{\mathbf{e}}^{-1} \mathbb{E}(\mathbf{r}_{t}
				\mathbf{r}_{t}^{\intercal}) \mathbf{P}_{\mathbf{e}}^{-\intercal} \mathbf{J}_{t} (\mathbf{J}_{t}^{\intercal} \mathbf{P}_{\mathbf{e}}^{-1} \mathbf{J}_{t})^{-\intercal} \quad {\color{Mahogany}\leftarrow \mathbb{E}(\mathbf{r}_{t}\mathbf{r}_{t}^{\intercal}) = \mathbf{P}_{\mathbf{e}} } \\  
				=& \left(
				{\color{Mahogany}{\mathbf{J}_{t}^{\intercal} \mathbf{P}_{\mathbf{e}}^{-1} \mathbf{J}_{t}}}
				\right)^{-1} \\
				=& \left(
				\begin{bmatrix} \mathbf{I} & \mathbf{H}_{t}^{\intercal} \end{bmatrix}
				\begin{bmatrix} \mathbf{P}_{t|t-1} & \mathbf{0} \\ \mathbf{0} & \mathbf{R}_{t} \end{bmatrix}^{-1}
				\begin{bmatrix} \mathbf{I} \\ \mathbf{H}_{t} \end{bmatrix}
				\right)^{-1} \\
				=& \left( \mathbf{P}_{t|t-1}^{-1} + \mathbf{H}_{t}^{\intercal}\mathbf{R}_{t}^{-1}\mathbf{H}_{t} \right)^{-1} \\
				=& \mathbf{P}_{t|t-1} - \mathbf{P}_{t|t-1}\mathbf{H}_{t}^{\intercal} (\mathbf{H}_{t}\mathbf{P}_{t|t-1}\mathbf{H}_{t}^{\intercal} + \mathbf{R}_{t})^{-1} \mathbf{H}_{t} \mathbf{P}_{t|t-1} \quad {\color{Mahogany} \leftarrow \text{matrix inversion lemmas} } \\
				=& {\color{Mahogany}{(\mathbf{I} - \mathbf{K}_{t}\mathbf{H}_{t}) \mathbf{P}_{t|t-1}}}
		\end{aligned} }
	\end{equation}

As can be seen from the fifth line of the above equation, the inverse of the approximate Hessian matrix $\tilde{\mathbf{H}}^{-1}$ obtained through GN and the EKF's posterior covariance $\mathbf{P}_{t|t}$ are the same.

\begin{equation}  
	\begin{aligned}
		\tilde{\mathbf{H}}^{-1} = (\mathbf{J}_{t}^{\intercal} \mathbf{P}_{\mathbf{e}}^{-1} \mathbf{J}_{t})^{-1} = \mathbf{P}_{t|t}
	\end{aligned} 
\end{equation}

~\\
\textbf{Posterior mean $\mathbf{x}_{t|t}$:}
~\\ \\
By iteratively performing GN, the $j$th $\mathbf{x}_{t|t, j}$ can be determined as follows.

	\begin{equation}  
		\boxed{ \begin{aligned}
				\hat{\mathbf{x}}_{t|t,j+1}
				=& \hat{\mathbf{x}}_{t|t,j} + \delta \hat{\mathbf{x}}_{t|t,j} \\
				=& \hat{\mathbf{x}}_{t|t,j} + (\mathbf{J}_{t}^{\intercal} \mathbf{P}_{t|t-1}^{-1} \mathbf{J}_{t})^{-1} (\mathbf{J}_{t}^{\intercal} \mathbf{P}_{t|t-1}^{-1} \mathbf{r}_{t}) \\
				=& (\mathbf{J}_{t}^{\intercal} \mathbf{P}_{t|t-1}^{-1} \mathbf{J}_{t})^{-1} \mathbf{J}_{t}^{\intercal} \mathbf{P}_{t|t-1}^{-1} (\mathbf{y}_{t} - \mathbf{g}(\hat{\mathbf{x}}_{t|t,j}) + \mathbf{J}_{t} \hat{\mathbf{x}}_{t|t,j}) \quad {\color{Mahogany} \leftarrow \mathbf{r}_{t} = \mathbf{y}_{t} - \mathbf{g}(\hat{\mathbf{x}}_{t|t,j}) }\\
				=&
				\left( \mathbf{P}_{t|t-1}^{-1} + \mathbf{H}_{t}^{\intercal}\mathbf{R}_{t}^{-1} \mathbf{H}_{t} \right)^{-1}
				\begin{bmatrix} \mathbf{P}_{t|t-1}^{-1} & \mathbf{H}_{t}^{\intercal}\mathbf{R}_{t}^{-1} \end{bmatrix}
				\begin{bmatrix}
					\hat{\mathbf{x}}_{t|t-1} \\
					\mathbf{z}_{t} - \mathbf{h}(\hat{\mathbf{x}}_{t|t,j}) + \mathbf{H}_{t} \hat{\mathbf{x}}_{t|t,j}
				\end{bmatrix} \quad {\color{Mahogany} \leftarrow \text{expand } \mathbf{J}_{t} } \\
				=&
				\left( \mathbf{P}_{t|t-1}^{-1} + \mathbf{H}_{t}^{\intercal}\mathbf{R}_{t}^{-1} \mathbf{H}_{t} \right)^{-1}
				\left(
				\mathbf{H}_{t}^{\intercal}\mathbf{R}_{t}^{-1}(\mathbf{z}_{t} - \mathbf{h}(\hat{\mathbf{x}}_{t|t,j}) + \mathbf{H}_{t} \hat{\mathbf{x}}_{t|t,j}) + \mathbf{P}_{t|t-1}^{-1} \hat{\mathbf{x}}_{t|t-1}
				\right) \\
				=&
				\left( \mathbf{P}_{t|t-1}^{-1} + \mathbf{H}_{t}^{\intercal}\mathbf{R}_{t}^{-1} \mathbf{H}_{t} \right)^{-1}
				\left(
				\mathbf{H}^{\intercal}\mathbf{R}_{t}^{-1}(\mathbf{z}_{t} - \mathbf{h}(\hat{\mathbf{x}}_{t|t,j}) - \mathbf{H}_{t}(\hat{\mathbf{x}}_{t|t-1} - \hat{\mathbf{x}}_{t|t,j}) + \mathbf{H}_{t} \hat{\mathbf{x}}_{t|t-1}) + \mathbf{P}_{t|t-1}^{-1} \hat{\mathbf{x}}_{t|t-1}
				\right) \\
				=&
				\left( \mathbf{P}_{t|t-1}^{-1} + \mathbf{H}_{t}^{\intercal}\mathbf{R}_{t}^{-1} \mathbf{H}_{t} \right)^{-1}
				\left(
				\mathbf{H}^{\intercal}\mathbf{R}_{t}^{-1}(\mathbf{z}_{t} - \mathbf{h}(\hat{\mathbf{x}}_{t|t,j}) - \mathbf{H}_{t}(\hat{\mathbf{x}}_{t|t-1} - \hat{\mathbf{x}}_{t|t,j})) + \underbrace{(\mathbf{P}_{t|t-1}^{-1} +  \mathbf{H}_{t}^{\intercal}\mathbf{R}_{t}^{-1} \mathbf{H}_{t} )\hat{\mathbf{x}}_{t|t-1}}_{\hat{\mathbf{x}}_{t|t-1}}
				\right) \\
				=&
				\hat{\mathbf{x}}_{t|t-1} +
				{\color{Mahogany}{
						\left( \mathbf{P}_{t|t-1}^{-1} + \mathbf{H}_{t}^{\intercal}\mathbf{R}_{t}^{-1} \mathbf{H}_{t} \right)^{-1} \mathbf{H}_{t}^{\intercal}\mathbf{R}_{t}^{-1}
				}}
				(\mathbf{z}_{t} - \mathbf{h}(\hat{\mathbf{x}}_{t|t,j}) - \mathbf{H}_{t}(\hat{\mathbf{x}}_{t|t-1} - \hat{\mathbf{x}}_{t|t,j})) \\
				=&
				\hat{\mathbf{x}}_{t|t-1} + {\color{Mahogany}{ \mathbf{K}_{t} }} (\mathbf{z}_{t} - \mathbf{h}(\hat{\mathbf{x}}_{t|t,j}) - \mathbf{H}_{t}(\hat{\mathbf{x}}_{t|t-1} - \hat{\mathbf{x}}_{t|t,j}))
		\end{aligned} } 
	\end{equation}

The above equation is identical to the IEKF (\ref{eq:iekf-cor}) equation. As seen in the last equation, estimating the solution of the EKF through Gauss-Newton and estimating the solution through IEKF have the same meaning. If it is the first iteration $j=0$, $\hat{\mathbf{x}}_{t|t,0} = \hat{\mathbf{x}}_{t|t-1}$, then the equation is summarized as follows.

\begin{equation}  
	\boxed{ \begin{aligned}
			\hat{\mathbf{x}}_{t|t} = \hat{\mathbf{x}}_{t|t-1} + \mathbf{K}_{t} (\mathbf{z}_{t} - \mathbf{h}(\hat{\mathbf{x}}_{t|t-1}))
	\end{aligned} } 
\end{equation}

This is identical to the solution of the EKF. That is, the EKF implies that GN iteration=1, and IEKF performs the same operations as GN.

\section{Derivation of IESKF Update Step} \label{sec:deriv}
In this section, the intermediate derivation process of obtaining $\delta \hat{\mathbf{x}}$ from the update step in the IESKF process is explained. The derivation primarily references the contents of a link. First, let us revisit the update formula (\ref{eq:ieskf6})

\begin{equation}  
	\boxed{ \begin{aligned}  
			\arg\min_{\delta \hat{\mathbf{x}}_{t|t,j}} \quad & \left\| \mathbf{z}_{t} - \mathbf{h}(\hat{\mathbf{x}}_{t|t-1}, 0, 0) - \mathbf{H}_{t} \delta\hat{\mathbf{x}}_{t|t,j} \right\|_{\mathbf{H}_{\mathbf{v}}\mathbf{R}^{-1}\mathbf{H}_{\mathbf{v}}^{\intercal}}  + \left\| \hat{\mathbf{x}}_{t|t,j} - \hat{\mathbf{x}}_{t|t-1} + \mathbf{J}_{t,j}\delta\hat{\mathbf{x}}_{t|t,j} \right\|_{\mathbf{P}_{t|t-1}^{-1}}  
	\end{aligned} } 
\end{equation}

The symbols in the above equation are simplified as follows:
\begin{equation*}  \begin{aligned}  
		\mathbf{x}
		= \arg\min_{\mathbf{x}} \quad \left\| \mathbf{z} - \mathbf{H}\mathbf{x} \right\|^{2}_{\mathbf{R}} + \left\| \mathbf{c} + \mathbf{J}\mathbf{x} \right\|^{2}_{\mathbf{P}}  
	\end{aligned}  
\end{equation*}
- $\mathbf{z} = \mathbf{z}_{t} - \mathbf{h}(\hat{\mathbf{x}}_{t|t-1}, 0, 0)$\\
- $\mathbf{c} = \hat{\mathbf{x}}_{t|t,j} - \hat{\mathbf{x}}_{t|t-1}$\\
- $\mathbf{R} = \mathbf{H}_{\mathbf{v}}\mathbf{R}^{-1}\mathbf{H}_{\mathbf{v}}^{\intercal}$\\
- $\mathbf{x} = \delta \hat{\mathbf{x}}$\\

Expanding the norm in the above equation gives:
\begin{equation*}  \begin{aligned}  
		\mathbf{x} = \arg\min_{\mathbf{x}} \ \bigg[ \underbrace{ (\mathbf{z} - \mathbf{H}\mathbf{x})^{\intercal}\mathbf{R}^{-1}(\mathbf{z} - \mathbf{H}\mathbf{x}) + (\mathbf{c} + \mathbf{J}\mathbf{x})^{\intercal}\mathbf{P}^{-1}(\mathbf{c} + \mathbf{J}\mathbf{x}) }_{\mathbf{r}} \bigg]
	\end{aligned}  
\end{equation*}

If we expand only the $\mathbf{r}$ part in the above equation, it becomes:
\begin{equation} 
	\begin{aligned} 
		\mathbf{r} & = \mathbf{z}^{\intercal}\mathbf{R}^{-1}\mathbf{z} {\color{Mahogany} - \mathbf{x}^{\intercal}\mathbf{H}^{\intercal}\mathbf{R}^{-1}\mathbf{z}- \mathbf{z}^{\intercal}\mathbf{R}^{-1}\mathbf{H}\mathbf{x} } + \mathbf{x}^{\intercal}\mathbf{H}^{\intercal}\mathbf{R}^{-1}\mathbf{H}\mathbf{x} + \mathbf{c}^{\intercal}\mathbf{P}^{-1}\mathbf{c} + {\color{Mahogany} \mathbf{x}^{\intercal}\mathbf{J}^{\intercal}\mathbf{P}^{-1}\mathbf{c} + \mathbf{c}^{\intercal}\mathbf{P}^{-1}\mathbf{J}\mathbf{x} } + \mathbf{x}^{\intercal}\mathbf{J}^{\intercal}\mathbf{P}^{-1}\mathbf{J}\mathbf{x} \\
		& = \mathbf{z}^{\intercal}\mathbf{R}^{-1}\mathbf{z} {\color{Mahogany} - 2\mathbf{x}^{\intercal}\mathbf{H}^{\intercal}\mathbf{R}^{-1}\mathbf{z} } + \mathbf{x}^{\intercal}\mathbf{H}^{\intercal}\mathbf{R}^{-1}\mathbf{H}\mathbf{x} + \mathbf{c}^{\intercal}\mathbf{P}^{-1}\mathbf{c} {\color{Mahogany} + 2\mathbf{x}^{\intercal}\mathbf{J}^{\intercal}\mathbf{P}^{-1}\mathbf{c} } + \mathbf{x}^{\intercal}\mathbf{J}^{\intercal}\mathbf{P}^{-1}\mathbf{J}\mathbf{x}
	\end{aligned} 
\end{equation}

Differentiating $\mathbf{r}$ with respect to $\mathbf{x}$ yields:
\begin{equation} \label{eq:deriveieskf1}
	\begin{aligned}  
		\frac{\partial \mathbf{r}}{\partial {\mathbf{x}}} = - \mathbf{H}^{\intercal}\mathbf{R}^{-1}\mathbf{z} + \mathbf{H}^{\intercal}\mathbf{R}^{-1}\mathbf{H}\mathbf{x} +\mathbf{J}^{\intercal}\mathbf{P}^{-1}\mathbf{c} + \mathbf{J}^{\intercal}\mathbf{P}^{-1}\mathbf{J}\mathbf{x} & = 0 \\ 
		{\color{Mahogany} (\mathbf{H}^{\intercal}\mathbf{R}^{-1}\mathbf{H} + \mathbf{J}^{\intercal}\mathbf{P}^{-1}\mathbf{J}) } \mathbf{x} & = 
		\mathbf{H}^{\intercal}\mathbf{R}^{-1}\mathbf{z} - \mathbf{J}^{\intercal}\mathbf{P}^{-1}\mathbf{c} \\ 
		\mathbf{x} & = {\color{Mahogany}(\mathbf{H}^{\intercal}\mathbf{R}^{-1}\mathbf{H} + \mathbf{J}^{\intercal}\mathbf{P}^{-1}\mathbf{J})^{-1} } ( 
		\mathbf{H}^{\intercal}\mathbf{R}^{-1}\mathbf{z} - \mathbf{J}^{\intercal}\mathbf{P}^{-1}\mathbf{c}) 
	\end{aligned}  
\end{equation}

The $(\mathbf{H}^{\intercal}\mathbf{R}^{-1}\mathbf{H} + \mathbf{J}^{\intercal}\mathbf{P}^{-1}\mathbf{J})^{-1}$ part can be expanded using the matrix inversion lemma as follows:
\begin{equation}  
	\begin{aligned}  
		{\color{Mahogany}(\underbrace{\mathbf{J}^{\intercal}\mathbf{P}^{-1}\mathbf{J}}_{\bar{\mathbf{P}}^{-1}} +\mathbf{H}^{\intercal}\mathbf{R}^{-1}\mathbf{H})^{-1} = (\bar{\mathbf{P}} -\bar{\mathbf{P}}\mathbf{H}^{\intercal}(\mathbf{R} + \mathbf{H}\bar{\mathbf{P}}\mathbf{H}^{\intercal})^{-1}\mathbf{H}\bar{\mathbf{P}} }
	\end{aligned}  
\end{equation}
- $\bar{\mathbf{P}} = \mathbf{J}^{-1}\mathbf{P}\mathbf{J}^{-\intercal}$ \\

\mybox{Tip}{gray!40}{gray!10}{
	\textbf{Matrix Inversion Lemma} \\
	$(\mathbf{A}+\mathbf{UCV})^{-1}=\mathbf{A}^{-1}-\mathbf{A}^{-1}\mathbf{U}(\mathbf{C}^{-1}+\mathbf{VA}^{-1}\mathbf{U})^{-1}\mathbf{VA}^{-1}$
}

Substituting this equation into (\ref{eq:deriveieskf1}) yields:
\begin{equation} 
	\begin{aligned} 
		{\mathbf{x}} = {\color{Mahogany}(\bar{\mathbf{P}} -\bar{\mathbf{P}}\mathbf{H}^{\intercal}(\mathbf{R} + \mathbf{H}\bar{\mathbf{P}}\mathbf{H}^{\intercal})^{-1}\mathbf{H}\bar{\mathbf{P}})} (
		\mathbf{H}^{\intercal}\mathbf{R}^{-1}\mathbf{z} - \mathbf{J}^{\intercal}\mathbf{P}^{-1}\mathbf{c})
	\end{aligned} 
\end{equation}

Expanding the above equation results in:
\begin{equation}  
	\begin{aligned}   
		{\mathbf{x}} & = (\bar{\mathbf{P}} -\bar{\mathbf{P}}\mathbf{H}^{\intercal}(\mathbf{R} + \mathbf{H}\bar{\mathbf{P}}\mathbf{H}^{\intercal})^{-1}\mathbf{H}\bar{\mathbf{P}})( 
		\mathbf{H}^{\intercal}\mathbf{R}^{-1}\mathbf{z} - \mathbf{J}^{\intercal}\mathbf{P}^{-1}\mathbf{c}) \\  
		& = \bar{\mathbf{P}}\mathbf{H}^{\intercal}\mathbf{R}^{-1}\mathbf{z} - \bar{\mathbf{P}}\mathbf{J}^{\intercal}\mathbf{P}^{-1}\mathbf{c} 
		-\bar{\mathbf{P}}\mathbf{H}^{\intercal}(\mathbf{R} + {\color{Mahogany} \mathbf{H}\bar{\mathbf{P}}\mathbf{H}^{\intercal})^{-1} } \mathbf{H}\bar{\mathbf{P}}\mathbf{H}^{\intercal}\mathbf{R}^{-1}  \mathbf{z} +\bar{\mathbf{P}}\mathbf{H}^{\intercal}(\mathbf{R} + \mathbf{H}\bar{\mathbf{P}}\mathbf{H}^{\intercal})^{-1}\mathbf{H}\bar{\mathbf{P}}\mathbf{J}^{\intercal}\mathbf{P}^{-1}\mathbf{c} \\ 
		& = \bar{\mathbf{P}}\mathbf{H}^{\intercal}\mathbf{R}^{-1}\mathbf{z} - \bar{\mathbf{P}}\mathbf{J}^{\intercal}\mathbf{P}^{-1}\mathbf{c} 
		-\bar{\mathbf{P}}\mathbf{H}^{\intercal} \cancel{ (\mathbf{R} + \mathbf{H}\bar{\mathbf{P}}\mathbf{H}^{\intercal})^{-1} }  ({\color{Mahogany} \cancel{ (\mathbf{H}\bar{\mathbf{P}}\mathbf{H}^{\intercal}+\mathbf{R})} \mathbf{R}^{-1} - \mathbf{I}) }  \mathbf{z}  +\bar{\mathbf{P}}\mathbf{H}^{\intercal}(\mathbf{R} + \mathbf{H}\bar{\mathbf{P}}\mathbf{H}^{\intercal})^{-1}\mathbf{H}\bar{\mathbf{P}}\mathbf{J}^{\intercal}\mathbf{P}^{-1}\mathbf{c} \\ 
		& = \cancel{ \bar{\mathbf{P}}\mathbf{H}^{\intercal}\mathbf{R}^{-1}\mathbf{z} } - \bar{\mathbf{P}}\mathbf{J}^{\intercal}\mathbf{P}^{-1}\mathbf{c} 
		- \cancel{ \bar{\mathbf{P}}\mathbf{H}^{\intercal}\mathbf{R}^{-1}\mathbf{z} }  + \bar{\mathbf{P}}\mathbf{H}^{\intercal}(\mathbf{R} + \mathbf{H}\bar{\mathbf{P}}\mathbf{H}^{\intercal})^{-1}\mathbf{z}  +\bar{\mathbf{P}}\mathbf{H}^{\intercal}(\mathbf{R} + \mathbf{H}\bar{\mathbf{P}}\mathbf{H}^{\intercal})^{-1}\mathbf{H}\bar{\mathbf{P}}\mathbf{J}^{\intercal}\mathbf{P}^{-1}\mathbf{c} \\ 
		& =  - \bar{\mathbf{P}}\mathbf{J}^{\intercal}\mathbf{P}^{-1}\mathbf{c} 
		+ \bar{\mathbf{P}}\mathbf{H}^{\intercal}(\mathbf{R} + \mathbf{H}\bar{\mathbf{P}}\mathbf{H}^{\intercal})^{-1}\mathbf{z}   +\bar{\mathbf{P}}\mathbf{H}^{\intercal}(\mathbf{R} + \mathbf{H}\bar{\mathbf{P}}\mathbf{H}^{\intercal})^{-1}\mathbf{H}\bar{\mathbf{P}}\mathbf{J}^{\intercal}\mathbf{P}^{-1}\mathbf{c} \\ 
		& =  - \mathbf{J}^{-1}\underbrace{\mathbf{P}\mathbf{J}^{-\intercal}\mathbf{J}^{\intercal}\mathbf{P}^{-1}}_{\mathbf{I}} \mathbf{c} 
		+ {\color{Cyan} \mathbf{J}^{-1}\mathbf{P}\mathbf{J}^{-\intercal}\mathbf{H}^{\intercal}( \underbrace{\mathbf{R} + \mathbf{H}\mathbf{J}^{-1}\mathbf{P}\mathbf{J}^{-\intercal}\mathbf{H}^{\intercal}}_{\mathbf{S}})^{-1} } \mathbf{z}  +{\color{Cyan}  \mathbf{J}^{-1}\mathbf{P}\mathbf{J}^{-\intercal}\mathbf{H}^{\intercal}(\underbrace{\mathbf{R} + \mathbf{H}\mathbf{J}^{-1}\mathbf{P}\mathbf{J}^{-\intercal}\mathbf{H}^{\intercal}}_{\mathbf{S}})^{-1} } \mathbf{HJ}^{-1}\underbrace{\mathbf{PJ}^{-\intercal}\mathbf{J}^{\intercal}\mathbf{P}^{-1}}_{\mathbf{I}} \mathbf{c} \\ 
		& = {\color{Cyan} \mathbf{K} } \mathbf{z} + {\color{Cyan} \mathbf{K} } \mathbf{H}\mathbf{J}^{-1}\mathbf{c} - \mathbf{J}^{-1}\mathbf{c} 
	\end{aligned}     
\end{equation}
- $\mathbf{K} = \mathbf{J}^{-1}\mathbf{PJ}^{-\intercal}\mathbf{H}^{\intercal}\mathbf{S}^{-1}$ \\

Restoring the symbols substituted in the last line of the equation yields the final update equation (\ref{eq:ieskf8}) as follows:
\begin{equation} 
	\boxed{ \begin{aligned}  
			\therefore \delta \hat{\mathbf{x}}_{t|t,j} = \mathbf{K}_{t,j} \bigg(  \mathbf{z}_{t} - \mathbf{h}(\hat{\mathbf{x}}_{t|t,j}, 0 ,0) + \mathbf{H}_{t,j}\mathbf{J}^{-1}_{t,j}(\hat{\mathbf{x}}_{t|t,j} - \hat{\mathbf{x}}_{t|t-1}) \bigg) - \mathbf{J}^{-1}_{t,j}(\hat{\mathbf{x}}_{t|t,j} - \hat{\mathbf{x}}_{t|t-1})
	\end{aligned} }
\end{equation}
- $\mathbf{z} = \mathbf{z}_{t} - \mathbf{h}(\hat{\mathbf{x}}_{t|t-1}, 0, 0)$\\
- $\mathbf{c} = \hat{\mathbf{x}}_{t|t,j} - \hat{\mathbf{x}}_{t|t-1}$\\
- $\mathbf{R} = \mathbf{H}_{\mathbf{v}}\mathbf{R}^{-1}\mathbf{H}_{\mathbf{v}}^{\intercal}$\\
- $\mathbf{x} = \delta \hat{\mathbf{x}}$\\

The derivation process up to now can be illustrated as follows.
\begin{figure}[h!]
	\centering
	\includegraphics[width=16cm]{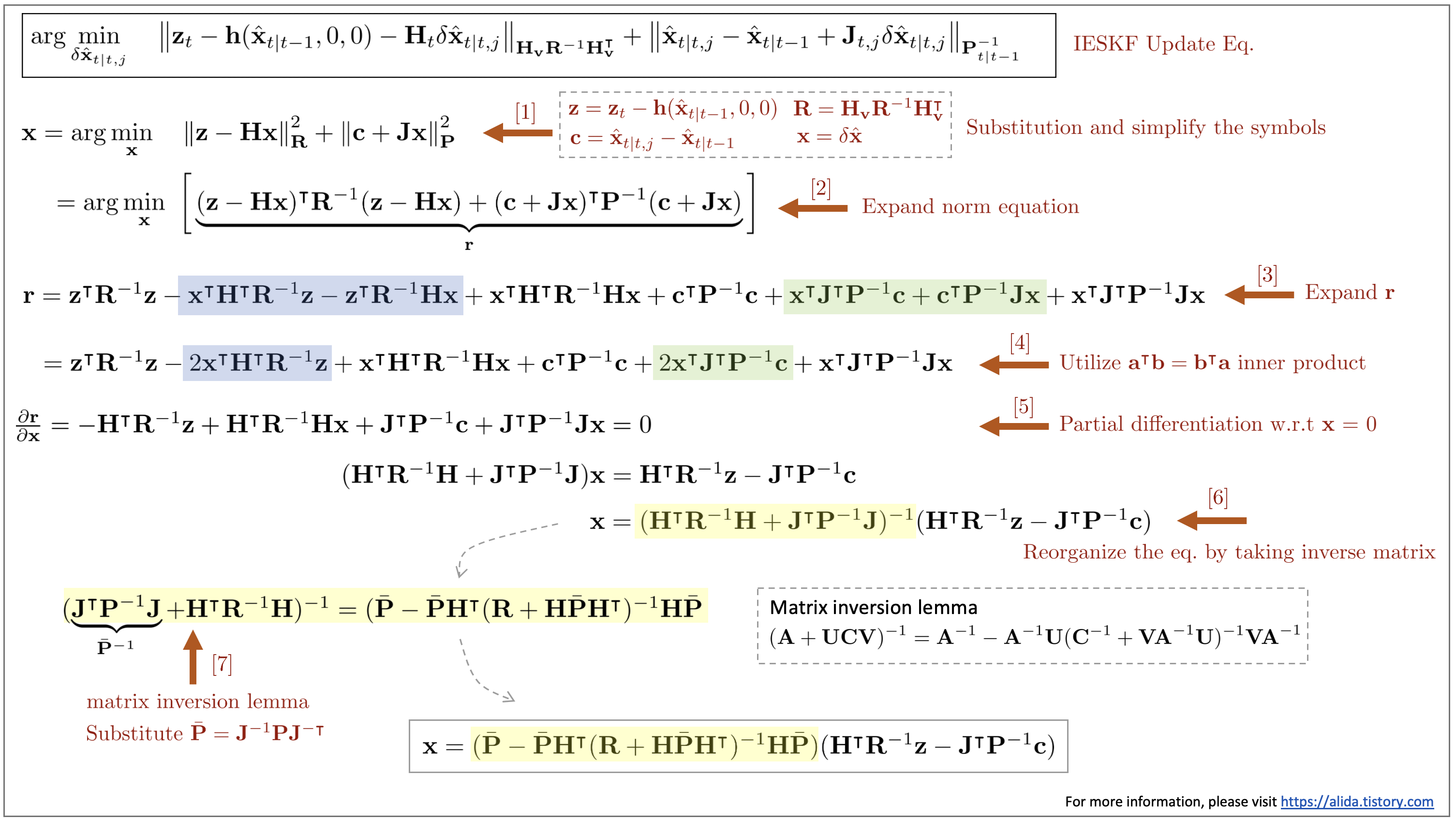}
\end{figure}
\begin{figure}[h!]
	\centering
	\includegraphics[width=16cm]{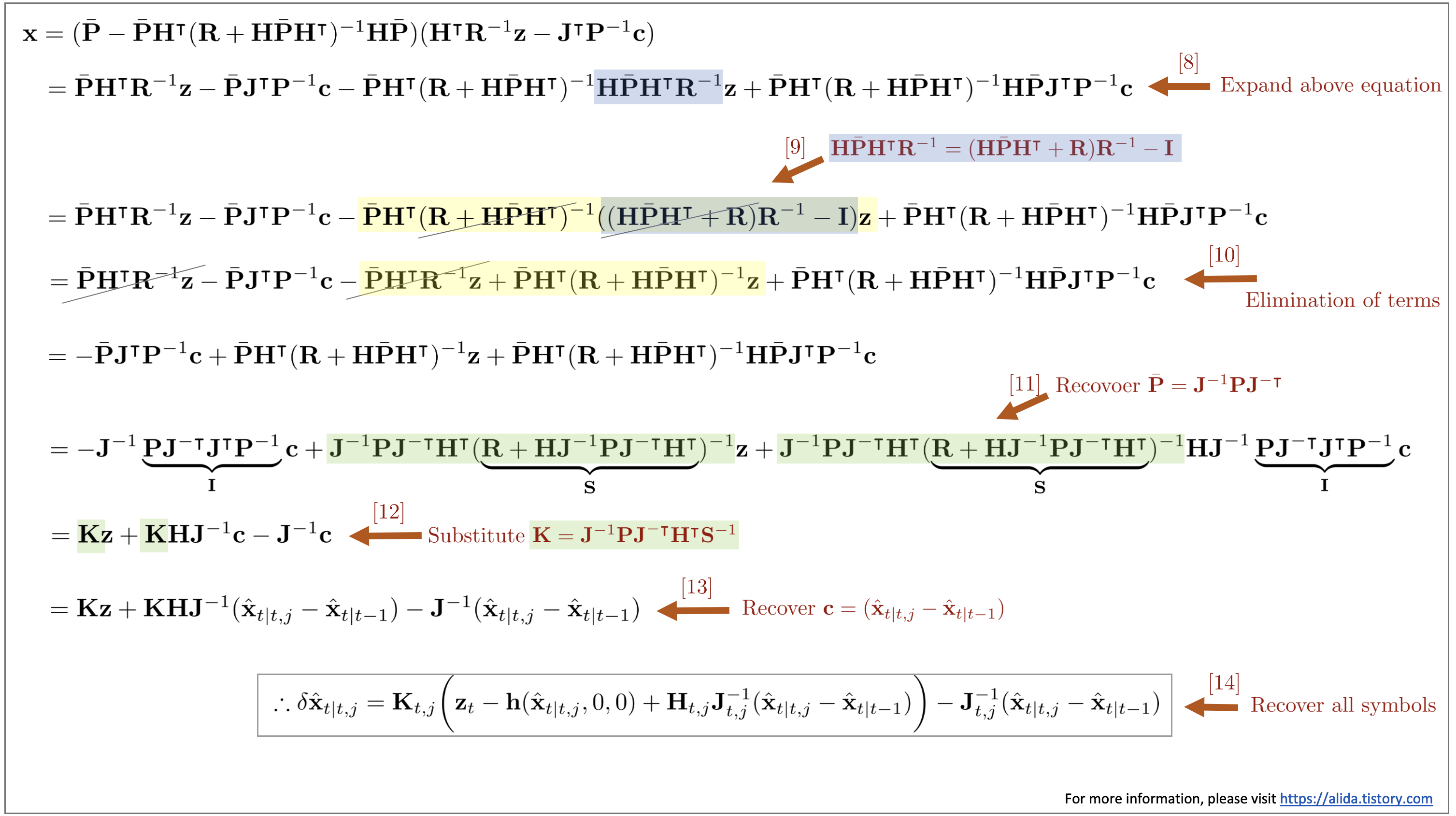}
\end{figure}

\section{Wrap-up}
The explanation of KF, EKF, ESKF, IEKF, IESKF so far can be represented on a single slide as follows. Click to see the larger image.

	\subsection{Kalman Filter (KF)}
	\begin{figure}[h!]
		\centering
		\includegraphics[width=16cm]{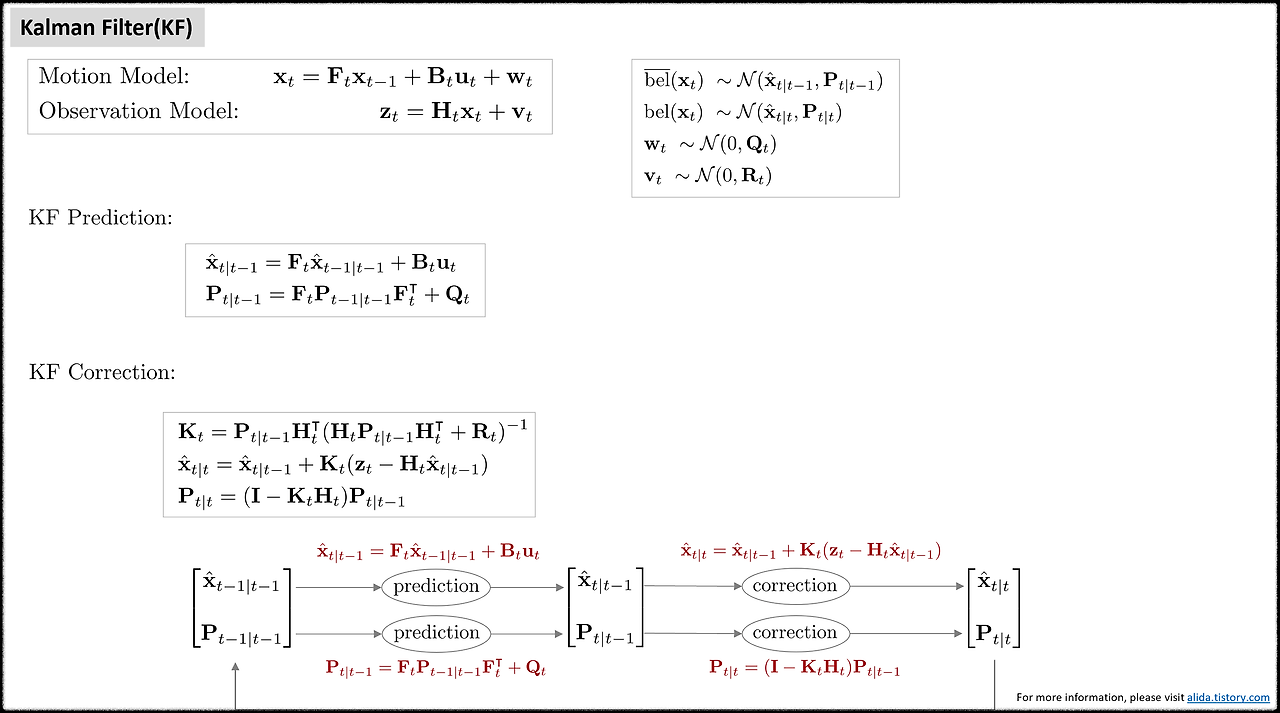}
	\end{figure}
	\subsection{Extended Kalman Filter (EKF)}
	\begin{figure}[h!]
		\centering
		\includegraphics[width=16cm]{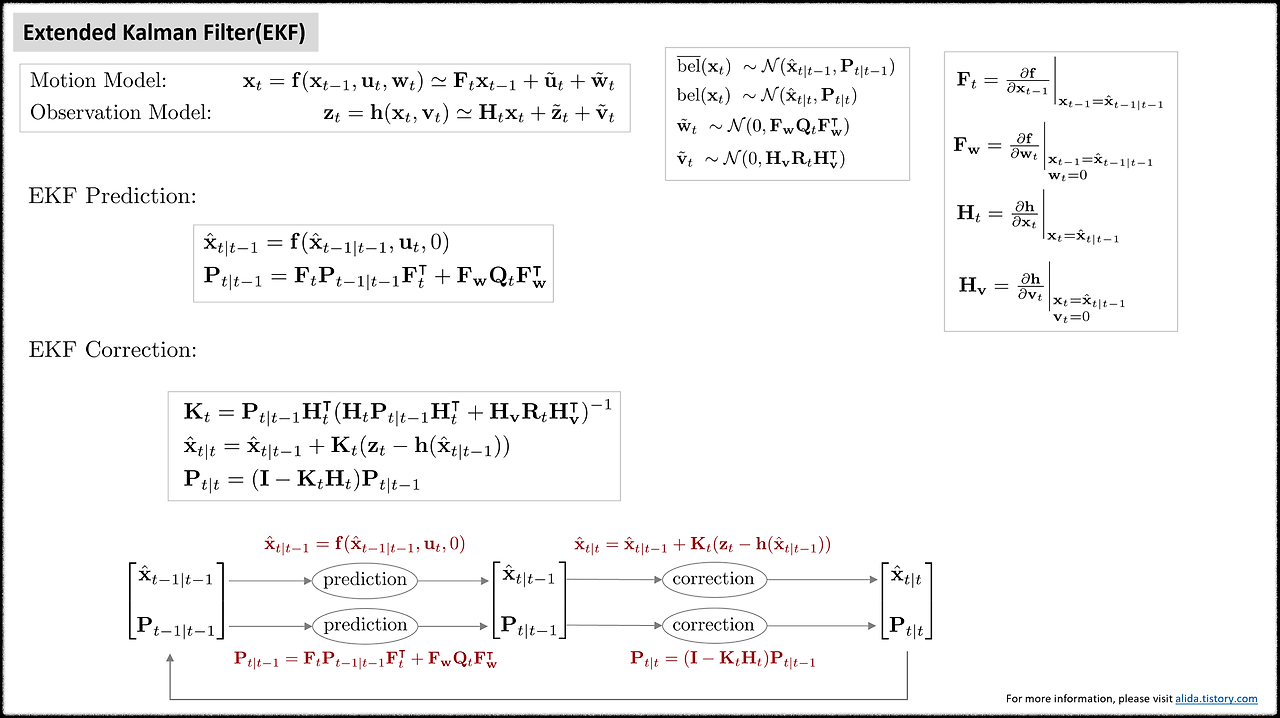}
	\end{figure}
	\subsection{Error-state Kalman Filter (ESKF)}
	\begin{figure}[h!]
		\centering
		\includegraphics[width=16cm]{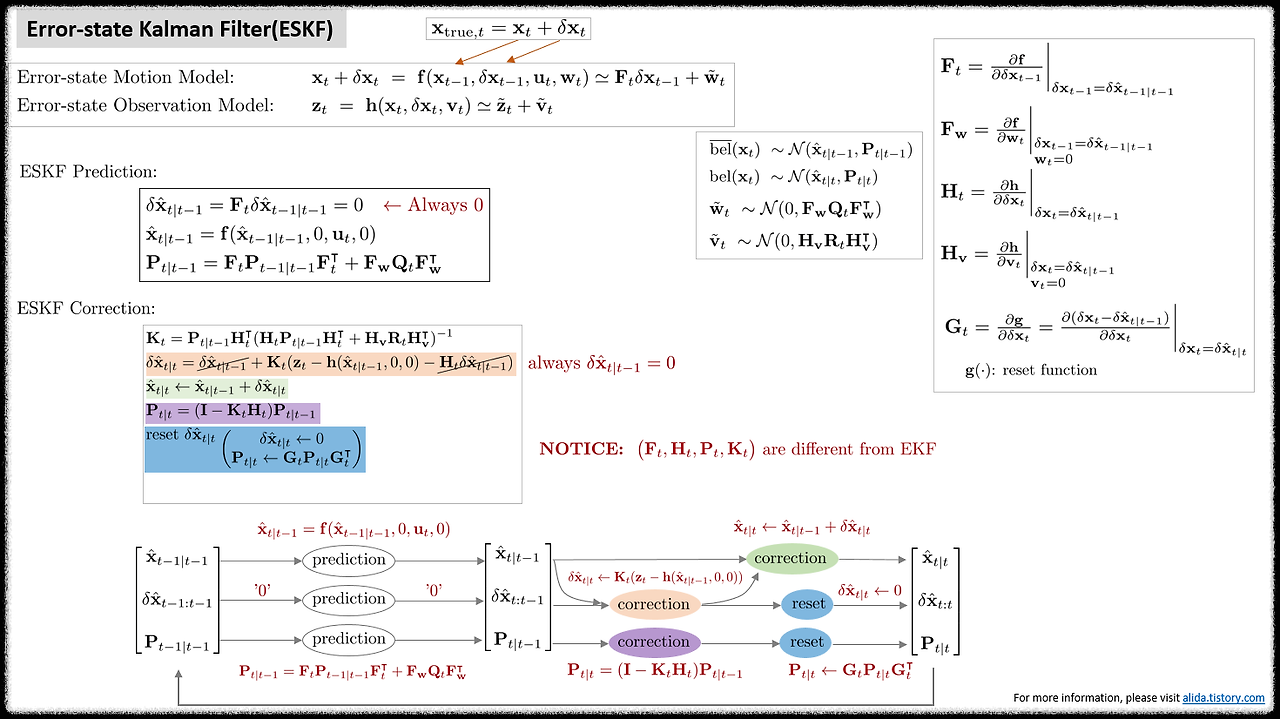}
	\end{figure}
	\subsection{Iterated Extended Kalman Filter (IEKF)}
	\begin{figure}[h!]
		\centering 
		\includegraphics[width=16cm]{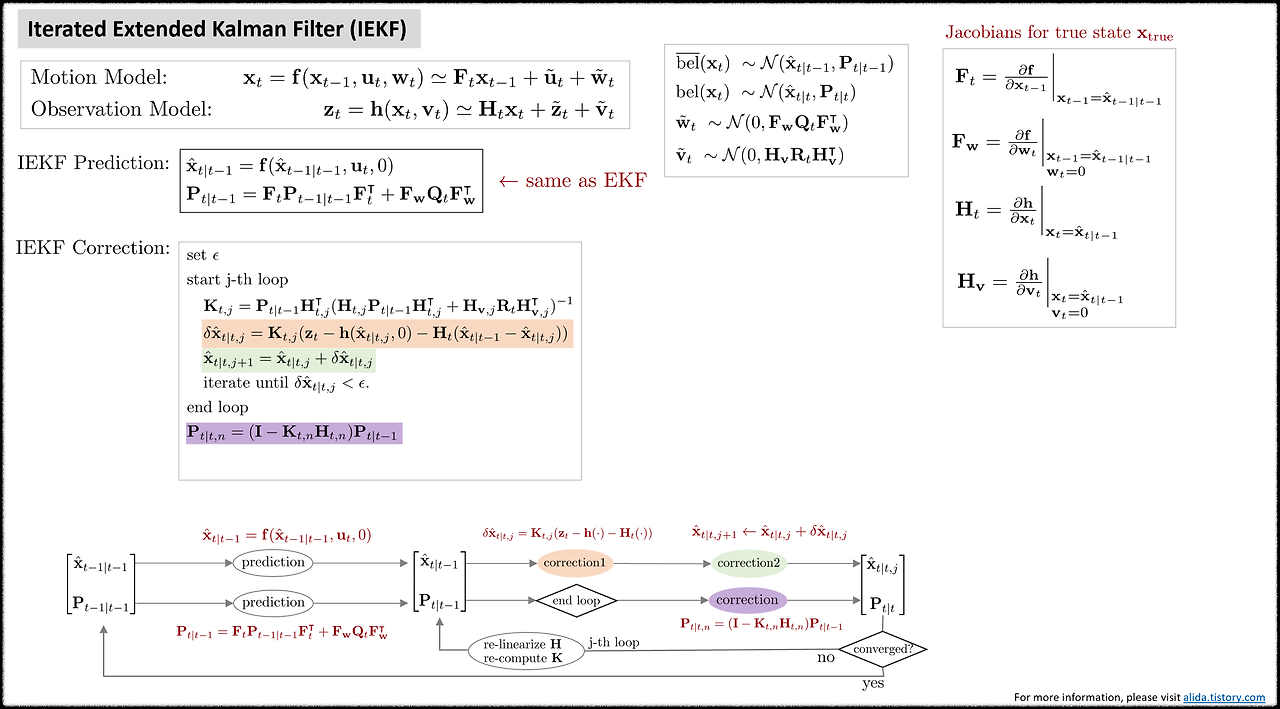}
	\end{figure}
	\subsection{Iterated Error-state Kalman Filter (IESKF)}
	\begin{figure}[h!]
		\centering
		\includegraphics[width=16cm]{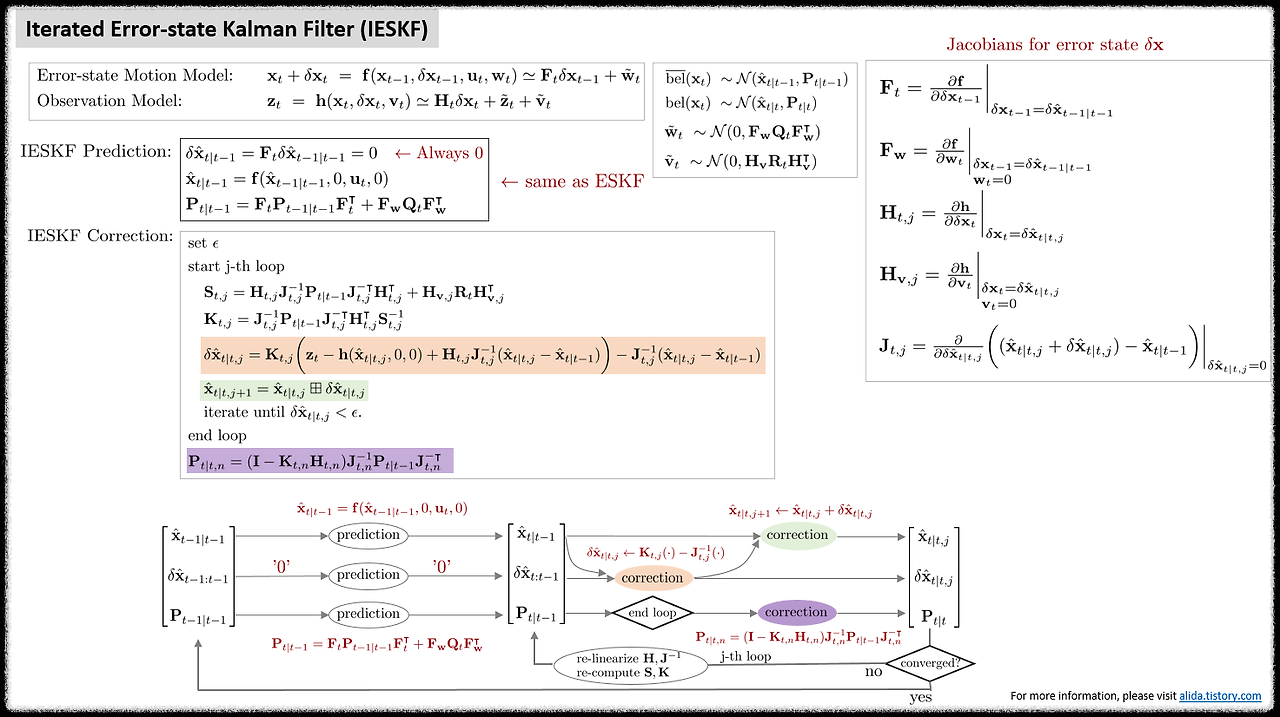}
	\end{figure}
	
	\section{Reference}
	\begin{enumerate}[label={[\arabic*]}]
		\item \href{https://en.wikipedia.org/wiki/Kalman_filter}{(Wiki) Kalman Filter}
		\item \href{https://arxiv.org/abs/1711.02508}{(Paper) Sola, Joan. "Quaternion kinematics for the error-state Kalman filter." arXiv preprint arXiv:1711.02508 (2017).}
		\item \href{https://youtu.be/wVsfCnyt5jA}{(Youtube) Robot Mapping Coure - Freiburg Univ}
		\item \href{http://jinyongjeong.github.io/2017/02/14/lec03_kalman_filter_and_EKF/}{(Blog) [SLAM] Kalman filter and EKF(Extended Kalman Filter)
			- Jinyong Jeong} \label{ref:4}
		\item \href{https://blog.csdn.net/liu3612162/article/details/114120772}{(Blog) Error-State Kalman Filter understanding and formula derivation - CSDN} \label{ref:5}
		\item \href{https://arxiv.org/pdf/2102.03804.pdf}{(Paper) He, Dongjiao, Wei Xu, and Fu Zhang. "Kalman filters on differentiable manifolds." arXiv preprint arXiv:2102.03804 (2021).} \label{ref:6}
		\item \href{https://github.com/gaoxiang12/slam_in_autonomous_driving}{(Book) SLAM in Autonomous Driving book (SAD book)}\label{ref:7}
		\item \href{https://arxiv.org/pdf/2010.08196.pdf}{(Paper) Xu, Wei, and Fu Zhang. "Fast-lio: A fast, robust lidar-inertial odometry package by tightly-coupled iterated kalman filter." IEEE Robotics and Automation Letters 6.2 (2021): 3317-3324.} \label{ref:8}
		\item \href{https://arxiv.org/pdf/2307.09237.pdf}{(Paper) Huai, Jianzhu, and Xiang Gao. "A Quick Guide for the Iterated Extended Kalman Filter on Manifolds." arXiv preprint arXiv:2307.09237 (2023).} \label{ref:9}
		\item \href{https://www.research-collection.ethz.ch/bitstream/handle/20.500.11850/263423/1/ROVIO.pdf}{(Paper) Bloesch, Michael, et al. "Iterated extended Kalman filter based visual-inertial odometry using direct photometric feedback." The International Journal of Robotics Research 36.10 (2017): 1053-1072.} \label{ref:10}
		\item \href{https://ieeexplore.ieee.org/stamp/stamp.jsp?tp=&arnumber=7266781}{(Paper) Skoglund, Martin A., Gustaf Hendeby, and Daniel Axehill. "Extended Kalman filter modifications based on an optimization view point." 2015 18th International Conference on Information Fusion (Fusion). IEEE, 2015.} \label{ref:11}
		\item \href{https://cgabc.xyz/posts/784a80cb/}{(Blog) From MAP, MLE, OLS, GN to IEKF, EKF} \label{ref:12}
		\item \href{https://docs.ufpr.br/~danielsantos/ProbabilisticRobotics.pdf}{(Book) Thrun, Sebastian. "Probabilistic robotics." Communications of the ACM 45.3 (2002): 52-57.} \label{ref:13}
	\end{enumerate}
	
	\section{Revision log}
	\begin{itemize}
		\item 1st: 2020-06-23
		\item 2nd: 2020-06-24
		\item 3rd: 2020-06-26
		\item 4th: 2023-01-21
		\item 5th: 2023-01-31
		\item 6th: 2023-02-02
		\item 7th: 2023-02-04
		\item 8th: 2024-02-08
		\item 9th: 2024-02-09
		\item 10th: 2024-05-02
		\item 11th: 2024-06-12
		\item 12th: 2024-06-23
		\item 13th: 2024-06-28
	\end{itemize}

\end{document}